\PassOptionsToPackage{hyphens}{url}
\documentclass[%
 reprint,
 aps,
 pre,
 showkeys,
 floatfix
]{revtex4-2}

\usepackage{graphicx}
\usepackage{dcolumn}
\usepackage{bm}
\usepackage{color}
\usepackage{xcolor}
\usepackage[normalem]{ulem}

\usepackage{url}
\usepackage{amssymb}
\usepackage{amsmath}
\usepackage{hyperref}

\begin{document}

\preprint{APS/123-QED}

\title{AI Teaches Humans to Handle Complexity in the Game of Go}

\author{Adamo Cerioli\textsuperscript{1}, 
Vito D.~P.~Servedio\textsuperscript{2},
Gabriele Cerioli\textsuperscript{3},
Antonio Villani\textsuperscript{1},
Carlo Metta\textsuperscript{4, *}}

\affiliation{
 \textsuperscript{1}Department of Mathematics, Physics and Computer Science, University of Parma, Parco Area delle Scienze, 7/A, 43124, Parma, Italy
}%
\affiliation{
 \textsuperscript{2}Complexity Science Hub, Metternichgasse 8, 1030, Vienna, Austria
}
\affiliation{
  \textsuperscript{3}Department of Physics, University of Trieste, Via A. Valerio 2, 34127 Trieste, Italy
}%
\affiliation{
 \textsuperscript{4}Center for Nanotechnology Innovation (NEST-IIT), P.zza San Silvestro 12, 56127 Pisa, Italy
}%
\affiliation{
 \textsuperscript{*}Corresponding author: carlo.metta@isti.cnr.it
}%

\begin{abstract}
\noindent
Strategic decision-making in complex environments requires balancing local stability against global coordination. Using the game of Go as a bounded empirical testbed, we investigate how biological and artificial agents navigate this trade-off by formalizing a network-derived metric of strategic tension defined via the spectral radius of spatial dependency matrices. Across human skill progression, tension systematically decays from novices to masters as uncoordinated tactical skirmishes give way to disciplined territorial stabilization. However, a longitudinal analysis of professional play spanning 1980--2026 reveals a striking historical paradox: following the superhuman breakthrough of AlphaGo in 2016, elite human matches display a marked surge in sustained tension. A two-dimensional projection based on move distance and captures places modern professionals close to neural engines, separating their tension from the local contact fighting of novices and from random noise. Consequently, strategic tension acts as a macroscopic order parameter for decision paradigms rather than raw computational strength. Beyond the game of Go, these findings establish a constructive paradigm for human-AI interaction, indicating how artificial systems can act as cognitive scaffolding that expands biological heuristics and elevates human capacity to navigate higher-order complexity across professional training and decision-making domains.
\end{abstract}


\maketitle

\noindent 
Many competitive systems face a basic trade-off between expansion and resilience: agents can consolidate what they already control or extend their reach into new domains, at the risk of becoming fragile \cite{march1991exploration, hills2015exploration}. The same dilemma appears at very different scales, from foraging animals deciding whether to keep exploiting a local patch or move on \cite{charnov1976optimal}, to states trading internal cohesion for territorial growth \cite{kennedy2010rise}, to firms choosing between their core competencies and uncertain new markets \cite{levinthal1993myopia}. Moreover, the trade-off rarely concerns a single front: agents usually have to manage several unresolved sectors at once, distributing limited resources among them \cite{roberson2006colonel}. Overexpansion spreads resources too thin and leaves the system exposed to collapse \cite{ scheffer2001catastrophic}, while excessive caution leads to premature consolidation, stagnation and loss of initiative \cite{levitt1988organizational}. In natural and socioeconomic systems, however, the mechanisms behind these decisions are hard to isolate, because information is incomplete and outcomes are noisy. Games of perfect information avoid this problem: every move and position is observable, which makes them a controlled setting for studying these trade-offs quantitatively \cite{vonNeumann1944}. For this reason, they have long served as model systems both in cognitive science \cite{chase1973perception, reitman1976skilled} and in the physics of complex networks \cite{georgeot2012game, kandiah2014move, coquide2017distinguishing}.

Among board games, Go is a natural model system for these questions: it is inherently spatial, and its state space is far too large for exhaustive search \cite{harre2011aggregate, tromp2006combinatorics}, making it a benchmark for testing artificial intelligence (AI) performance \cite{morandin2019sai, allis1994searching}. Played on a $19\times 19$ grid, two players alternate placing stones to enclose territory. Once placed, stones remain on their intersections unless captured. Orthogonally connected stones of the same color form a chain, here termed a \textit{group}, and share their \textit{liberties}, the empty intersections directly adjacent to the group. A group is captured and removed from the board when its last liberty is occupied by the opponent. This simple rule set generates a state space so vast that brute-force calculation is impossible, forcing human players to rely heavily on visual heuristics, memory and deep geometric intuition \cite{bossomaier2012neuro}. 

The core strategic tension in Go manifests in the continuous assessment of local stability versus global potential. A player must constantly assess whether a group is alive, meaning secure against capture \cite{benson1976life}, and whether the local position requires another move or can be left to pursue an opportunity elsewhere. Equally challenging is the decision of whether to expend valuable resources saving a struggling local group, or to intentionally sacrifice it to build broader global influence.

Historically, navigating these spatial trade-offs relied on centuries of human experience and heuristics. While computational Go steadily advanced from hand-crafted expert systems to early Monte Carlo Tree Search (MCTS)---marked by milestones such as \textit{Crazy Stone} around 2006 \cite{coulom2006efficient, gelly2012grand}---this changed abruptly in 2016, when AlphaGo defeated a top professional player \cite{silver2016mastering, silver2017mastering}. Its self-play successors later mastered the game without human knowledge~\cite{silver2018general}. This breakthrough led human professionals to rethink long-standing heuristics: by integrating neural engines into their daily training~\cite{kang2022ai}, players learned to abandon traditional biases toward premature simplification~\cite{shin2021human}. Yet, human players remain bounded by finite working memory and cognitive load \cite{kahneman2011thinking, sweller1988cognitive}. This evolutionary sequence, proceeding from classical human intuition to superhuman synthetic play and ultimately to AI-informed human adaptation, offers an ideal natural experiment to track how biological agents assimilate non-human solutions to immense combinatorial complexity.

To quantify how humans and artificial agents navigate this strategic domain, and specifically to assess the degree to which human play adapted under synthetic influence, we model the board as a spatial interaction network of mutual dependencies between established stone groups and dynamically identified \textit{contested zones}. Utilizing the open-source engine KataGo \cite{wu2019accelerating} as an evaluative oracle, we compute the co-variation in spatial ownership across the grid. We define strategic tension as the spectral radius $\lambda_1$ of the resulting dependency matrix, which provides, to our knowledge, the first position-level measure of strategic tension in Go. Macroscopically, a low $\lambda_1$ reflects a modular, decoupled board state where local skirmishes can be evaluated independently without systemic risk. In contrast, an elevated $\lambda_1$ characterizes an entangled, highly coupled regime where tactical uncertainties across disjoint fronts are globally interdependent, ensuring that a single local intervention can trigger cascading strategic consequences across the entire board so that a single local intervention can affect the status of several groups and zones at once.

We evaluate this framework across an extensive longitudinal corpus of professional tournament games spanning nearly half a century (1980--2026), benchmarked against matches generated across a wide computational spectrum of AI systems, from mixed-agent tournaments to superhuman neural engines. By comparing these trajectories across historical eras and computational budgets, we examine how human play changed after the introduction of AI \cite{chen2025generative, liu2026tool}. Previous work has linked AI exposure in board games to higher move quality and greater novelty~\cite{shin2023superhuman, choi2025does, gaessler2023training}, but also to convergence onto fewer patterns~\cite{choi2025dual} and to a period of rapid change in professional opening strategies after AlphaGo, followed by stabilization~\cite{beheim2025opening}. Our framework complements these findings with a spatial perspective: after AI, professional players sustain more strongly coupled board configurations, closer to those of the engines.

To ensure a comprehensive evaluation, we also analyze gameplay across a broad spectrum of proficiencies, comparing human matches across skill tiers \footnote{Go players are commonly classified using the kyu--dan ranking system. Amateur ranks progress from higher kyu numbers (weaker players) down to 1-kyu, followed by amateur dan ranks. Professional players are ranked separately from 1-dan professional (1p) to 9-dan professional (9p).} with KataGo simulations constrained by varying search budgets, measured via MCTS visit counts \footnote{In Monte Carlo Tree Search (MCTS), the visit count specifies the computational budget allocated to evaluate a position, defined as the number of search simulations executed to expand candidate branches. Higher visit counts correspond to deeper tactical lookahead and greater playing strength.}.

Our analyses reveal that strategic tension follows a characteristic macroscopic trajectory reflecting the distinct evolutionary phases of a game. Rather than a simple dichotomy between biological and synthetic play, these profiles establish a nuanced structural hierarchy: mixed AI tournaments sustain the highest tension due to tactical asymmetries, whereas pure superhuman engines operate at levels comparable to, or only slightly above, modern AI-informed professionals, who in turn systematically surpass historical pre-AI cohorts.

Although these temporal profiles parallel recent findings in chess \cite{cerioli2025ai}, a striking divergence emerges across skill gradients. In stark contrast to chess, where strategic tension rises monotonically with proficiency, tension in Go decays systematically with human rank, as localized beginner skirmishes that produce tactical noise give way to global spatial efficiency. 

Crucially, projecting gameplay onto a structured topological two-dimensional space reveals that elevated tension is structurally degenerate, resolving it into three distinct operational regimes: the localized, high-friction contact fighting typical of amateur play, the uncoordinated dispersion of stochastic noise, and the structured, whole-board coordination sustained by expert humans and advanced AI architectures. By isolating these dynamics within a mathematically bounded laboratory, our results suggest that the operational disparity between biological cognition and synthetic agents reflects fundamental differences in how systems manage multi-front interdependencies, balancing immediate local stabilization against global structural resilience.

\section{Defining Tension in Go}
\noindent
In the game of Go, ``tension'' does not emerge from the direct threat of piece-to-piece capture as in chess, but rather from the unresolved ambiguity of spatial control. It manifests as a state of latent structural dependency, where the stability of a stone group or the potential of a contested region is contingent upon future placements across the board. Local exchanges may resolve these dependencies, but they may also preserve or create tactical possibilities that affect play elsewhere. Tension persists while such dependencies remain unresolved, requiring both players to assess how local continuations may affect group safety and territorial control across the board.

To quantify this dynamic, we shift from individual stones to higher-order structural units. When the status of a group is unsettled, the uncertainty does not stay local but affects the evaluation of neighbouring groups and regions. We capture these dependencies with a network whose nodes are stone groups and contested zones, and whose links measure their mutual influence as evaluated by KataGo (see Methods).

\begin{figure}[!htbp]
\centering
\includegraphics[width=1\linewidth]{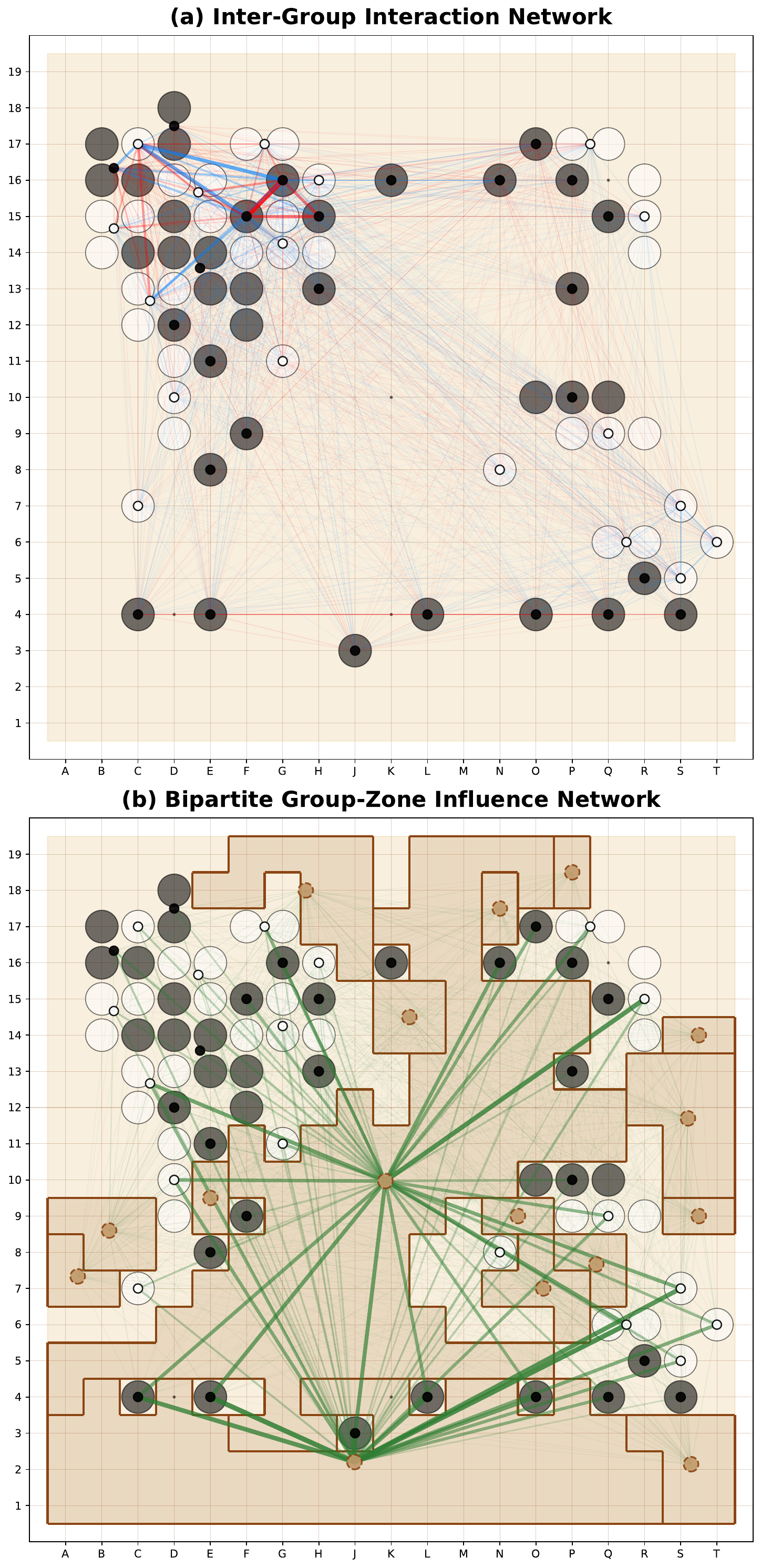}
\caption{\label{fig:network} \textbf{Topological decomposition of the strategic interaction network.} 
Evaluated via KataGo queries (100 base visits, 10 perturbation visits) at ply 75 of Game 2 from \textit{Lee Sedol vs. AlphaGo} (Seoul, March 10, 2016). 
\textbf{(a)} Direct tactical dependencies between stone groups (solid black and white nodes). Blue and red links denote positively ($\Delta_i \Delta_j > 0$) and negatively ($\Delta_i \Delta_j < 0$) correlated local perturbations, respectively. 
\textbf{(b)} Structural coupling between active stone groups and contested zones (shaded tan polygons with brown boundaries), connected by green links. 
Link widths and opacities are normalized to scale with interaction strength $|w(i,j)|$; sub-threshold interactions are suppressed to highlight dominant board-wide dependencies.}
\end{figure}

Given the state of a Go board as in Fig.~\ref{fig:network}, we evaluate the board's structural configurations at each ply. To do so, we first employ the open-source engine KataGo to generate an ownership map, which assigns to each intersection a value $O \in [-1, +1]$ representing the predicted probability of eventual territorial control. To ensure this spatial field reflects game properties rather than engine-specific artifacts, we cross-validated it against Leela Zero~\cite{LeelaZero}(see Appendix~\ref{sec:katago_ownership_and_robustness}). Based on this ownership field, we partition the board into two distinct types of structural units that constitute the nodes of our network:
\begin{description}
    \item[Stone Groups] Orthogonally connected chains of same-color stones, here termed groups, which share liberties and are captured as a unit (see Fig.~\ref{fig:network}a). Their liberty counts provide a local indicator of tactical vulnerability, although their safety also depends on eye shape and the surrounding position.
    \item[Contested Zones] Unresolved spatial regions composed of vacant intersections where territory is actively fought over (see Fig.~\ref{fig:network}b). These are identified from the ownership map where the territorial control is highly ambiguous, and are dynamically grouped into connected components representing distinct areas of strategic competition (see Methods).
\end{description}

The links of the network capture the latent interaction between these spatial entities. We define the weight of the link connecting node $i$ and node $j$ as:
\begin{equation}
w(i,j) = F(d_{ij}) \left| \overline{\Delta_i \Delta_j} \right| \frac{\sqrt{E_i E_j}}{N}
\label{eq:link_weight}
\end{equation}
where $N=19$ is the board size. The weights $w(i,j)$ form the weighted adjacency matrix $W$ of the network, and their components reflect specific structural properties:
\begin{description}
    \item[Spatial Locality ($F(d_{ij})$)] An exponential decay factor $F(d_{ij}) = e^{-d_{ij}/L_{0}}$ that penalizes interactions over large distances. Critically, $d_{ij}$ is defined as the minimum Manhattan distance between the two entities on the grid—yielding $d_{ij} = 1$ for spatially adjacent elements—thereby capturing the immediate shortest path of tactical engagement. The characteristic length $L_0=\sqrt{m/\pi}$, where $m$ is the number of legal moves, is the radius of a disc whose area equals the free space on the board \footnote{The number of legal moves $m$ coincides with the number of empty intersections, except for those where a move is forbidden by the rules, such as suicide or an immediate ko recapture.}; it shrinks as the board fills, reflecting that conflicts become increasingly local in the endgame.
    \item[\textbf{Susceptibility ($|\overline{\Delta_i \Delta_j}|$)}] A measure of coupled positional sensitivity. The signed shift $\Delta_i = \langle O_i^{\text{pert}} \rangle - \langle O_i^{\text{base}} \rangle$ is the change in mean ownership over the liberties of group $i$, or over the intersections of zone $i$, under a perturbation in which the player to move places a stone on a randomly chosen empty intersection. The sign of $\Delta_i \Delta_j$ indicates whether the ownership of the two entities shifts toward the same player ($\Delta_i \Delta_j > 0$, blue links in Fig.~\ref{fig:network}) or toward opposite players ($\Delta_i \Delta_j < 0$, red links) \footnote{Since ownership is signed (positive for Black, negative for White), the strategic meaning depends on colour: for two groups of the same colour, a positive product means they strengthen or weaken together, whereas for groups of opposite colours it means that one gains what the other loses.}. The product is then averaged over many independent perturbations, and its absolute value $|\overline{\Delta_i \Delta_j}|$ gives the net magnitude of the interdependence while ensuring that all weights are non-negative.
    
    \item[\textbf{Vulnerability and Influence ($E_i, E_j$)}] Scaling factors whose role depends on the link type. In group-group links, liberties measure vulnerability: a group with few liberties is closer to be captured and more tightly bound to its neighbours, so $E_i = 1/l_i$. In group-zone links, liberties measure the group's contact with open space, so $E_i = l_i$, while $E_j = n_j^2/N$, where $n_j$ is the number of intersections in zone $j$, i.e.\ the points at stake there. The link weight thus grows linearly with the zone's area.
\end{description}

The resulting Go network is intrinsically weighted to capture continuous tactical influence, leveraging a hybrid foundation that merges deterministic, rule-based features (position-derived geometry) with probabilistic neural network evaluations (engine-derived KataGo ownership maps). As illustrated in Fig.~\ref{fig:network}, the complete interaction network synthesizes two distinct topologies: panel (a) isolates direct tactical dependencies between stone groups (group-group links), while panel (b) maps the influence correlation between groups and territorial vacuums (group-zone links). Zones are not linked directly to each other: their interactions are mediated by groups. Together, these interconnected layers synthesize the complete interaction network. 

Among the observables extracted from this network, we use the spectral radius $\lambda_1$, the leading eigenvalue of the weighted adjacency matrix $W$, as our measure of global tension, following the approach introduced for chess \cite{cerioli2025ai}. Unlike pairwise measures, $\lambda_1$ reflects the collective structure of the network: it grows when many entities are strongly and simultaneously coupled, so that a single move can affect the status of several groups and zones at once. Since $\lambda_1$ sets the asymptotic growth rate of walks on the graph \cite{van2023graph} and increases with the density of strongly connected subgraphs, it responds both to tightly coupled local fights, such as capturing races (\textit{semeai}), and to broader dependencies, such as invasions or reductions of large frameworks (\textit{moyo}).

Because $W$ is non-negative, the Perron--Frobenius theorem \cite{pillai2005perron} guarantees that $\lambda_1$ is real and non-negative, with a non-negative leading eigenvector whose components identify the entities contributing most to the tension (see Appendix~\ref{sec:spectral_radius_tension}). In dynamical terms \cite{wang2003epidemic, castellano2017relating}, $\lambda_1$ measures how strongly a perturbation can propagate and amplify through chains of dependencies: the higher its value, the harder it is to resolve one area of the board without affecting others. Since its absolute value depends on normalization choices, we use $\lambda_1$ comparatively, across game phases, eras and skill levels.

Before presenting the statistical analysis across large game ensembles, we first illustrate the temporal evolution of strategic tension in a single representative game (see Appendix~\ref{subsec:example_tension}) and verify the robustness of the metric under variations of the main computational parameters. The single-game profile highlights the local fluctuations underlying the macroscopic trends. Across many games, tension turns out to be largely independent of the win-rate, indicating that it captures something distinct from who is ahead (see Appendix~\ref{subsec:ensemble_tension_analysis}). Robustness analyses on KataGo self-play games show that the tension profile is stable across engine search depths and perturbation schemes (see Appendix~\ref{subsec:appendix_robustness}), while variations in the spatial interaction scale (see Appendix~\ref{subsec:robustness_locality}) and contested-zone segmentation thresholds modify the absolute tension magnitude but preserve the overall temporal evolution of the tension profiles (see Appendix~\ref{subsec:robustness_zones}).

\section{Result 1: Strategic Tension Hierarchy and Historical Evolution}
\label{sec:result_tension}
\begin{figure}[!htbp]
\centering
\includegraphics[width=1\linewidth]{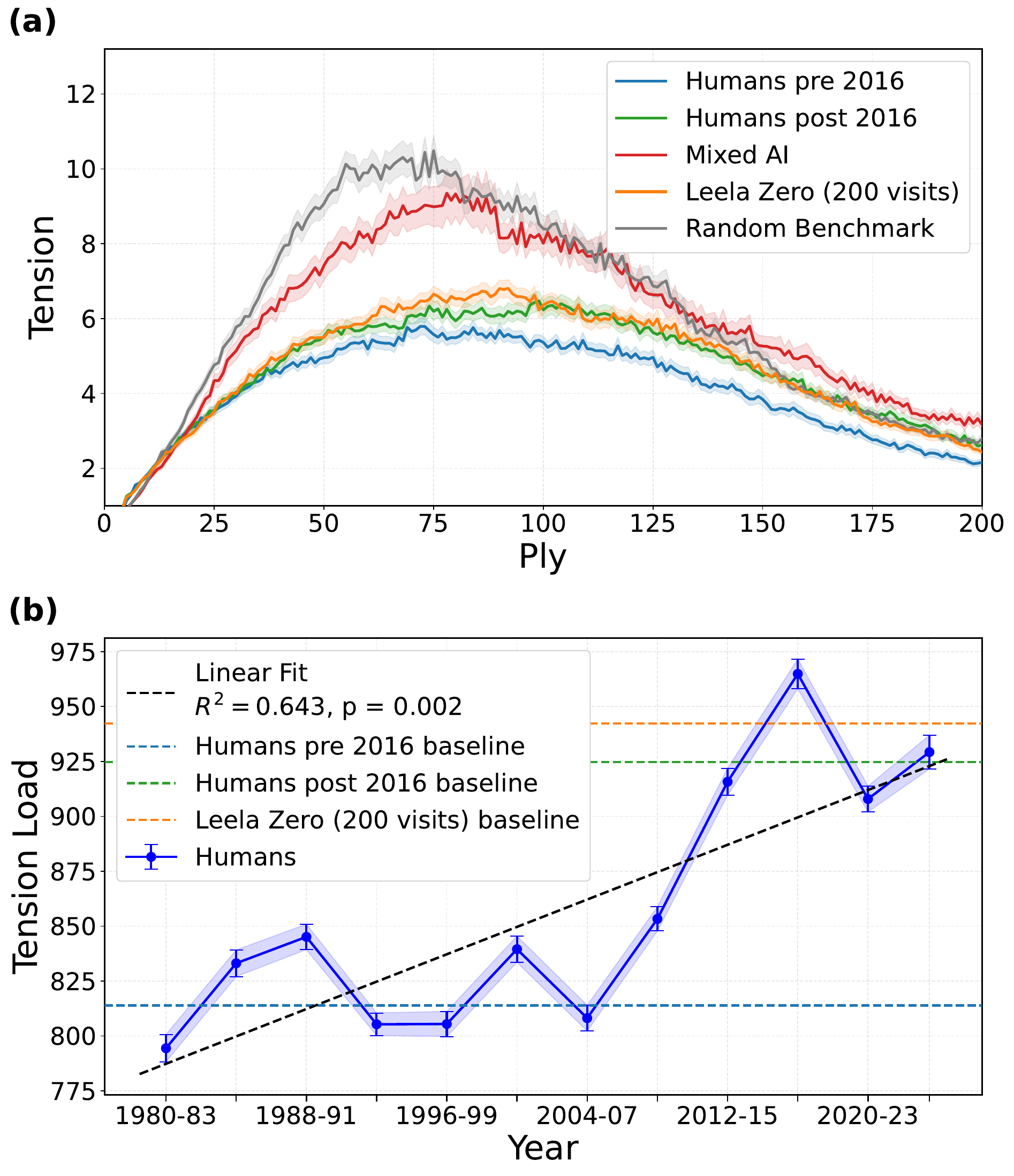}
\caption{\label{fig:tension_evolution} \textbf{Strategic tension dynamics and historical evolution across human and artificial intelligence play.} 
\textbf{(a)} Move-by-move mean strategic tension ($\lambda_1$) across 200 plies for pre-2016 human professionals (2000--2015, blue), post-2016 human professionals (2016--2026, green), Leela Zero at 200 visits (orange), tournament Mixed AI (red), and the Random Benchmark (gray). 
\textbf{(b)} Evolution of human Tension Load (integrated tension across plies 1--200) across 47 years (1980--2026), aggregated into 4-year bins (with the final bin covering 2024--2026; $N \approx 150$ games per bin). The black dashed line shows a weighted linear fit ($R^2 = 0.643$, $p = 0.002$). Horizontal dashed lines denote the cumulative baselines for pre-2016 humans (blue), post-2016 humans (green), and Leela Zero at 200 visits (orange). Error bars and shaded ribbons denote $\pm 1$ standard error of the mean.}
\end{figure}

\noindent
We measure strategic tension as the spectral radius $\lambda_1$ of the interaction network and follow it move by move (see Fig.~\ref{fig:tension_evolution}a). We examine five distinct ensembles: human professionals before 2016 (2000--2015) and after 2016 (2016--2026), when top-level play became influenced by deep reinforcement learning engines, Leela Zero with a fixed budget of 200 visits, a mixed ensemble of tournament engines (Mixed AI), and a Random Benchmark in which each move is drawn uniformly from the legal moves, establishing a null model of non-strategic play.

All ensembles follow the same overall shape: tension builds up during the opening, peaks in the middlegame, when fights are most open, and declines in the endgame as groups are settled (the mechanisms behind this decline are discussed in Appendix~\ref{subsec:robustness_drops}). 

Random play produces the highest tension for most of the game. This is expected, since uncoordinated moves leave many weak and unresolved groups, indicating how much tension arises without any strategy. Among strategic players, a clear hierarchy emerges. Mixed AI stays well above all other ensembles and, from roughly ply 130 onward, even exceeds the random baseline. Leela Zero sits slightly above post-2016 professionals around the middlegame peak, and post-2016 professionals are consistently above pre-2016 ones, suggesting that modern players accept more unresolved situations on the board than their predecessors did.

Furthermore, decomposing the interaction matrix into the group--group and group--zone contributions confirms that macroscopic tension is predominantly governed by contested territorial frontiers (see Appendix~\ref{subsec:tension_decomposition}).

The higher tension of Mixed AI is not a by-product of more lopsided results (see Appendix~\ref{subsec:robustness_competitive_balance}). When comparing games with similar final score margins, restricting the analysis to the range of margins observed in both groups, Mixed AI exhibits a higher Tension Load than modern professionals.

These differences reflect systematic shifts in board topology (see Appendix~\ref{sec:structural_analysis}). In particular, mixed-agent AI tournaments feature notably more stone captures per game and fewer liberties per group in the early stages of the game, indicative of a more fragmented and tactically intensive interaction network.

To quantify the historical evolution of human gameplay, we define the \textit{Tension Load} of an ensemble as the sum of its mean tension profile over plies 1--200,
\begin{equation}
\mathcal{T} = \sum_{k=1}^{200} \bar{\lambda}_1(k), \qquad \bar{\lambda}_1(k) = \frac{1}{N_k} \sum_{g \,:\, L_g \ge k} \lambda_1^{(g)}(k),
\end{equation}
where $\lambda_1^{(g)}(k)$ is the tension of game $g$ at ply $k$, $L_g$ is the length of game $g$, and $N_k$ is the number of games that reach ply $k$. Averaging only over games still in progress at each ply prevents shorter games from biasing the profile toward zero (see Fig.~\ref{fig:tension_evolution}b). Tracking 47 years of professional matches (1980--2026) reveals a statistically significant increase ($R^2 = 0.643$, $p = 0.002$). 

The increase, however, did not start with AlphaGo. Following a reasonably stable plateau from 1980 to 2007, human tension load accelerated after 2008, coinciding with early MCTS milestones like Crazy Stone, and surged to an acute peak in 2016--2019 following AlphaGo's breakthrough before consolidating at an elevated baseline. This timing matches that reported for professional opening strategies, which changed most rapidly shortly after AlphaGo and then stabilized~\cite{beheim2025opening}. This progression reflects an ongoing strategic alignment of human play with AI paradigms, marking an evolutionary transition from classical heuristics toward the high-coupling, multi-front coordination characteristic of neural engines.

To understand what drives this hierarchy, we examine local spatial move distributions (see Appendix~\ref{sec:spatial_analysis}), measuring edge preferences, distances between consecutive moves, and local stone adjacencies. Mixed AI plays farther from the previous move, closer to the board edge, and less often in contact with enemy stones than human professionals, whereas Leela Zero remains close to human values. Modern professionals have not moved toward these patterns: compared with pre-2016 players, they play slightly closer to the previous move, farther from the edge, and more often in contact with enemy stones. The rise in human tension is therefore unlikely to reflect changes in simple spatial habits such as move distance, edge preference, or contact frequency.

To test whether such spatial habits alone explain the tension levels, we build null models that reproduce each ensemble’s move statistics without any lookahead (see Appendix~\ref{sec:null_model}). Crucially, the Mixed AI-inspired spatial null model yields a significantly lower tension profile than the others. The high tension of Mixed AI therefore does not come from where its moves land, but from how they are coordinated across the board.

Finally, error profiles help interpret the gap between humans and engines (see Appendix~\ref{sec:appendix_error_analysis}). In balanced positions, where the outcome is still open, human error rates rise steadily through the late middlegame as fights accumulate. Mixed AI engines, in contrast, seem to maintain low error rates across all plies. Within limits, the comparison suggests that human players have learned to play in more complex positions, but that handling this complexity remains harder for them than for engines.

\section{Result 2: Skill-Dependent Scaling of Strategic Tension}
\label{sec:result_rank}
\begin{figure}[!htbp]
\centering
\includegraphics[width=1\linewidth]{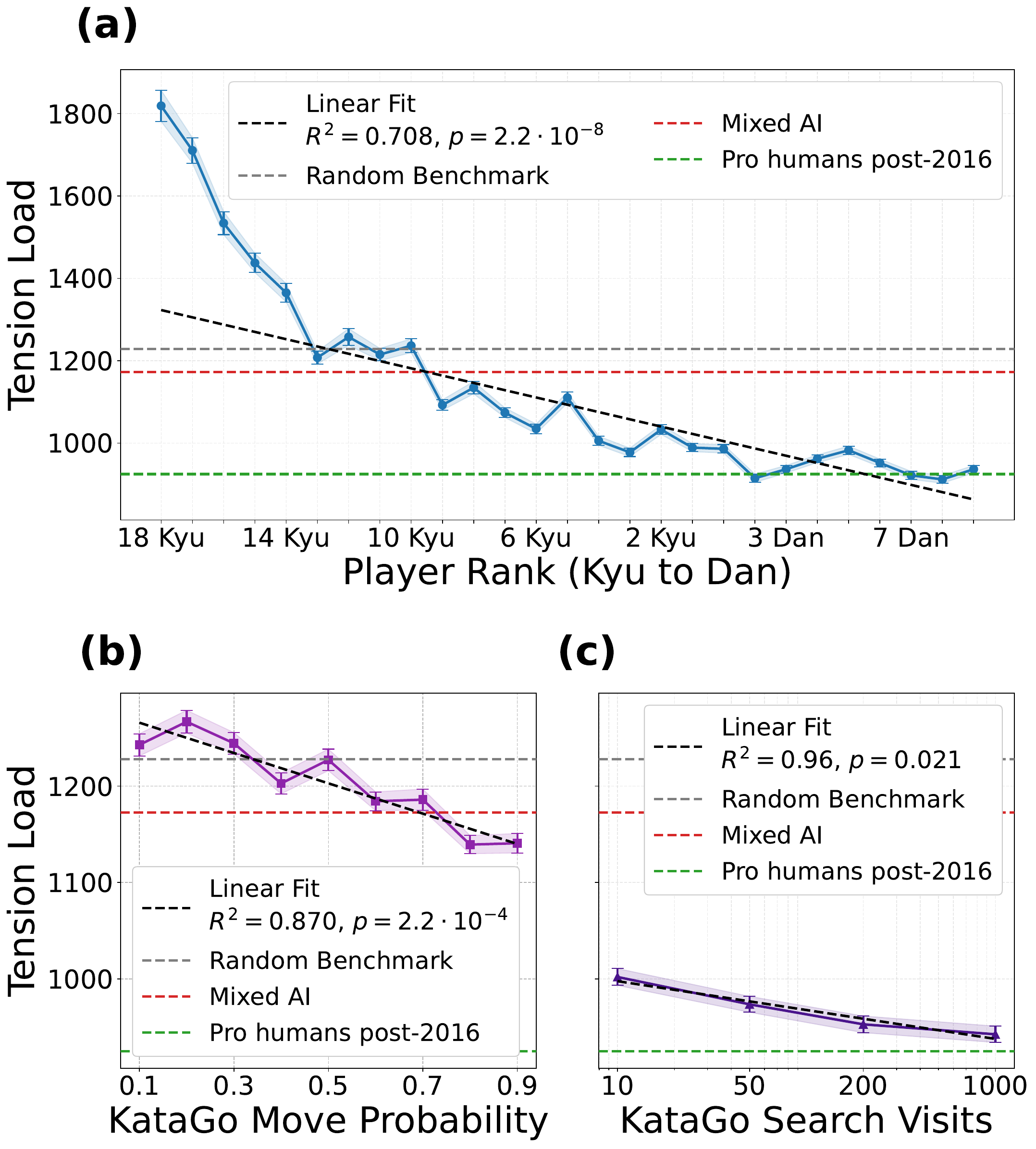}
\caption{\label{fig:tension_rank} \textbf{Strategic tension load across human expertise, policy noise, and MCTS search depth.} 
\textbf{(a)} Skill-dependent scaling of tension load across human player ranks from 18 Kyu to 9 Dan, all played between 2018 and 2019. Solid blue line and shaded area denote the mean tension load and standard error. The black dashed line shows a weighted linear fit. 
\textbf{(b)} Effect of uniform noise in an $\epsilon$-greedy policy as a function of KataGo move probability $1 - \epsilon$. 
\textbf{(c)} Scaling of tension load across MCTS search depth ($10$ to $1000$ visits) on a logarithmic scale. All synthetic KataGo games in (b) and (c) were initialized using the first 15 plies extracted from
real professional human openings. Each data point corresponds to 200 games. Horizontal dashed lines across all panels provide reference baselines: Random Benchmark (grey), Mixed AI baseline (red), and post-2016 professional humans (green).}
\end{figure}

\noindent
To examine how spatial dependencies scale across player expertise and decision noise, we analyze a rank-stratified corpus of modern post-2016 human games ranging from 18 Kyu to 9 Dan, alongside controlled experiments varying AI policy noise and search depth (see Fig.~\ref{fig:tension_rank}).

Tracking tension load across human ranks reveals a sharp, statistically significant inverse scaling with expertise (Fig.~\ref{fig:tension_rank}a). Novice play (18 Kyu) generates an extreme tension load far exceeding the random baseline, which then systematically declines across the amateur spectrum toward high-Dan ranks.

To uncover the physical mechanisms behind this inverse relationship, we examined the corresponding macroscopic structural metrics and micro-spatial move preferences across ranks (see Appendix~\ref{sec:rank_metrics_appendix}). This analysis suggests that the extreme tension in novice games reflects an excessive focus on local skirmishes at the expense of the whole-board balance of territory, influence and group safety. At low Kyu levels (18k--14k), players exhibit short inter-move distances ($d_{\text{prev}}$) and exceptionally high probabilities of both friendly and enemy adjacencies. Novices often play in direct contact with both their own and their opponent's stones, and their games contain many captures (see Appendix~\ref{sec:spatial_analysis}). This local, contact-heavy play raises tension through the interaction kernel: stones placed close to each other have small distances $d(i,j)$, which strengthens their spatial coupling, and frequent fights keep nearby areas unsettled. Together, these effects produce large link weights $w(i,j)$ in Eq.~\eqref{eq:link_weight} and raise the spectral radius, even though this tension comes from local crowding rather than from strategic planning.

As player expertise advances from 18 Kyu toward Dan ranks, players internalize shape efficiency, group connectivity, and whole-board awareness. Successive moves span larger distances across the grid and local stone adjacencies decline, reflecting selective local engagement and timely transitions to global opportunities (\textit{tenuki}). Meanwhile, captured stones drop markedly as players master group safety and avoid doomed skirmishes. Consequently, the tension load crosses below the random benchmark around 10 Kyu and asymptotically converges toward the post-2016 professional baseline at high-Dan levels.

To test whether non-strategic play inherently inflates tension or whether this effect is specific to human contact-fighting biases, we evaluated the impact of uniform move noise (Fig.~\ref{fig:tension_rank}b) alongside search depth scaling in KataGo (Fig.~\ref{fig:tension_rank}c). To model decision noise, moves were selected according to a noise-perturbed policy where, at each ply, the agent follows KataGo's policy evaluated at 10 visits with probability $1-\epsilon$, and samples a move uniformly at random from all legal moves with probability $\epsilon$. Unlike novice humans, whose blunders cluster into tight contact battles, uniform stochastic noise scatters stones across the grid without forming local clusters or global coordination. Consequently, as the KataGo move probability $1-\epsilon$ decreases, tension load rises steadily toward the random benchmark. Conversely, varying the search visits from 10 to 1000 in symmetric self-play produces a smooth, log-linear modulation of tension, remaining consistently bounded between professional and superhuman baselines. 

Notably, an unbalanced encounter between KataGo at 2 visits and 1000 visits yields a tension load of $1028.3 \pm 6.2$, with the 1000-visit engine winning by an average score margin of $+57.8 \pm 30.0$ points (evaluated at 500 visits per position). This confirms that while symmetric expertise contains tension within a controlled regime, sharp skill asymmetries trigger severe tactical instability as the stronger agent exploits overplays across the board.

\section{Result 3: Three Regimes of Strategic Tension}
\label{sec:result_phase_space}
\begin{figure}[!htbp]
\centering
\includegraphics[width=1\linewidth]{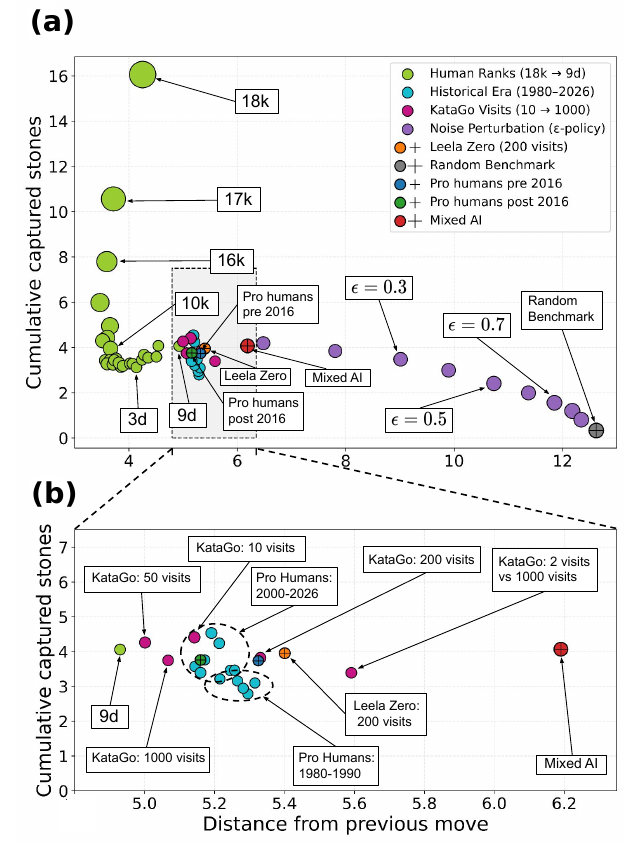}
\caption{\label{fig:phase_space_scatter} \textbf{Go ensembles in the space of move distance and captured stones.} 
\textbf{(a)} Global distribution plotted across move-to-move spatial distance and cumulative captured stones. Marker area scales as (Tension Load)$^{2}$ after normalization. Categories include Human Ranks ($18\text{k} \to 9\text{d}$, light green), Historical Era bins (1980--2026, cyan), KataGo search depths ($10 \to 1000$ visits, magenta), and $\epsilon$-policy noise perturbations (purple). Circles with internal crosses denote reference baselines: Random Benchmark (grey), Leela Zero (orange), Pre-2016 Pro Humans (blue), Post-2016 Pro Humans (green), and Mixed AI (red). 
\textbf{(b)} Magnified view of the dashed bounding box in \textbf{(a)}, showing the coordinates of expert human play, historical professional cohorts, and engines playing at equal search budgets.}
\end{figure}

\noindent
While the scalar tension load quantifies overall spatial coupling, it cannot distinguish between three distinct generative regimes: localized tactical friction from amateur contact fighting, stochastic perturbation from decision noise, and structured, long-range strategic coordination. To separate them, we project all empirical and synthetic ensembles into a two-dimensional space spanned by two macroscopic observables: the mean move-to-move spatial distance and the cumulative number of captured stones. The inter-move distance measures the spatial scope of play, from moves played in direct contact with the previous one to moves elsewhere on the board. The cumulative number of captured stones measures how often groups die, and thus how fragile groups are and how intense the fighting is. Together, they separate how spread out the play is from how violent it is, and the ensembles fall into distinct regions (see Fig.~\ref{fig:phase_space_scatter}a).

Crucially, this bidimensional space representation exposes a fundamental distinction in the physical mechanisms generating strategic tension across different playstyles:

\begin{enumerate}
    \item \textbf{Local-Instability Regime (Low-Skill / Novice Play):} In beginner games ($18\text{k}$--$14\text{k}$), high tension load is accompanied by minimal move distances and many stone captures. As skill advances toward $9\text{d}$, players increase their spatial dispersion and avoid unnecessary captures, causing tension load to decline.

    \item \textbf{Stochastic Fragmentation Regime (Random \& High-Noise Policies):} At the opposite extreme, uniform move selection (Random Benchmark and policies with noise, up to $\epsilon = 0.9$) exhibits large spatial jump distances due to unconstrained sampling across the $19\times19$ grid. Here tension comes from fragmentation: scattered, uncoordinated stones form many small, unstable groups with high susceptibilities $\Delta_i$.

    \item \textbf{Global-Coordination Regime (Superhuman AI \& Modern Pros):} Strikingly, both superhuman AI engines and post-2016 human professionals occupy a distinct, highly coordinated domain in the two-dimensional space (Fig.~\ref{fig:phase_space_scatter}a). The magnified view of this region (Fig.~\ref{fig:phase_space_scatter}b) resolves the fine-grained trajectory of human adaptation toward AI playstyles over the past decades. Tracking professional play from historical cohorts (1980--1990) to modern eras reveals a systematic convergence toward the AI cluster. This migration is driven primarily by an increase in cumulative captured stones. Conversely, asymmetric AI matchups (KataGo 2 vs.\ 1000 visits; Mixed AI) lie at much larger move distances, as the deeper-searching engine exploits expansive, non-local maneuvers against an under-searching opponent.
\end{enumerate}

\noindent
By decoupling local tactical attrition from long-range spatial coupling, this space mapping is consistent with the view that the tension of strong players, human and artificial, reflects coordination across the board rather than local friction or noise.

\section{Discussion}
\noindent
In this work, we use the game of Go as a model system to investigate how agents navigate complex decision landscapes. Beyond its vast combinatorial scale, Go couples intricate spatial geometry with centuries of recorded play, offering a uniquely suited empirical environment to compare human and artificial intelligence. Across a wide spectrum of play, ranging from human experts across distinct eras and skill levels to neural engines with controlled computational budgets and stochastic baselines, we examine how decision-makers coordinate spatial dependencies while balancing immediate local defense against global expansion.

We quantify strategic tension using spatial interaction networks, adapting methods recently applied to chess~\cite{cerioli2025ai} and complementing earlier complex-network studies of Go, including one that distinguished human from computer play~\cite{coquide2017distinguishing}. Related work on chess has also modeled how human and engine behavior differ~\cite{mcilroy2020aligning}. However, these domains operate under fundamentally distinct physical boundary conditions. Whereas chess is subtractive and dynamic, with mobile pieces preserved for downstream utility, Go is additive and cumulative, requiring stationary stones to be committed early to configurations that must remain optimal through the terminal state. 

This additive nature helps explain a clear difference in how tension changes with skill. In chess, tension rises with proficiency, whereas in Go it declines from novices to experts, and the same trend appears across AI search depths. Novice games are marked by short move distances and frequent captures, producing dense local fights that strongly increase spatial coupling. One plausible interpretation is that novices try to simplify the position by capturing stones, and in doing so create the very tension they aim to reduce \cite{hammond2006hidden}.

Consequently, elevated strategic tension $\lambda_1$ is structurally degenerate, conflating three fundamentally distinct operational regimes: topological fragmentation driven by stochastic noise in zero-intelligence baselines \cite{gode1993allocative, mckelvey1995quantal}, localized grid coupling induced by the myopic contact skirmishes of novice humans, and deliberate, whole-board coordination sustained by top professionals and superhuman AI. Because a single scalar metric cannot discriminate between amateur tactical friction and purposeful global architecture, we project game trajectories onto a two-dimensional macroscopic space spanned by the mean inter-move distance and cumulative stone captures, effectively decoupling spatial dispersion from material attrition.

Armed with this macroscopic projection, our longitudinal analysis of professional human games spanning 1980 to 2026 reveals a progressive increase in sustained strategic tension. Specifically, this upward trajectory initiated around 2008, historically coinciding with the emergence of early MCTS engines such as \textit{Crazy Stone}, and culminated in 2016 with the advent of superhuman deep reinforcement learning architectures exemplified by AlphaGo. Through this technological progression, the operational tension sustained by human masters systematically shifted closer to the AI regime.

This trajectory reflects a clear shift in human strategy driven by learning from AI. Importantly, the historical rise in tension cannot be explained by general skill improvement alone: across player ranks, expertise actually correlates with lower tension. Rather than simply playing better in the classical sense, modern masters have adopted a distinct, highly coupled style characteristic of neural engines. This is consistent with recent discussions of cultural transmission from machines to humans~\cite{brinkmann2023machine}.
In this setting, AI can serve as cognitive scaffolding~\cite{popenici2017exploring,gaessler2023training}, helping human players to move past traditional habits and avoid simplifying positions prematurely. Looking ahead, incorporating such spatial metrics into adaptive training systems \cite{diaz2024artificial} could help decision-makers learn to navigate dense, multi-front interdependencies across other complex domains.

Nevertheless, neural engines maintain a small margin in strategic tension over top professionals, which persists even at a low KataGo search budget (approximately 100 visits per move). This residual gap likely reflects biological cognitive constraints: bounded by working memory \cite{baddeley2012working} and processing capacity \cite{sweller1988cognitive}, human players proactively stabilize vulnerable groups to keep operational complexity within manageable mental horizons \cite{inglis2000central, heinemann2009measuring}. Conversely, artificial architectures face no such biological bottleneck, enabling them to sustain elevated whole-board coupling across multiple overlapping, unresolved fronts without succumbing to tactical collapse \cite{wurman2023improving}.

Beyond the full-information setting of Go, extending this framework to multiplayer games with imperfect information, such as \textit{Diplomacy} \cite{meta2022human} and \textit{Stratego} \cite{perolat2022mastering}, represents a natural next frontier. In these environments, strategic tension couples with partial observability, shifting alliances, and the threat of betrayal, providing a model system to study how agents manage deception and collective stability across decentralized networks.

More broadly, strategic tension offers an interpretable metric to monitor how decision-makers handle dense interdependencies, identify cascading vulnerabilities \cite{helbing2013globally}, and anticipate systemic failure in hybrid human-AI systems \cite{scheffer2012anticipating, punzi2026learning}. In Go, our framework captures how human professionals moved past long-standing tactical heuristics by assimilating AI paradigms, demonstrating that artificial agents can expand human problem-solving and illuminate persistent conceptual blind spots \cite{riva2026architecture, chen2025generative}. However, leveraging AI as an effective learning partner requires balance. While machine intelligence can guide humans toward richer strategic ground, educational tools must respect cognitive constraints. Without thoughtful pedagogical design, learners risk passively copying AI outputs and offloading critical reasoning \cite{jain2026cognitive, garcia2025epistemic}. Developing macroscopic, interpretable measures of complexity thus provides a principled foundation for adaptive learning environments that challenge human intuition without overwhelming cognitive capacity.

\section{Methods}

To investigate strategic tension and spatial coupling in Go, we developed an end-to-end analytical pipeline integrating game records, neural engine evaluations, and spatial network models. The mathematical formalization of contested zones and network interaction metrics was established in close consultation with advanced domain expertise (co-author CM holds a 5~Dan rank), ensuring that heuristic thresholds and graph-theoretic formulations remain faithful to established principles of Go strategy and territorial influence.

\subsection{Data}
\noindent
Go games are universally archived and processed using the Smart Game Format (SGF). This standard plain-text representation records the full sequential tree of stone placements, coordinate properties, and essential metadata, including player identities, historical contexts, and tournament time controls. Being highly structured and computationally efficient, the SGF framework serves as the ideal baseline for automated board reconstruction and large-scale engine parsing.

To capture elite human strategic decision-making across the modern era (1980--2026), professional games played by top-tier human masters were compiled from a public archive of professional tournaments \footnote{\texttt{https://homepages.cwi.nl/\textasciitilde aeb/go/games}}. This comprehensive dataset encompasses elite international majors as well as premier classical title matches spanning both the pre- and post-AI eras. Specifically, it includes major international competitions, such as the Chunlan Cup, the Japan--China Super Go, the Fujitsu Cup, the LG Cup, the Samsung Cup, and the Ing Cup, alongside the traditional Japanese major titles: the Kisei, Meijin, Honinbo, Judan, Tengen, Gosei, and Oza tournaments. These competitions were selected because of their exceptionally long time controls, which often grant several hours of main thinking time per player. This ensures that the analyzed plies reflect deep, deliberate human calculation rather than time-pressure heuristics.

To evaluate machine play across diverse algorithmic paradigms, 200 games were collected from open computer Go tournaments \footnote{\texttt{https://uec-go.com/}}. Encompassing architectures from classical heuristics to modern deep neural networks, this corpus exhibits pronounced heterogeneity in playing strength, spanning amateur to superhuman tiers. Crucially, these pairings introduce natural skill asymmetries, establishing an ideal Mixed AI benchmark to investigate how strategic tension behaves during unbalanced encounters.

To assess how strategic dynamics scale across the full spectrum of player expertise, we complemented the elite professional corpus with a rank-stratified dataset of human games spanning from 18 kyu (18k) to 9 dan (9d) extracted from \footnote{\texttt{https://github.com/featurecat/go-dataset}}. Games within each rank tier were sampled under strict criteria—specifically filtering out handicap matches and restricting pairings to opponents with a maximum rank difference of two levels—thereby providing a clean progression of human playstyles from amateur foundational play to high-dan mastery.

Finally, we also compiled a dedicated corpus of $70$ landmark games played by DeepMind's AlphaGo against elite human professionals extracted from public archives \footnote{\texttt{https://gomagic.org/download/alphago-sgf-pack-70-games/}}. This collection prominently features the historic five-game challenge series between South Korean world champion Lee Sedol and AlphaGo in March 2016, alongside subsequent high-profile encounters against top professionals. This dataset provides a unique empirical window into the direct strategic collision between classical human intuition and early superhuman deep reinforcement learning architectures.

These combined datasets provide a comprehensive evolutionary cross-section of Go strategy, spanning from deep human intuition under classical tournament conditions to the high-coupling territorial paradigms established by modern reinforcement learning systems.

\subsection{Software}
\noindent
The computational pipeline was implemented in Python, using \texttt{sgfmill}~\cite{sgfmill} to parse SGF records, \texttt{NetworkX}~\cite{hagberg2008exploring} for graph-theoretic analysis, and custom routines to enforce the deterministic rules of Go (capture mechanics, liberty tracking, suicide prohibition, and the \textit{Ko} rule). To extract spatial ownership and tactical priors, the architecture interfaces asynchronously via JSON with KataGo (v1.16.4, CUDA~12.8, cuDNN~9.8.0). Concurrently, Leela Zero (v0.17) is controlled via standard GTP commands through a multiprocess GPU worker pool, orchestrating parallel Monte Carlo rollouts under deterministic Tromp-Taylor scoring for cross-engine robustness validation and multi-tier seeded self-play generation.

\subsection{Identification of Contested Zones}
\noindent
While stone groups are objectively and unequivocally defined by the deterministic rules of Go, the concept of a contested zone is inherently more elusive. To formalize these vague spatial domains, we implement a two-stage spatial clustering framework based on KataGo's ownership map that isolates unresolved regions and segments them into locally consistent areas of competition.

\subsubsection{Spatial Masking and Components}
We first isolate vacant intersections characterized by high territorial uncertainty. Given the board matrix $B$ ($B_{x,y} = 0$ for vacant sites) and the net ownership map $O_{x,y} \in [-1, +1]$, we define a binary spatial mask:
\begin{equation}
M_{x,y} = \mathbb{I}\left(B_{x,y} = 0 \ \wedge \ \theta_{\text{lower}} < |O_{x,y}| < \theta_{\text{upper}}\right)
\label{eq:spatial_mask}
\end{equation}
where $\mathbb{I}(\cdot)$ is the indicator function. The upper threshold $\theta_{\text{upper}} = 0.75$ filters out intersections that are already heavily biased toward a specific player and can no longer be considered genuinely contested. The lower bound $\theta_{\text{lower}} = 0.05$ excludes intersections with near-zero predicted ownership, with the aim of filtering out neutral points (\textit{dame}). However, near-zero ownership can also occur at genuinely unsettled intersections and does not, by itself, identify tactically irrelevant points. We therefore test a permissive setting with $\theta_{\text{lower}} = 0.01$, together with stricter and looser values of the other thresholds: these choices change the absolute level of tension but not the shape of its profile (see Appendix~\ref{subsec:robustness_zones}). 

The connected regions of intersections satisfying the criteria above, where two intersections are connected if they are orthogonally adjacent (4-connectivity, as for stone groups in Go), are then isolated as candidate zones. Within this framework, any component containing fewer than $\kappa_{\text{min}} = 2$ intersections is discarded as background noise, as these lack the spatial capacity to form viable tactical bases.

\subsubsection{Splitting Components into Sub-Zones}
Because large-scale components often span multi-front sectors, each region undergoes a specialized agglomerative clustering to split them into localized tactical units based on ownership similarity.

Initially, each cell $(x,y)$ within a component forms a singleton cluster $z_i$ with value $z_i^{\text{own}} = O_{x,y}$. The algorithm then iteratively executes the following steps:
\begin{enumerate}
    \item \textbf{Adjacency Check:} Identify all pairs of clusters $(z_i, z_j)$ that are adjacent, i.e., contain at least one pair of orthogonally adjacent intersections.
    \item \textbf{Minimization:} Compute the ownership distance $\delta(z_i, z_j) = |z_i^{\text{own}} - z_j^{\text{own}}|$ for all adjacent pairs, selecting the pair $(z_i, z_j)$ that minimizes $\delta$.
    \item \textbf{Termination:} If $\min(\delta) > \theta_{\text{merge}}$ (where $\theta_{\text{merge}} = 0.20$) or no adjacent pairs remain, the loop terminates.
    \item \textbf{Merger:} Merge the optimal pair into a new cluster $z_{\text{new}} = z_i \cup z_j$, whose updated average ownership is defined via the area-weighted formula:
    \begin{equation}
    z_{\text{new}}^{\text{own}} = \frac{z_i^{\text{own}} |z_i| + z_j^{\text{own}} |z_j|}{|z_i| + |z_j|}
    \label{eq:merge_update}
    \end{equation}
    where $|z|$ denotes the number of intersections within a cluster.
\end{enumerate}
Only the final sub-zones satisfying $|z| \ge \kappa_{\text{min}}$ are retained and registered as contested zones.

\section*{Code availability}
\noindent
The Python code used to construct the spatial interaction networks, query the engine, and calculate strategic tension metrics in this study is openly available on GitHub at \url{https://github.com/Draghellon/Tension_Go}.

\section*{Authors' contributions}
\noindent
AC conceived the study, developed the primary code and analytical pipelines, conducted the formal analysis, generated all figures, and wrote the original draft of the manuscript. VDPS assisted in project planning and contributed to the writing and editing of the manuscript. GC assisted in implementing the KataGo analysis pipeline, curated secondary datasets, and contributed to the validation and critical revision of the manuscript. AV explored alternative formulations for the tension metric and contributed to the supplementary analysis of temporal error dynamics. CM, an experienced 5 Dan Go player, provided domain supervision, established the foundational guidelines for the tension metric, and contributed to the manuscript. All authors reviewed and approved the final manuscript.

\section*{Acknowledgements}
\noindent
The authors warmly thank the Taro-kun Go Club of Parma and its community of players for introducing them to the fundamentals of Go and for facilitating valuable connections with experienced players.

VDPS was funded by the Austrian Federal Ministry for Innovation, Mobility and Infrastructure (BMIMI) as part of the CSH Postdoc Program (GZ 2023-0.841.266). 

The authors used the Anthropic models Claude Opus 5.5 and Claude Fable 5.1 via Claude Code to refine the manuscript's wording and structure.
The authors directed every change, reviewed and corrected the suggested text, and took full responsibility for the content. 

\appendix

\section{Territorial Ownership: KataGo Heatmap and Cross-Engine Robustness}
\label{sec:katago_ownership_and_robustness}

\begin{figure}[!htbp]
\centering
\includegraphics[width=0.95\linewidth]{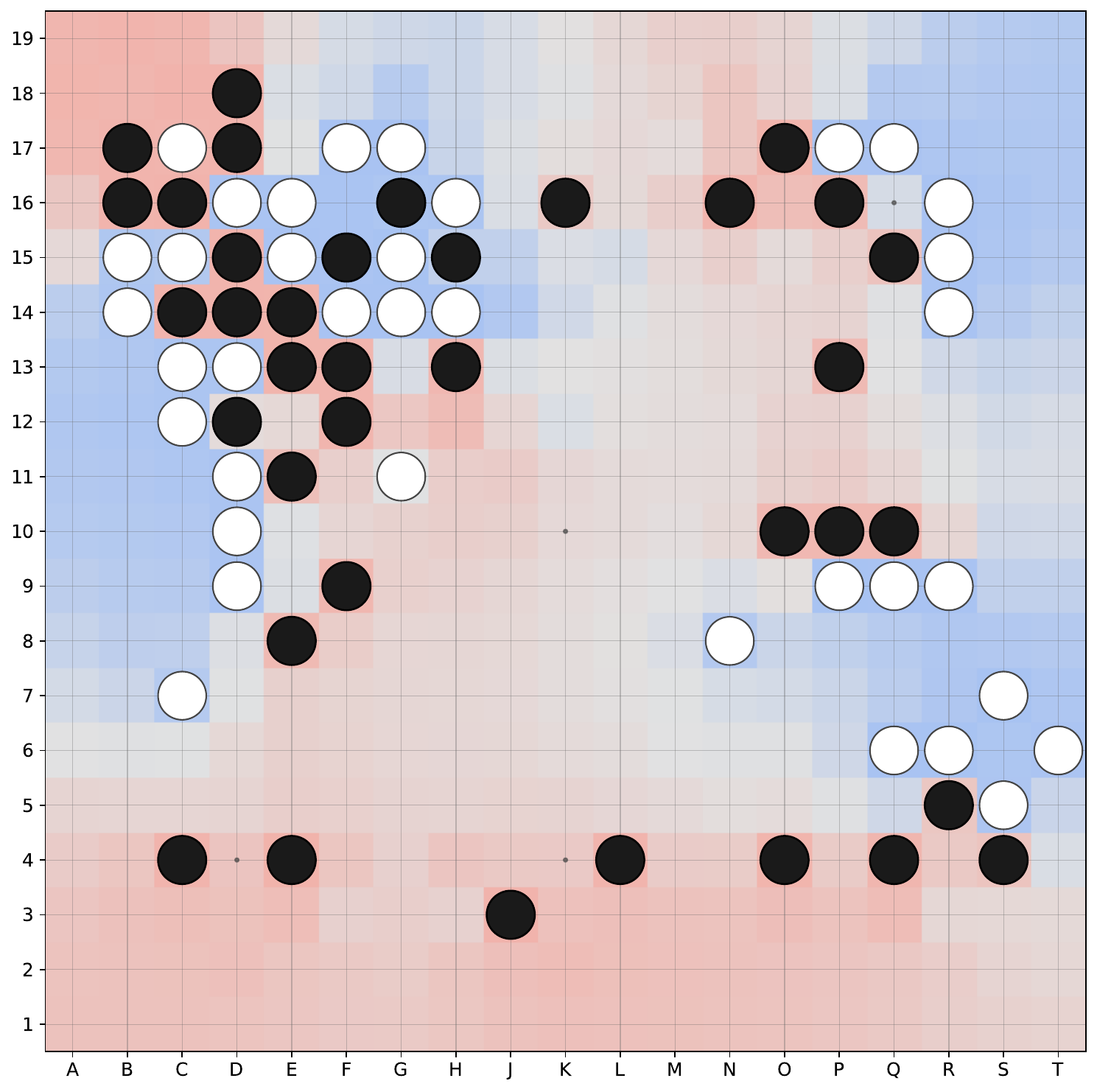}
\caption{\label{fig:own_heatmap} \textbf{Territorial ownership landscape at the strategic tension peak.} 
Evaluated with KataGo (500 analysis visits) at ply 75 of Game 2 from \textit{Lee Sedol vs. AlphaGo} (Seoul, March 10, 2016). 
Red and blue denote Black ($\to +1.0$) and White ($\to -1.0$) predicted territory, respectively, while gray regions mark unresolved territorial uncertainty.}
\end{figure}

In our interaction network framework, the structural weights of the links are fundamentally governed by the territorial control across the board. To quantify this field, we employ the continuous territorial ownership projection computed by KataGo, which assigns each intersection a value in the range $[-1, +1]$ denoting Black ($+1.0$) or White ($-1.0$) control, with intermediate values reflecting territorial uncertainty (Fig.~\ref{fig:own_heatmap}).

To verify that these ownership distributions reflect objective topological properties rather than architecture-dependent artifacts of KataGo's auxiliary head, we perform an independent cross-engine validation against Leela Zero. While KataGo outputs an end-to-end continuous expectation via its specialized head (evaluated at 20 MCTS visits), Leela Zero's ownership field is estimated empirically as the ensemble average over $N_{\text{rollouts}} = 200$ independent Monte Carlo playouts (allocated at 20 visits per move) scored deterministically with Tromp-Taylor rules. We sample $N_{\text{samples}} = 50$ configurations (game--ply pairs) uniformly across plies $20 \le t \le 200$ from the corpus of $N_{\text{games}} = 70$ games between AlphaGo and professional players. Crucially, all comparative metrics are evaluated exclusively on empty intersections ($s \in \mathcal{V}_{\text{empty}}$) to eliminate trivial stone self-correlations.

\begin{figure}[!htbp]
\centering
\includegraphics[width=1\linewidth]{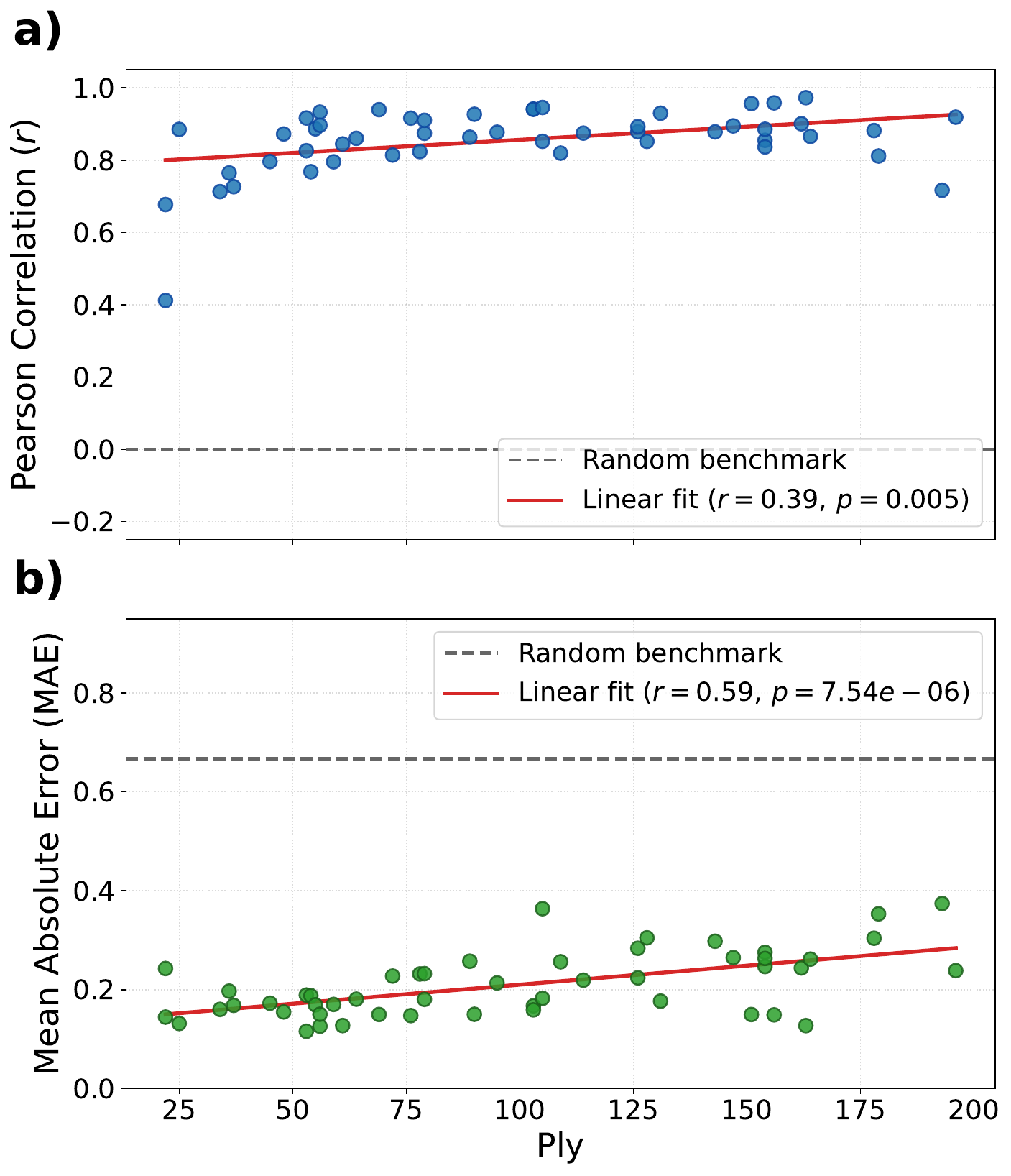}
\caption{\label{fig:scatter_robustness} \textbf{Cross-engine validation and robustness of spatial ownership predictions.} 
Evaluated on empty intersections across 50 game--ply configurations throughout game progression. 
\textbf{(a)} Pearson correlation coefficient $r$ compared against the stochastic null baseline ($\mathbb{E}[r] \approx 0$, gray dashed line). 
\textbf{(b)} Mean absolute error ($\mathrm{MAE}$) measured against the theoretical continuous uniform baseline $\mathcal{U}(-1, 1)$ ($\mathbb{E}[\mathrm{MAE}] \approx 0.667$, gray dashed line). 
Solid red lines denote linear fits.}
\end{figure}

As shown in Fig.~\ref{fig:scatter_robustness}, the agreement decisively rejects the null model $\mathcal{U}(-1, 1)$. The Pearson correlation mostly spans $r \in [0.70, 0.98]$ across the game progression, with the $\mathrm{MAE}$ remaining mainly bounded between $0.15$ and $0.35$. The positive slope in Fig.~\ref{fig:scatter_robustness}a ($p = 0.005$) captures the progressive stabilization of living groups and borders. Conversely, the slight increase in late-game $\mathrm{MAE}$ (Fig.~\ref{fig:scatter_robustness}b, $p = 7.54 \times 10^{-6}$) is a natural consequence of probability saturation: as values polarize into binary extremes ($\pm 1.0$), localized tactical discrepancies over boundary stones produce discrete steps $|\Delta O| \approx 2.0$, moderately elevating the average error without disrupting global topological alignment.

\begin{figure}[!htbp]
\centering
\includegraphics[width=1\linewidth]{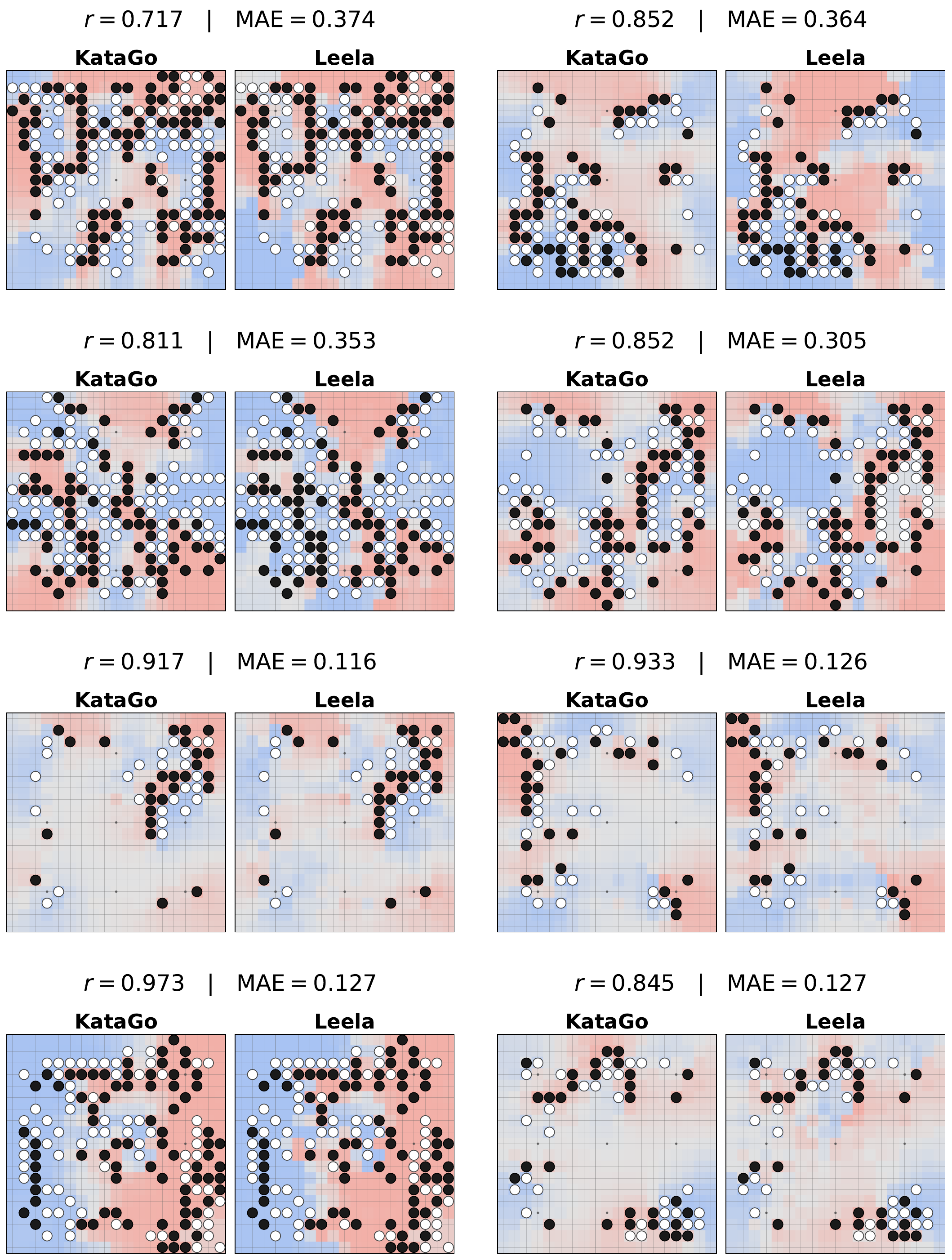}
\caption{\label{fig:extreme_mae_cases} \textbf{Extreme discrepancy cases in cross-engine territorial prediction.} 
Selected from 50 sampled game--ply configurations. The top four panels display instances with the highest mean absolute error ($\mathrm{MAE} \in [0.305, 0.374]$, $r \in [0.717, 0.852]$), whereas the bottom four panels display positions with the lowest error ($\mathrm{MAE} \in [0.116, 0.127]$, $r \in [0.845, 0.973]$).}
\end{figure}

A visual inspection of the extreme cases (Fig.~\ref{fig:extreme_mae_cases}) shows that even in the positions with the largest discrepancy ($\mathrm{MAE} \approx 0.37$), the two engines give broadly consistent assessments of which groups are alive and where the main territorial boundaries lie. In the lowest-discrepancy regimes ($\mathrm{MAE} \approx 0.11\text{--}0.13$), the spatial agreement is nearly indistinguishable.

\section{Spectral radius as a macroscopic measure of strategic tension}
\label{sec:spectral_radius_tension}
\noindent
The theoretical justification for adopting the spectral radius $\lambda_1$ as a macroscopic measure of strategic tension builds upon the dynamical systems and topological entropy framework established in our previous study on chess \cite{cerioli2025ai}. In that formulation, the complexity of a game state is governed by the asymptotic growth rate of walks across the interaction graph, where the topological entropy $h_{\mathrm{top}} = \lim_{n \to \infty} \frac{1}{n} \log N(n) = \log(\lambda_1)$~\cite{adler1965topological, dinaburg1970correlation, bowen1971entropy} is strictly determined by the dominant eigenvalue of the adjacency matrix (see~\cite{cerioli2025ai} for the complete spectral derivation).

This mathematical foundation generalizes naturally to the game of Go, despite a fundamental structural difference in how interactions are modeled: whereas chess networks rely on discrete, combinatorial attack-and-defense links, spatial dependencies in Go are mediated by continuous-valued coupling intensities defined via spatial proximity and territorial influence. Furthermore, rather than operating on a fixed set of piece or square indices, the Go network is represented by a dynamically evolving coupling matrix $W \in \mathbb{R}_{\ge 0}^{M \times M}$, where the effective dimensionality $M = G + Z$ fluctuates as collective stone groups ($G$) and contested territorial zones ($Z$) emerge, merge, or settle. Since $\lambda_1$ generally grows with the size of the network, a larger number of nodes could in principle raise tension on its own. We checked for this confounding factor: the average number of groups is very similar across ensembles and does not follow the tension hierarchy, so it cannot account for the observed differences. Tension therefore depends mainly on how strongly groups and zones are coupled, not simply on how many there are.

Because $W$ is real, symmetric, and non-negative, the Spectral Theorem and the Perron--Frobenius theorem guarantee that the total multi-step communicability across the board, $C(n) = \mathbf{1}^T W^n \mathbf{1}$, scales asymptotically with the leading eigenvalue $\lambda_1$:
\begin{equation}
\begin{aligned}
\lim_{n \to \infty} [C(n)]^{1/n} &= \lambda_1, \\
\text{or equivalently,} \quad h_{\mathrm{top}}(W) &= \lim_{n \to \infty} \frac{1}{n} \log C(n) = \log(\lambda_1).
\end{aligned}
\end{equation}
For an unweighted adjacency matrix, this quantity coincides with the topological entropy used in chess~\cite{cerioli2025ai}. For the weighted matrix $W$ used here, we refer to $\log\lambda_1$ as the growth rate of weighted walks, since rescaling all weights by a constant $c$ shifts it by $\log c$. The spectral radius summarizes the strength of coupling, but does not by itself determine its spatial extent.

In the context of Go, $\lambda_1$ thus acts as a macroscopic susceptibility index that captures the global interdependence of the board:
\begin{itemize}
    \item \textbf{Global Coupling Capacity:} Rather than merely tallying localized tactical skirmishes, $\lambda_1$ measures the asymptotic capacity of the spatial network to propagate multi-step dependency cascades. A large $\lambda_1$ indicates that the life-and-death status of stone groups and the territorial ownership of contested zones across distant sectors are deeply entangled into a single, highly sensitive equilibrium. Under this high-tension regime, a local perturbation (such as an invasion, attachment, or probing move, \textit{tesuji}) triggers structural reverberations across the entire board.
    \item \textbf{Structural Stabilization and Decoupling:} Conversely, a sharp decline in $\lambda_1$ marks the progressive partitioning of the board into mutually decoupled, self-contained subgraphs. In this low-tension regime, stone groups have secured definitive life-and-death status (\textit{seki} or two-eye stability) and contested zones have crystallized into settled territorial boundaries, preventing local endgame plays (\textit{yose}) from altering the global balance of the game.
\end{itemize}
Tracking $\lambda_1$ therefore provides a parameter for global strategic tension across the opening (\textit{fuseki}), middlegame (\textit{chuban}), and endgame (\textit{yose}) phases.

\section{Example of tension profile for a single game}
\label{subsec:example_tension}
\begin{figure}[!htbp]
\centering
\includegraphics[width=1\linewidth]{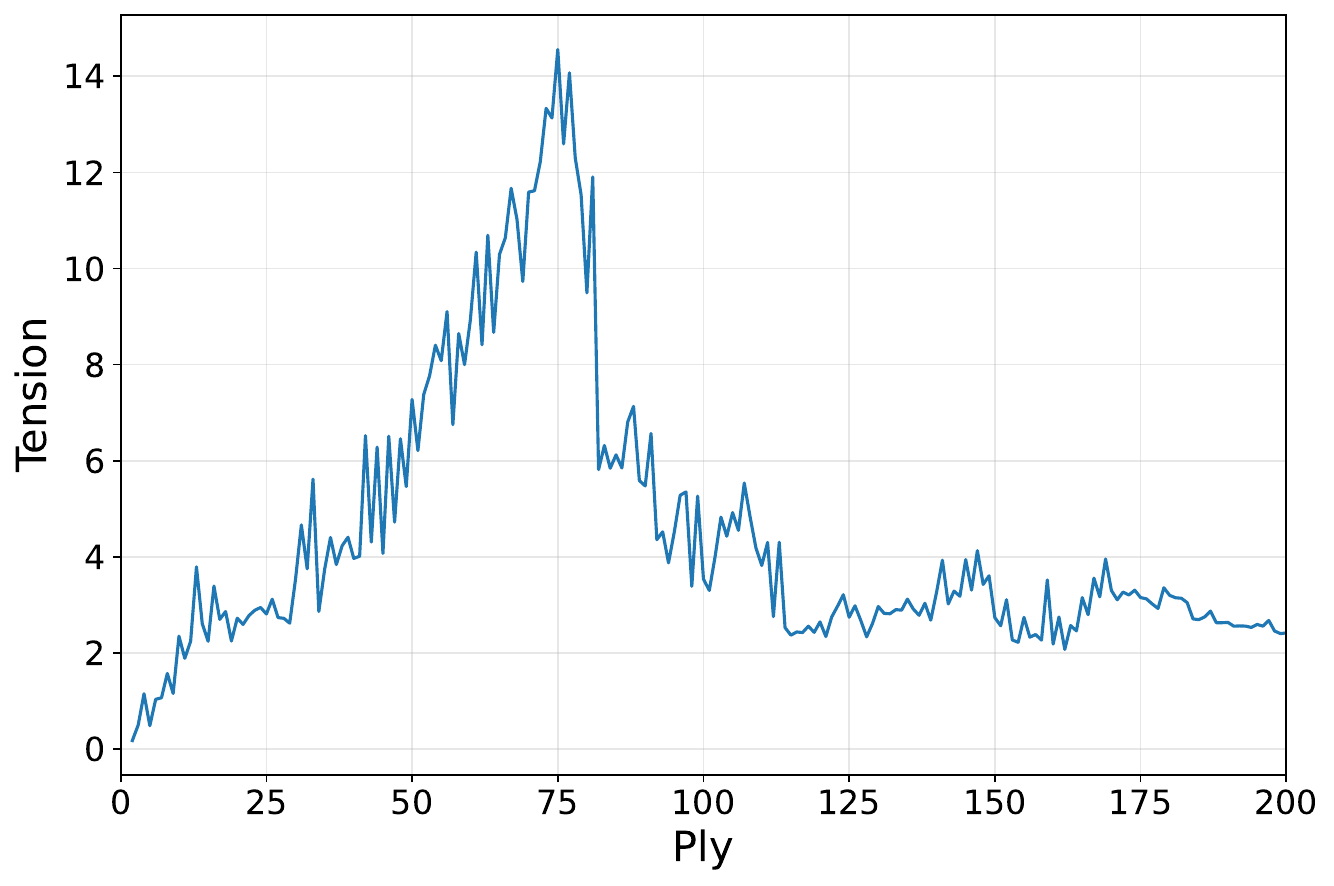}
\caption{\label{fig:single_game} \textbf{Move-by-move strategic tension profile in elite human--machine play.} 
Evolution of the network spectral radius $\lambda_1$ across game progression, evaluated via KataGo queries (100 base visits, 10 perturbation visits) for Game 2 of the historic challenge match between Lee Sedol and AlphaGo (Seoul, March 10, 2016).}
\end{figure}

\noindent
In this section, we present the tension profile of a single game, specifically Game 2 of the historic five-match challenge series between Lee Sedol and AlphaGo (Fig.~\ref{fig:single_game}), which concluded with a decisive victory for AlphaGo. Without the smoothing effect of averaging across a large dataset, pronounced macroscopic fluctuations become highly visible: individual stone placements can fundamentally reshape the spatial distribution of territory and influence, triggering sudden shifts in global tension. This behavior highlights the extreme sensitivity of our network-theoretic metric to local positional variations. Abrupt spikes and drops in the profile directly reflect critical strategic choices, such as the sudden invasion of a contested zone, the structural destabilization of a floating group, or the decisive resolution of a local life-and-death struggle.

The peak tension is observed at ply 75, with its interaction network illustrated in Fig.~\ref{fig:network}.

\begin{figure}[!htbp]
\centering
\includegraphics[width=1\linewidth]{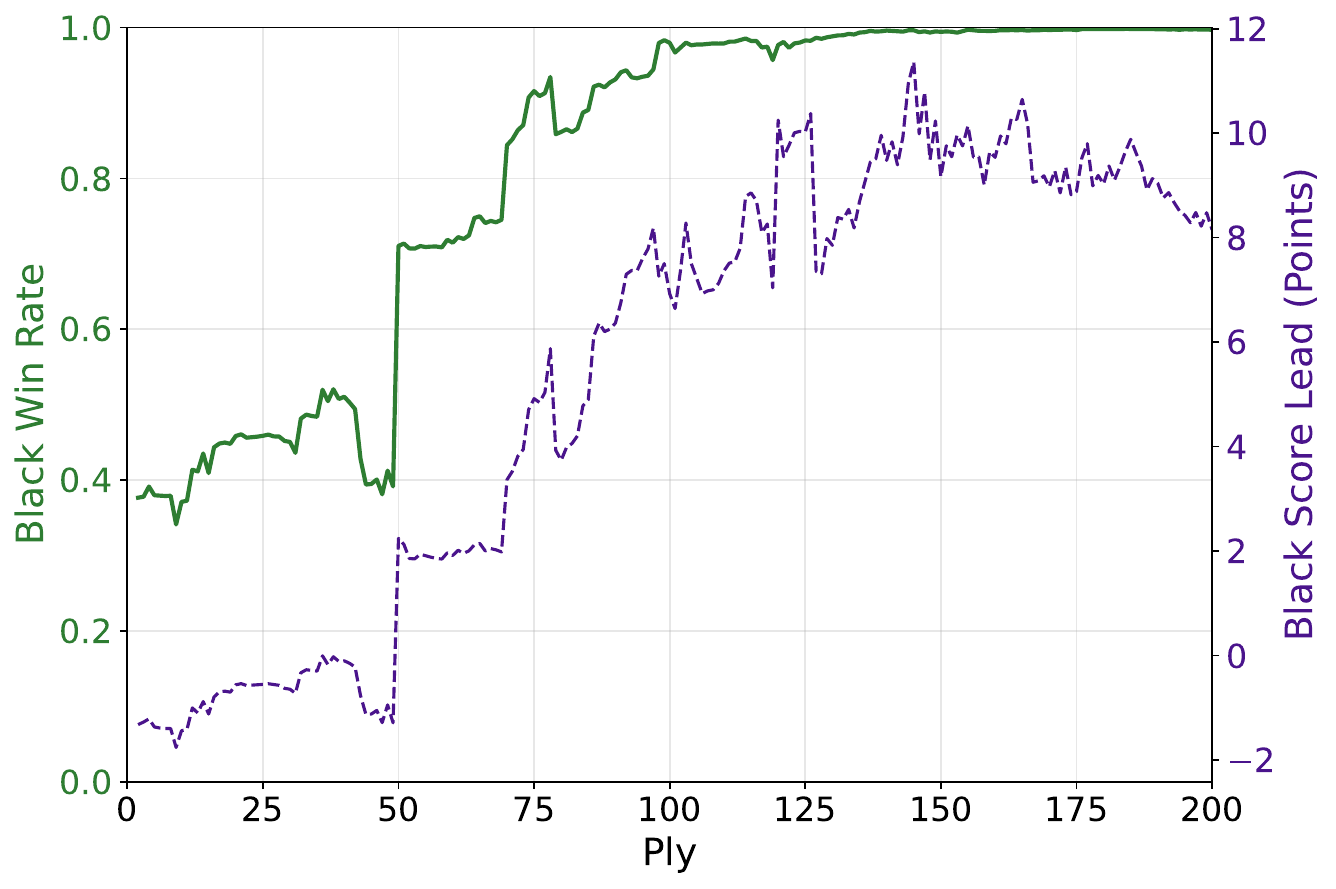}
\caption{\label{fig:evaluation_game} \textbf{Objective evaluation trajectory across move progression.} 
KataGo evaluation profile across plies for Game 2 of \textit{Lee Sedol vs. AlphaGo} (matching Fig.~\ref{fig:single_game}), computed at 500 analysis visits per position. Black win rate (left axis) and predicted score lead in points (right axis) track the shifting territorial balance throughout the match.}
\end{figure}

To provide a deeper strategic context, we also analyze the game's objective engine evaluation (Fig.~\ref{fig:evaluation_game}), computed using KataGo. Specifically, the engine's output is characterized by two distinct complementary metrics: the \textit{Black Win Rate} (WR), which quantifies AlphaGo's objective probability of winning the match in a range of $[0, 1]$, and the \textit{Black Score Lead}, which estimates the expected net point advantage on the board from Black's perspective. As shown in the evaluation profile, the game starts with a standard compressed win rate for Black ($\approx 40\%$) due to the strict $7.5$-point Komi rule, which translates to a slight virtual score disadvantage under KataGo's modern metric \cite{wu2019accelerating}. 

The most pronounced single-move spike in advantage for AlphaGo occurs exactly at ply 50, where Black's win rate takes its sharpest vertical leap, permanently locking in a decisive strategic dominance. Interestingly, around ply 75, while the win rate has already smoothly plateaued toward an imminent victory, the structural tension reaches its absolute global maximum. This clear divergence suggests that our tension metric does not merely mirror the game's outcome or win probability; rather, it captures the genuine topological complexity and structural vulnerability of the board configuration, suggesting to be fundamentally independent of who is leading and by how much.

\begin{figure}[!htbp]
\centering
\includegraphics[width=1\linewidth]{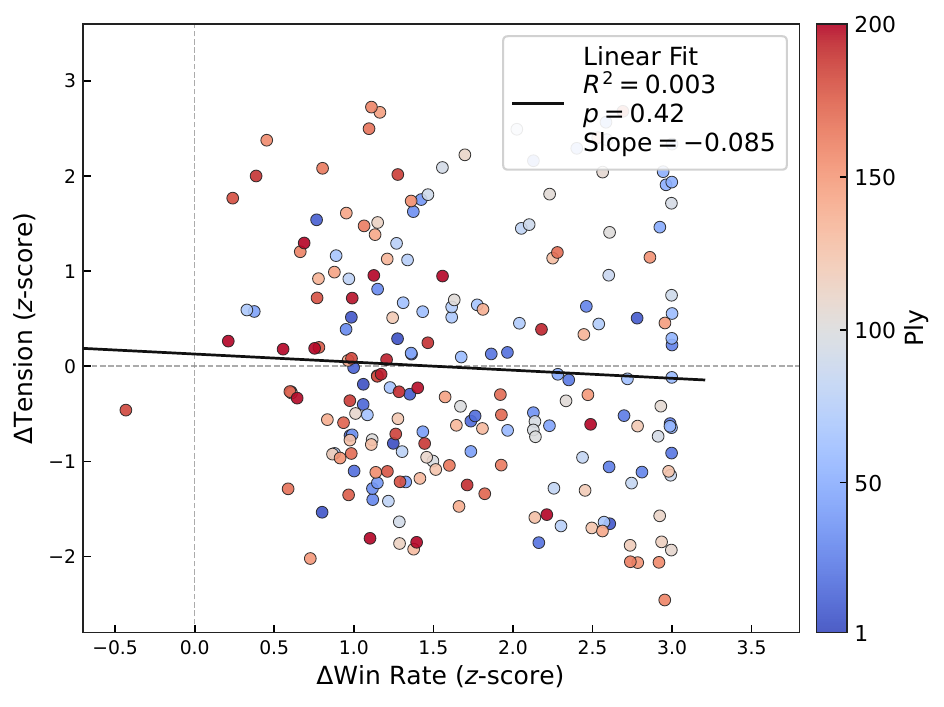}
\caption{\label{fig:decision_landscape} \textbf{Decoupling of win-rate optimization and strategic tension dynamics.} 
Standardized move-to-move change in active player win rate ($\Delta\mathrm{WR}$) plotted against standardized change in strategic tension ($\Delta\mathrm{Tension}$) for Game 2 of \textit{Lee Sedol vs. AlphaGo} (matching Fig.~\ref{fig:single_game}), with points colored by ply progression. At each ply, values are standardized as $z$-scores relative to a uniform sample of 10 legal candidate moves (5 evaluation visits each), evaluated against a base state of 20 visits and 5 internal perturbations (2 visits each) for tension estimation. The solid line denotes a linear regression fit ($R^2 = 0.003$, $p = 0.42$, slope $= -0.085$).}
\end{figure}

To quantitatively substantiate the orthogonality between structural tension and engine evaluation dynamics, we examine the move-to-move variations in both quantities across the game. Fig.~\ref{fig:decision_landscape} displays the relationship between standardized changes in win rate ($\Delta\text{WR}$) and standardized changes in tension ($\Delta\text{Tension}$), color-coded by ply index. A linear regression yields a vanishing coefficient of determination with no statistical significance. This statistical independence demonstrates that abrupt tactical turning points swinging the win probability do not necessarily dictate proportional reconfigurations in network tension, and vice versa—reinforcing that our network metric captures topological board complexity rather than mere engine evaluation fluctuations. 

\section{Ensemble analysis of tension dynamics and win-rate decoupling}
\label{subsec:ensemble_tension_analysis}
\noindent
To determine whether the temporal decoupling observed in Game~2 between Lee Sedol and AlphaGo (Appendix~\ref{subsec:example_tension}) is a general strategic property, we evaluate an ensemble of $N = 70$ games between AlphaGo and professional players. We quantify this relationship through two metrics: (i) the macroscopic \textit{lead-lag} offset, defined as the temporal delay $\Delta\tau = \mathrm{Ply}(T_{\max}) - \mathrm{Ply}(\max|\Delta\mathrm{WR}|)$ between the sharpest win-rate swing and the peak of global strategic tension $\lambda_1$, and (ii) the microscopic linear coupling between move-by-move tension variations $z(\Delta T)$ and win-rate shifts $z(\Delta\mathrm{WR})$, both standardized as $z$-scores relative to the ensemble of legal candidate moves available at each position.

\begin{figure}[!htbp]
\centering
\includegraphics[width=1\linewidth]{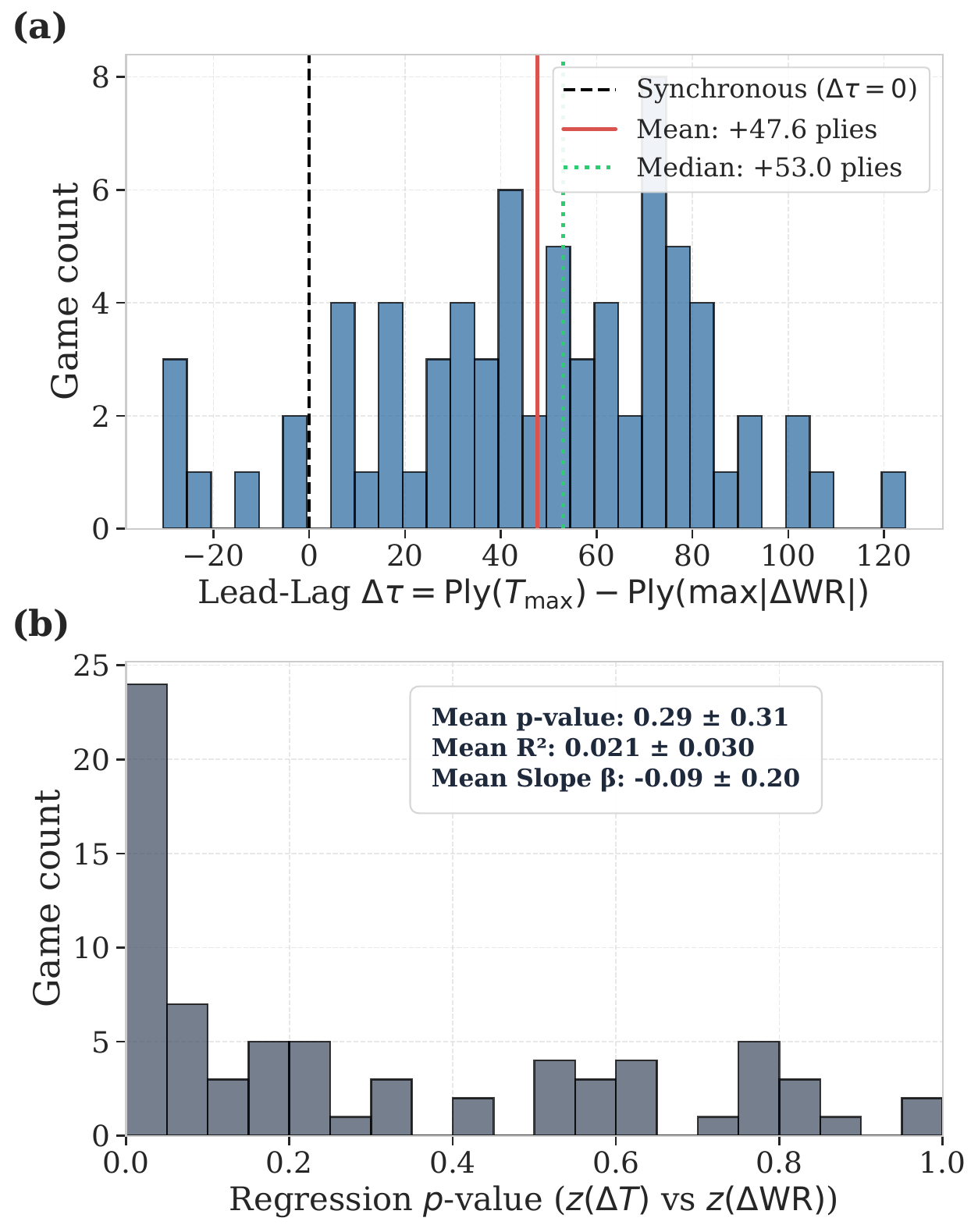}
\caption{\label{fig:ensemble_lead_lag} \textbf{Temporal lead-lag dynamics and microscopic decoupling in human--AI play.} 
Statistical ensemble analysis across $N = 70$ games between AlphaGo and human professionals. 
\textbf{(a)} Distribution of the lead-lag offset $\Delta\tau = \mathrm{ply}(T_{\max}) - \mathrm{ply}(\max|\Delta\mathrm{WR}|)$, where the dashed vertical line marks synchronous alignment ($\Delta\tau = 0$), while solid and dotted lines denote the sample mean ($+47.6$ plies) and median ($+53.0$ plies), respectively. 
\textbf{(b)} Distribution of linear regression $p$-values evaluating the microscopic coupling between $z(\Delta T)$ and $z(\Delta\mathrm{WR})$ across plies. Inset statistics report the mean $\pm$ standard deviation for the $p$-value, coefficient of determination $R^2$, and regression slope $\beta$. At each position, values are standardized as $z$-scores relative to a uniform sample of 10 legal candidate moves (5 evaluation visits each), evaluated against a base state of 20 visits and 5 internal perturbations (2 visits each) for tension estimation.}
\end{figure}

As shown in Fig.~\ref{fig:ensemble_lead_lag}a, the lead-lag offset
\begin{equation}
\Delta\tau = \mathrm{Ply}(T_{\max}) - \mathrm{Ply}(\max|\Delta\mathrm{WR}|)
\end{equation}
is mostly positive (mean $+47.6$ plies, median $+53$ plies): the largest win-rate swing usually occurs well before the tension peak. This is consistent with a separation of timescales, in which decisive shifts in evaluation tend to happen in the opening or early middlegame, while tension keeps building over the following plies as groups become entangled. Part of this offset, however, may simply reflect the fact that tension peaks at a similar stage in most games.

At the move level, the coupling between tension changes and win-rate changes is weak. Linear regressions of $z(\Delta T)$ on $z(\Delta\mathrm{WR})$ give a mean coefficient of determination of only $\langle R^2 \rangle = 0.021 \pm 0.030$ (Fig.~\ref{fig:ensemble_lead_lag}b). The coupling is nonetheless detectable in a subset of games: about one third of them have $p < 0.05$, far more than the $\sim$5\% expected under independence. The average slope is slightly negative ($\langle \beta \rangle = -0.094 \pm 0.196$), suggesting that moves that improve the mover's win rate tend, if anything, to reduce tension, consistent with a tendency toward simplification. Overall, tension and win rate carry largely different information: $\lambda_1$ captures aspects of the position that are only weakly related to who is winning.

\section{Robustness analysis on the number of KataGo visits and perturbations}
\label{subsec:appendix_robustness}
\noindent
To check that the tension metric $\lambda_1$ is not overly sensitive to the numerical choices made in its computation, we evaluate how the tension profile changes when varying three parameters: the engine's search depth (number of visits), the number of perturbations used to estimate the ownership variation, and the way perturbation moves are selected (random versus engine-preferred). All tests are performed on the same set of KataGo self-play games.

\subsection{Sensitivity to the number of KataGo visits}
\label{subsec:robustness_visits}
\noindent
The network construction relies on KataGo queries to generate the underlying ownership maps $O_{x,y}$ and simulate positional perturbations. To assess whether Monte Carlo Tree Search (MCTS) stochasticity at lower search depths distorts the structural topology, we evaluate the tension profiles across four distinct visit regimes: $20/2$, $50/5$, $100/10$, and $200/20$, maintaining a constant $10\%$ ratio between base evaluation visits ($V_{\text{base}}$) and perturbation visits ($V_{\text{perturb}}$).

\begin{figure}[!htbp]
\centering
\includegraphics[width=1\linewidth]{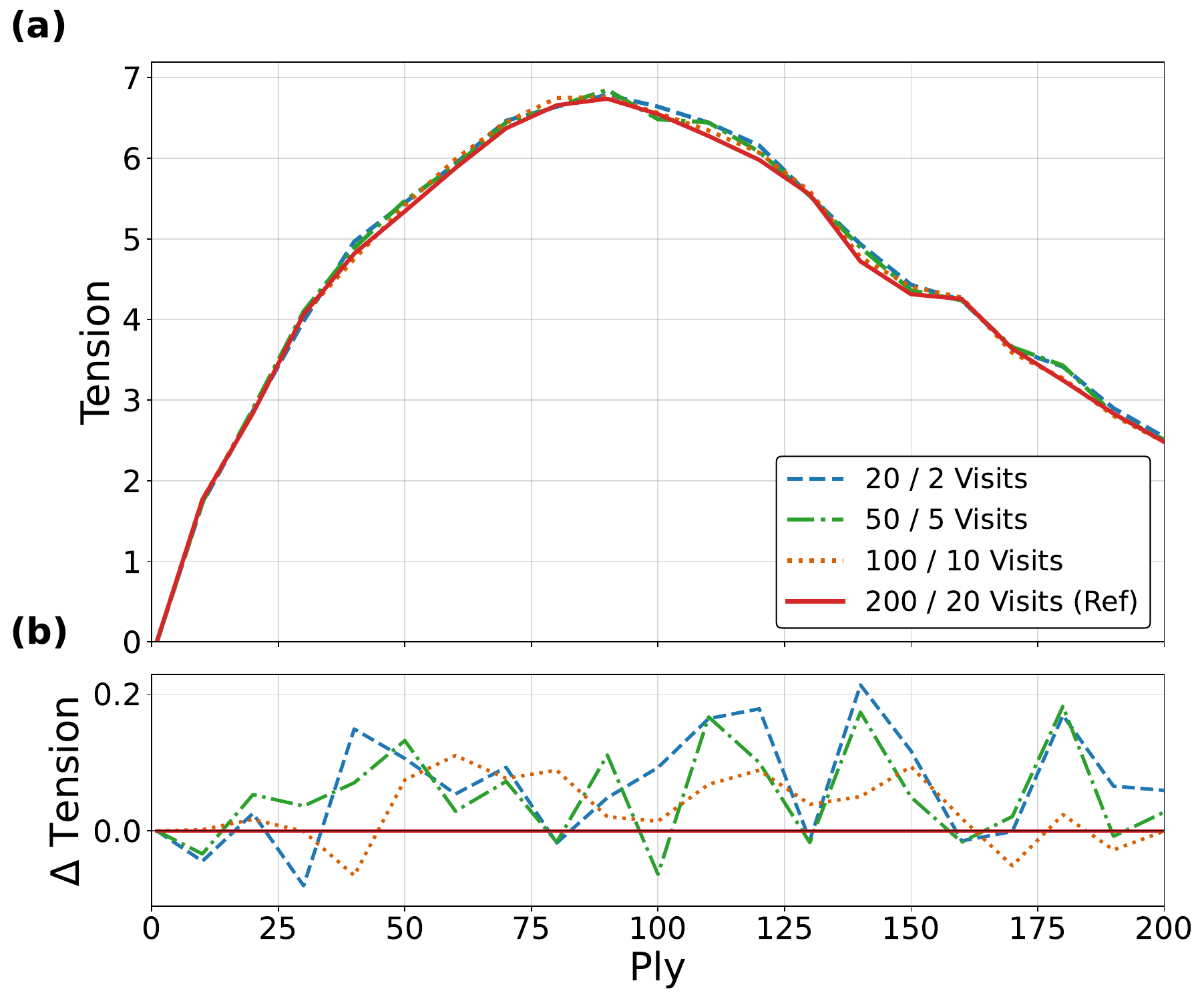}
\caption{\label{fig:robustness_visits} \textbf{Robustness of strategic tension across evaluation and perturbation budgets.} 
Analysis across $N = 200$ KataGo self-play games played at 1,000 visits per ply, sampled at 10-ply intervals. Strategic tension $\lambda_1$ is evaluated across varying base and perturbation search depths ($V_{\mathrm{base}} / V_{\mathrm{perturb}} \in \{20/2, 50/5, 100/10, 200/20\}$). 
\textbf{(a)} Mean strategic tension $\lambda_1$ across game progression. 
\textbf{(b)} Residual differences ($\Delta\mathrm{Tension}$) computed relative to the reference baseline ($200/20$ visits).}
\end{figure}

As illustrated in Fig.~\ref{fig:robustness_visits}, the macroscopic tension profile exhibits remarkable stability across all evaluated depth regimes. At extremely low search depths ($20/2$ visits), mild stochastic variance is observed at local tension peaks due to slight fluctuations in raw ownership probabilities. For intermediate regimes (e.g., $100/10$ visits), this residual deviation from the reference baseline progressively narrows, rendering the profile even closer to the $200/20$ baseline. Consequently, the $100/10$ evaluation depth was selected for all primary analyses conducted in this study, as it provides an optimal trade-off between structural accuracy and computational efficiency.

\subsection{Sensitivity to perturbation sample size}
\label{subsec:robustness_perturbations}
\noindent
The susceptibility term $\Delta_i \Delta_j$ in Eq.~\ref{eq:link_weight} captures positional variance by subjecting the board state to stochastic stone placements. To quantify the statistical convergence of the covariance matrix, we evaluate the strategic tension metric across four perturbation sample sizes: $N_{\text{perturb}} \in \{1, 5, 20, 100\}$.

\begin{figure}[!htbp]
\centering
\includegraphics[width=1\linewidth]{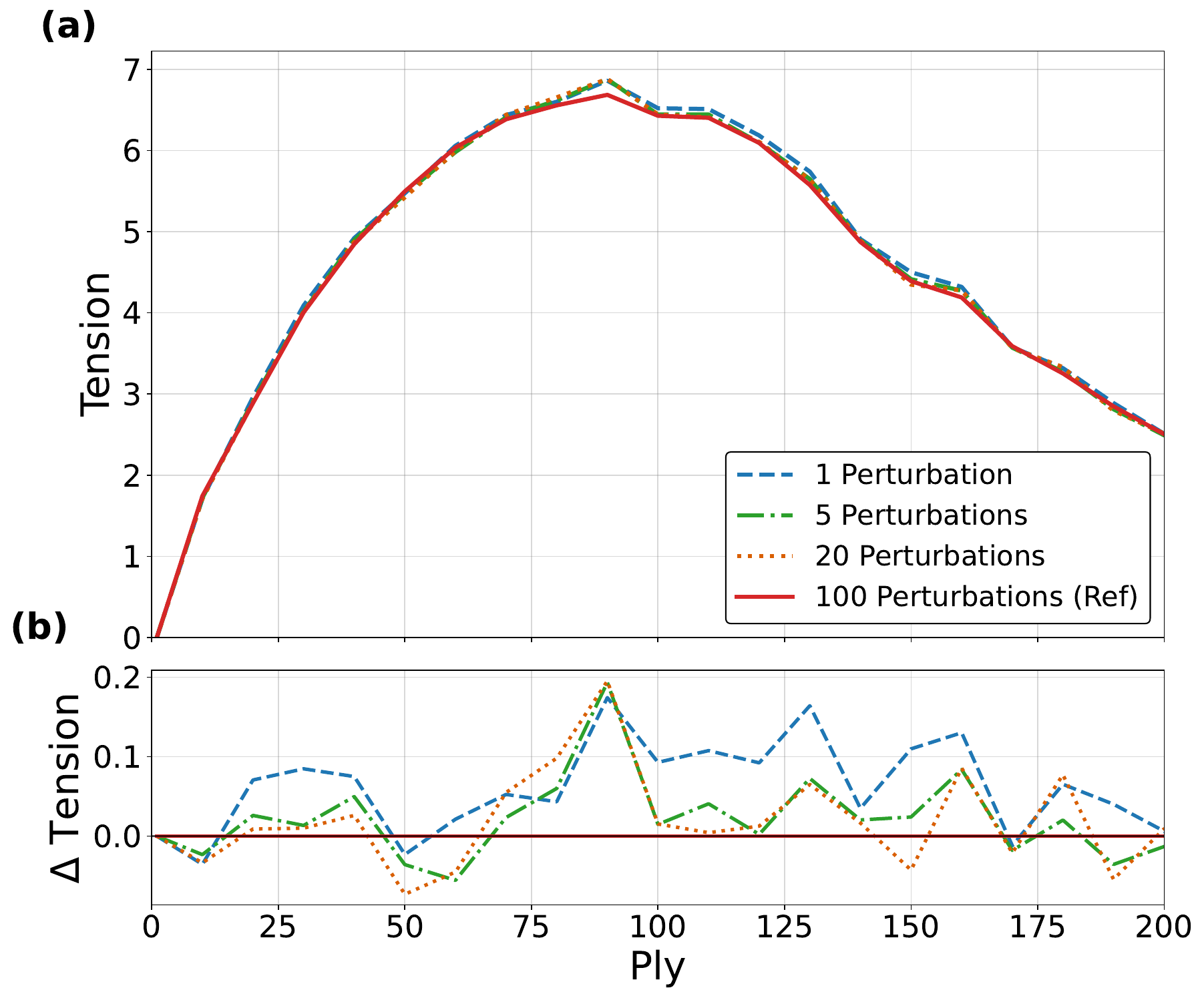}
\caption{\label{fig:robustness_perturbations} \textbf{Robustness of strategic tension to perturbation sample size.} 
Analysis across $N = 200$ KataGo self-play games sampled at 10-ply intervals, with fixed search parameters (100 base visits, 10 perturbation visits). Strategic tension $\lambda_1$ is evaluated across perturbation sample sizes ($N_{\mathrm{perturb}} \in \{1, 5, 20, 100\}$). 
\textbf{(a)} Mean strategic tension $\lambda_1$ across game progression. 
\textbf{(b)} Residual differences ($\Delta\mathrm{Tension}$) computed relative to the 100-perturbation reference baseline.}
\end{figure}

As illustrated in Fig.~\ref{fig:robustness_perturbations}, single-move perturbations ($N_{\text{perturb}} = 1$) introduce slightly more variance than larger perturbation samples, since the estimate then depends on the specific intersection chosen for the perturbation. Overall, however, the tension profile changes little with the number of perturbations: for all tested values, the residual differences with respect to the reference ($N_{\text{perturb}} = 100$) remain within about $0.2$, i.e., a few percent of the peak tension, and the shape and timing of the profile are unchanged. We use $N_{\text{perturb}} = 5$ in all primary analyses as a compromise between stability and computational cost.

\subsection{Sensitivity to perturbation strategic quality}\label{subsec:robustness_tactical_quality}
\begin{figure}[!htbp]
\centering
\includegraphics[width=1\linewidth]{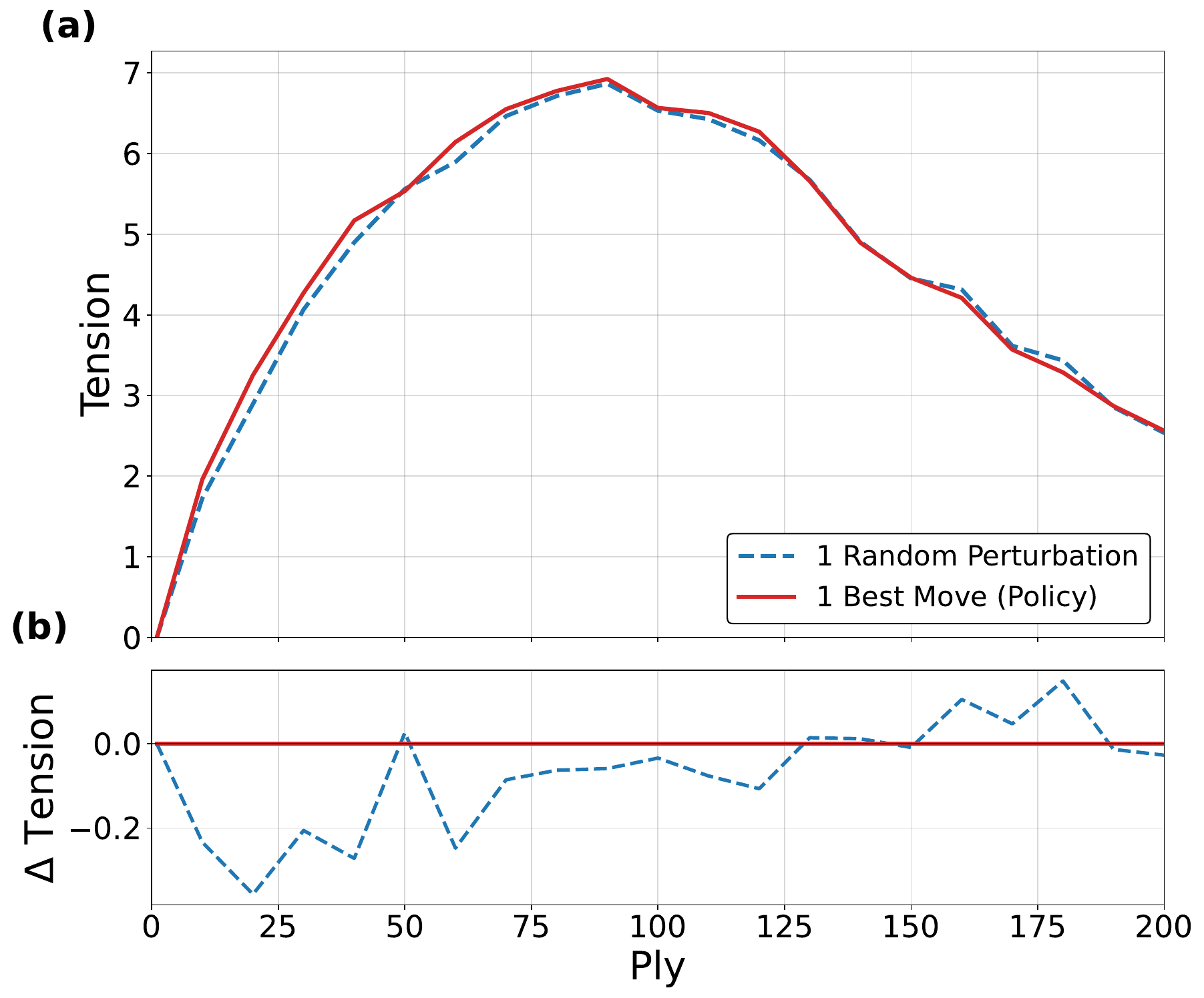}
\caption{\label{fig:robustness_mod_perturbations} \textbf{Robustness of strategic tension to perturbation selection policy.} 
Analysis across $N = 200$ KataGo self-play games sampled at 10-ply intervals, with fixed search parameters (100 base visits, 10 perturbation visits). Strategic tension $\lambda_1$ is compared between a single uniform random perturbation and a single policy-optimal perturbation (KataGo's top policy move). 
\textbf{(a)} Mean strategic tension $\lambda_1$ across game progression for both perturbation modes. 
\textbf{(b)} Residual differences ($\Delta\mathrm{Tension}$) of the random perturbation computed relative to the policy-optimal baseline.}
\end{figure}

\noindent
A critical question regarding the network construction is whether the tension metric merely measures the local threat generated by an optimal tactical move, or if it captures a true global property of the board's spatial topology. To isolate this effect, we evaluate the tension profile using a single random perturbation (sampled uniformly across all legal intersections) and compare it against a single optimal perturbation, where the stone is placed exactly at the coordinate with the highest probability assigned by KataGo's policy network. As illustrated in Fig.~\ref{fig:robustness_mod_perturbations}, the random and optimal perturbation methods yield similar macroscopic tension profiles. Despite the vast tactical disparity between a random placement and the engine's preferred move, both curves peak synchronously and exhibit the same structural evolution. Furthermore, the residual differences ($\Delta \text{Tension}$) between the random perturbation and the policy-optimal baseline are minimal, fluctuating within a narrow band of approximately $\pm 0.3$ throughout the game. This result demonstrates a fundamental property of the proposed framework: the macroscopic strategic tension is governed by the intrinsic geometric interdependencies and the inherent structural fragility of the board state, rendering it highly robust to the tactical quality of the applied perturbation.  

\section{Sensitivity to spatial locality scale}
\label{subsec:robustness_locality}
\noindent
The spatial interaction kernel $F(d_{ij}) = \exp(-d_{ij}/L_0)$ in Eq.~\ref{eq:link_weight} regulates the spatial range over which positional perturbations induce topological dependencies between board elements. The characteristic length scale $L_0 = \sqrt{m/\pi}$ acts as a spatial cut-off, balancing local tactical coupling against long-range strategic interactions. To verify how the tension profile is affected by the spatial cut-off parameter, we evaluate the tension metric across a range of multiplicative scale factors: $f \in \{0.25, 0.5, 1.0, 2.5, 5.0\}$, where the effective interaction range is set to $L = f \cdot L_0$.

\begin{figure}[!htbp]
\centering
\includegraphics[width=1\linewidth]{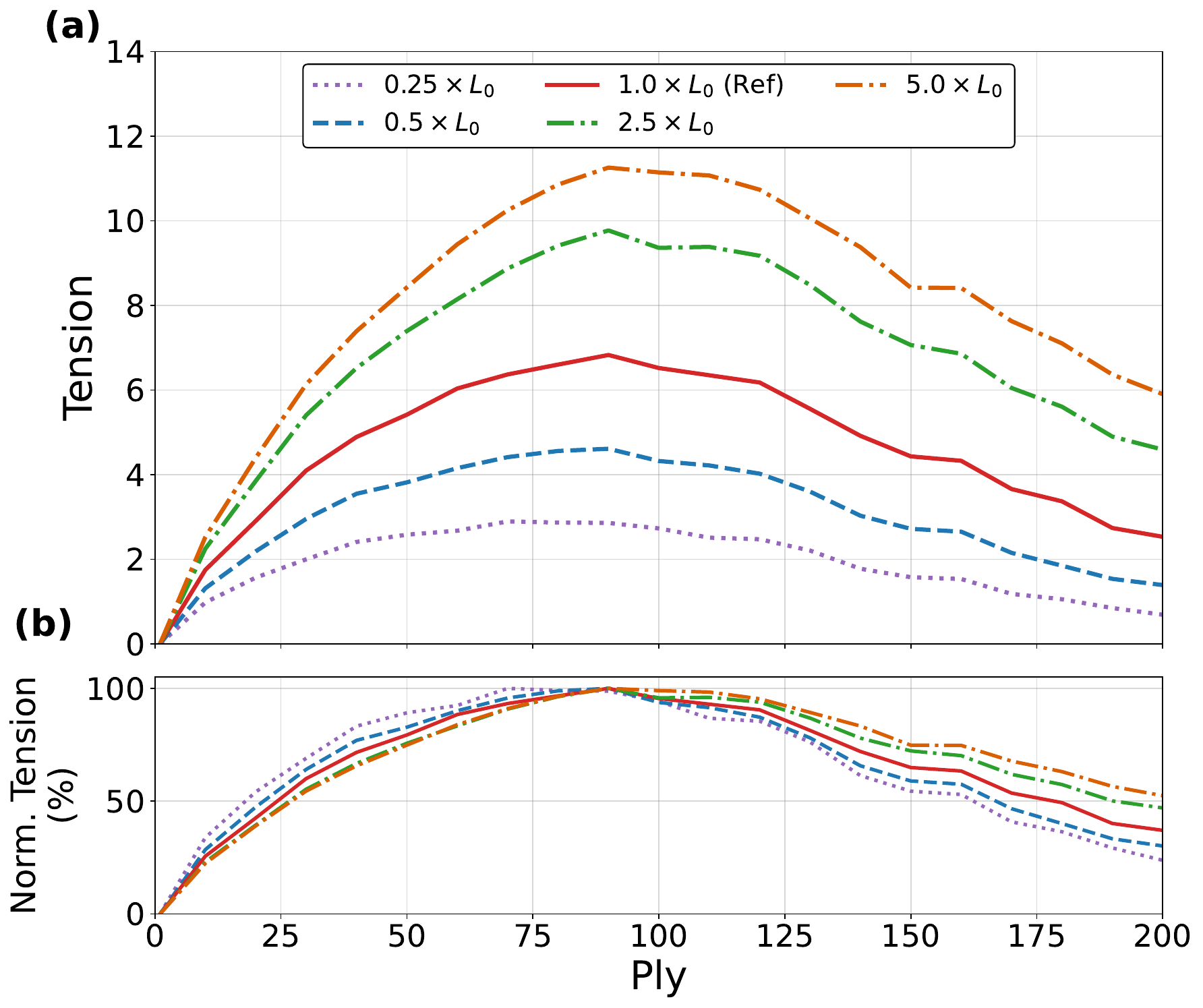}
\caption{\label{fig:robustness_locality} \textbf{Robustness of strategic tension to the spatial locality scale.} 
Analysis across $N = 200$ KataGo self-play games sampled at 10-ply intervals, with fixed search parameters (100 base visits, 10 perturbation visits). Strategic tension $\lambda_1$ is evaluated across spatial scale factors $f \in \{0.25, 0.5, 1.0, 2.5, 5.0\}$, where the interaction length scale is modulated as $L = f \cdot L_0$. 
\textbf{(a)} Mean strategic tension $\lambda_1$ across game progression. 
\textbf{(b)} Normalized tension profiles ($0$--$100\%$), highlighting the structural invariance of the dynamical trajectory and peak timing across spatial scales.}
\end{figure}

As illustrated in Fig.~\ref{fig:robustness_locality}a, varying the spatial locality scale naturally affects the absolute amplitude of the raw tension metric $\lambda_1$. Larger scale factors ($f > 1.0$) relax spatial attenuation, enabling distant groups to contribute additively to the dependency matrix $W$, thereby elevating the overall network density and spectral radius. Conversely, constrained spatial scales ($f < 1.0$) restrict interactions purely to immediate geometric neighbors, dampening the absolute amplitude.

As shown in Fig.~\ref{fig:robustness_locality}b, normalizing each profile to its maximum shows that the overall shape is largely preserved. The position of the tension peak ($\sim$ply 90) and the growth phase in the opening and middlegame are nearly identical across all scales. The endgame decay is more sensitive to the interaction range: with larger scale factors, tension decreases more slowly, since distant groups and zones remain coupled for longer. The choice of $L_0$ therefore affects the overall amplitude and the rate of endgame decay, but not the timing of the peak or the growth of tension during the game.

\section{Sensitivity to contested zone segmentation thresholds}
\label{subsec:robustness_zones}
\noindent
While stone groups are defined by deterministic rules, contested zones represent continuous areas of spatial uncertainty. To formalize these domains, our framework relies on uncertainty filtering bounds ($\theta_{\text{lower}}, \theta_{\text{upper}}$), a spatial clustering merge distance ($\theta_{\text{merge}}$), and a minimum component size threshold ($\kappa_{\text{min}}$). To confirm that the macroscopic tension metric is structurally stable and not an artifact of artificial partitioning choices, we systematically perturb these parameters across three distinct segmentation regimes:
\begin{enumerate}
    \item \textbf{Permissive Regime:} Broad uncertainty bounds ($\theta_{\text{lower}} = 0.01, \theta_{\text{upper}} = 0.85$), loose merging distance ($\theta_{\text{merge}} = 0.25$), and a minimal size cut-off ($\kappa_{\text{min}} = 1$).
    \item \textbf{Default Regime (Reference):} Balanced baseline thresholds ($\theta_{\text{lower}} = 0.05, \theta_{\text{upper}} = 0.75, \theta_{\text{merge}} = 0.20, \kappa_{\text{min}} = 2$).
    \item \textbf{Strict Regime:} Restrictive uncertainty bounds ($\theta_{\text{lower}} = 0.10, \theta_{\text{upper}} = 0.65$), tight spatial merging ($\theta_{\text{merge}} = 0.15$), and an elevated component size requirement ($\kappa_{\text{min}} = 3$).
\end{enumerate}

\begin{figure}[!htbp]
\centering
\includegraphics[width=1\linewidth]{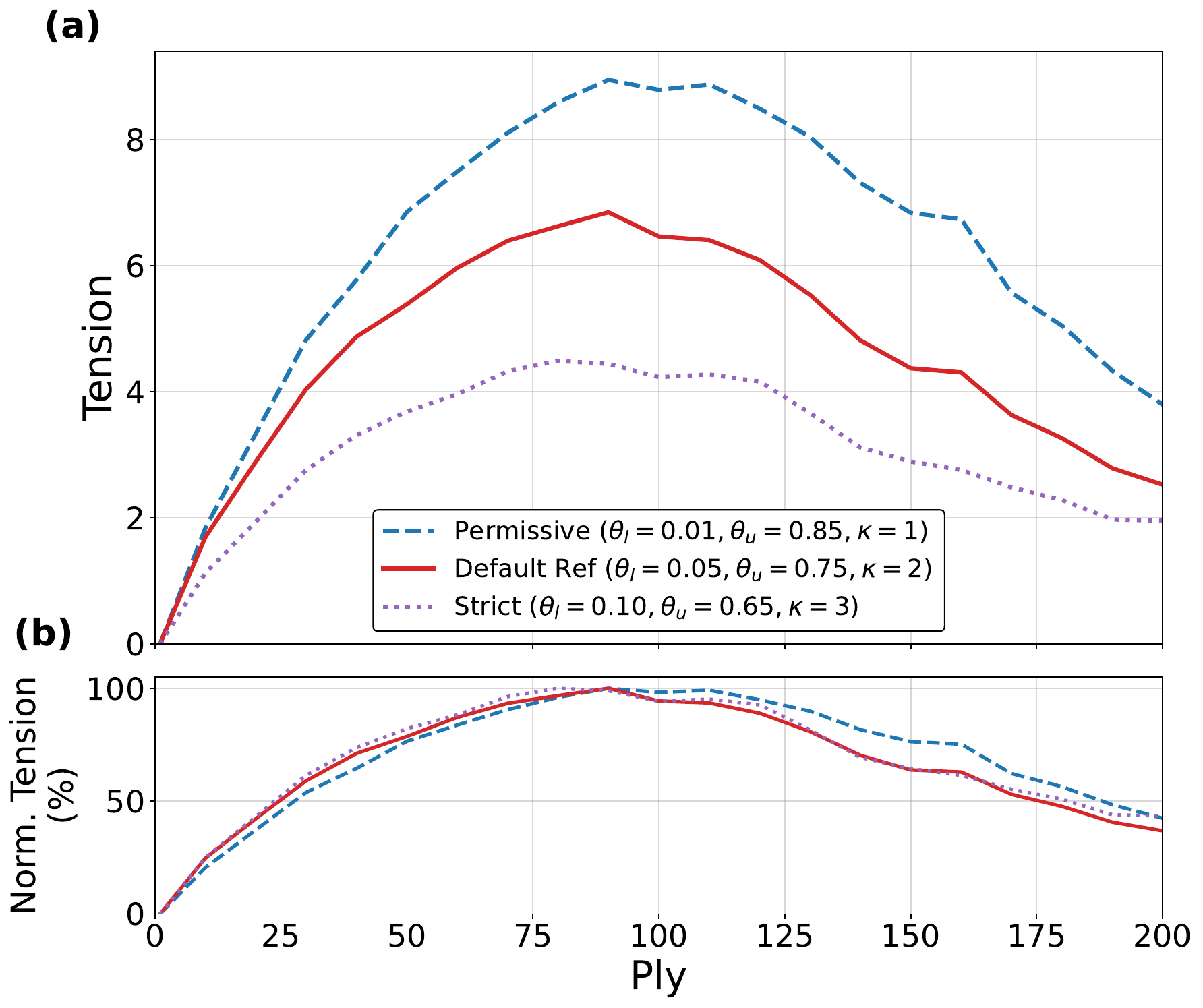}
\caption{\label{fig:robustness_zones} \textbf{Robustness of strategic tension to contested zone segmentation.} 
Analysis across $N = 200$ KataGo self-play games sampled at 10-ply intervals, with fixed search parameters (100 base visits, 10 perturbation visits). Strategic tension $\lambda_1$ is evaluated across three contested zone segmentation regimes: Permissive, Default, and Strict. 
\textbf{(a)} Raw mean strategic tension $\lambda_1$ across game progression. 
\textbf{(b)} Normalized tension profiles ($0$--$100\%$), demonstrating that the relative dynamical trajectory and peak timing remain robust to the choice of segmentation thresholds.}
\end{figure}

As illustrated in Fig.~\ref{fig:robustness_zones}(a), adopting a permissive segmentation scheme expands the total number and spatial extent of candidate zone nodes, resulting in a higher overall tension. Conversely, strict thresholding filters out small isolated pockets of uncertainty and weakly contested sectors, dampening the absolute tension amplitude.

As shown in Fig.~\ref{fig:robustness_zones}(b), normalizing profiles ($0$--$100\%$) yields a shared macroscopic tension behavior. Because phase transitions, growth slopes, and peak alignments remain consistent across regimes, the spatial zone decomposition is topologically robust, ensuring minor parameter variations do not distort the qualitative results.

\section{Classification and statistics of strategic tension drops}
\label{subsec:robustness_drops}
\noindent
To understand the structural reasons behind abrupt reductions in global strategic tension, we analyzed the statistical prevalence of the primary mechanisms causing sharp tension drops, defined as moves after which the spectral radius decreases by $\Delta \lambda_1 \le -1.0$. To ensure mutual exclusivity, each drop event is assigned to a single physical cause by evaluating a hierarchical set of topological conditions in strict sequential order (from top to bottom):
\begin{enumerate}
\item \textbf{Tenuki / New front:} The move is played far from the previous one ($d_{\text{prev}} \ge 8$) and either starts a new group ($\Delta N_{\text{groups}} > 0$) or captures stones ($N_{\text{captured}} > 0$), opening or reinforcing play in a different area of the board.
\item \textbf{Connection / Group merge:} Occurs when a move topologically fuses two or more same-color components into a single connected component ($\Delta N_{\text{groups}} < 0$), reducing overall graph fragmentation and suppressing local structural vulnerability.
\item \textbf{Tenuki / Distant extension:} The move is played far from the previous one ($d_{\text{prev}} \ge 8$) without capturing stones and without changing the number of groups ($\Delta N_{\text{groups}} = 0$), i.e., it extends an existing group in a different area of the board.
\item \textbf{Capture / Fighting resolution:} Tactical resolution of a local conflict resulting in the physical removal of enemy stones ($N_{\text{captured}} > 0$). This eliminates local spatial pressure, restores liberties to adjacent friendly groups, and settles territorial boundaries.
\item \textbf{Life / Eye formation:} Occurs when a move is associated with an increase in detected simple eyes
($\Delta \text{Eyes} > 0$), defined as empty intersections orthogonally
surrounded by stones of the same color, or with a substantial increase in
mean group liberties across the board
($\Delta L_{\text{mean}} \geq 2$), provided that no stones are captured
($N_{\text{captured}} = 0$).
\end{enumerate}

\begin{figure}[!htbp]
\centering
\includegraphics[width=1\linewidth]{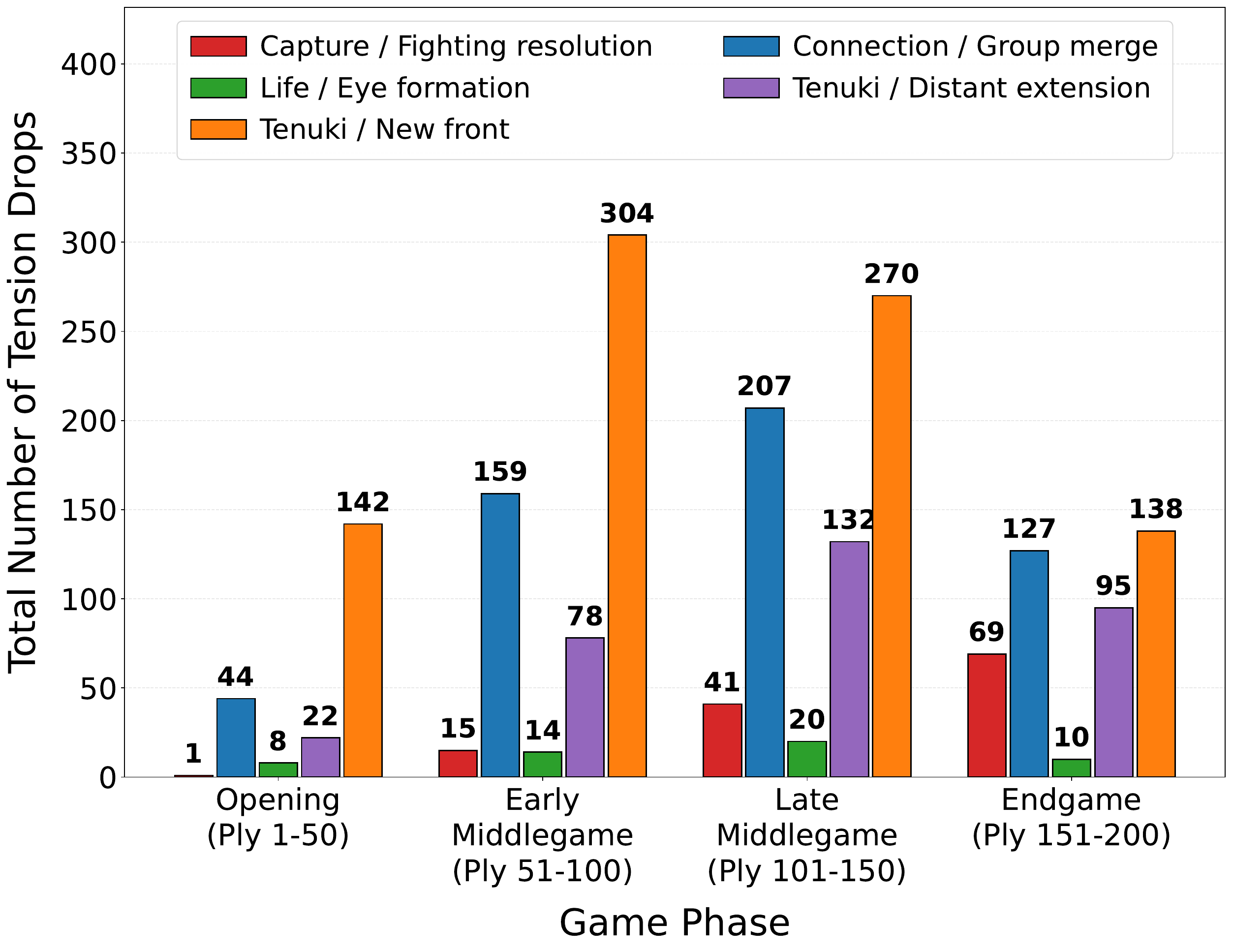}
\caption{\label{fig:tension_drop_causes} \textbf{Mechanisms driving strategic tension drops across game phases.} 
Frequency of sharp relaxation events ($\Delta\lambda_1 \le -1.0$) categorized across four developmental stages, evaluated in $N = 200$ KataGo self-play games (1,000 visits per ply). Events are classified into five physical mechanisms based on topological transition features: Capture/Fighting resolution (red), Connection/Group merge (blue), Life/Eye formation (green), Tenuki/Distant extension (purple), and Tenuki/New front (orange). Numbers above each bar indicate absolute event counts.}
\end{figure}

Evaluating these drop causes across the four canonical phases of play—Opening (plies 1--50), Early Middlegame (plies 51--100), Late Middlegame (plies 101--150), and Endgame (plies 151--200)—reveals distinct phase-dependent dynamical signatures (see Fig.~\ref{fig:tension_drop_causes}).

As shown in Fig.~\ref{fig:tension_drop_causes}, most sharp drops follow distant moves. \textit{Tenuki / New front} is the most frequent category in every phase, peaking in the early middlegame, and \textit{Tenuki / Distant extension} adds a further share in the middlegame. Large drops in tension thus typically occur when play shifts to a different area of the board, rather than through the local resolution of fights. \textit{Connection / Group merge} events also cluster in the middlegame, tracking the consolidation of isolated stones into larger groups. \textit{Capture / Fighting resolution} becomes more frequent as the game progresses and is most common in the endgame, when captures are more frequent overall. Finally, \textit{Life / Eye formation} remains rare in all phases, suggesting that groups usually secure their status through connections rather than through explicitly detected eyes.

\section{Decomposition of strategic tension}
\label{subsec:tension_decomposition}
\noindent
To dissect the structural origins of global strategic tension, we decompose the network interaction matrix into its two constituent topological channels: (i) direct tactical interactions between opposing stone groups $W_{\text{group--group}}$ (see Fig.~\ref{fig:network}a), and (ii) interactions coupling active stone groups to contested territorial zones $W_{\text{group--zone}}$ (see Fig.~\ref{fig:network}b). We evaluate the resulting spectral tension $\lambda_1$ independently for both channels across matches from elite human professionals (post-2016) and AI tournaments.

\begin{figure}[!htbp]
\centering
\includegraphics[width=1\linewidth]{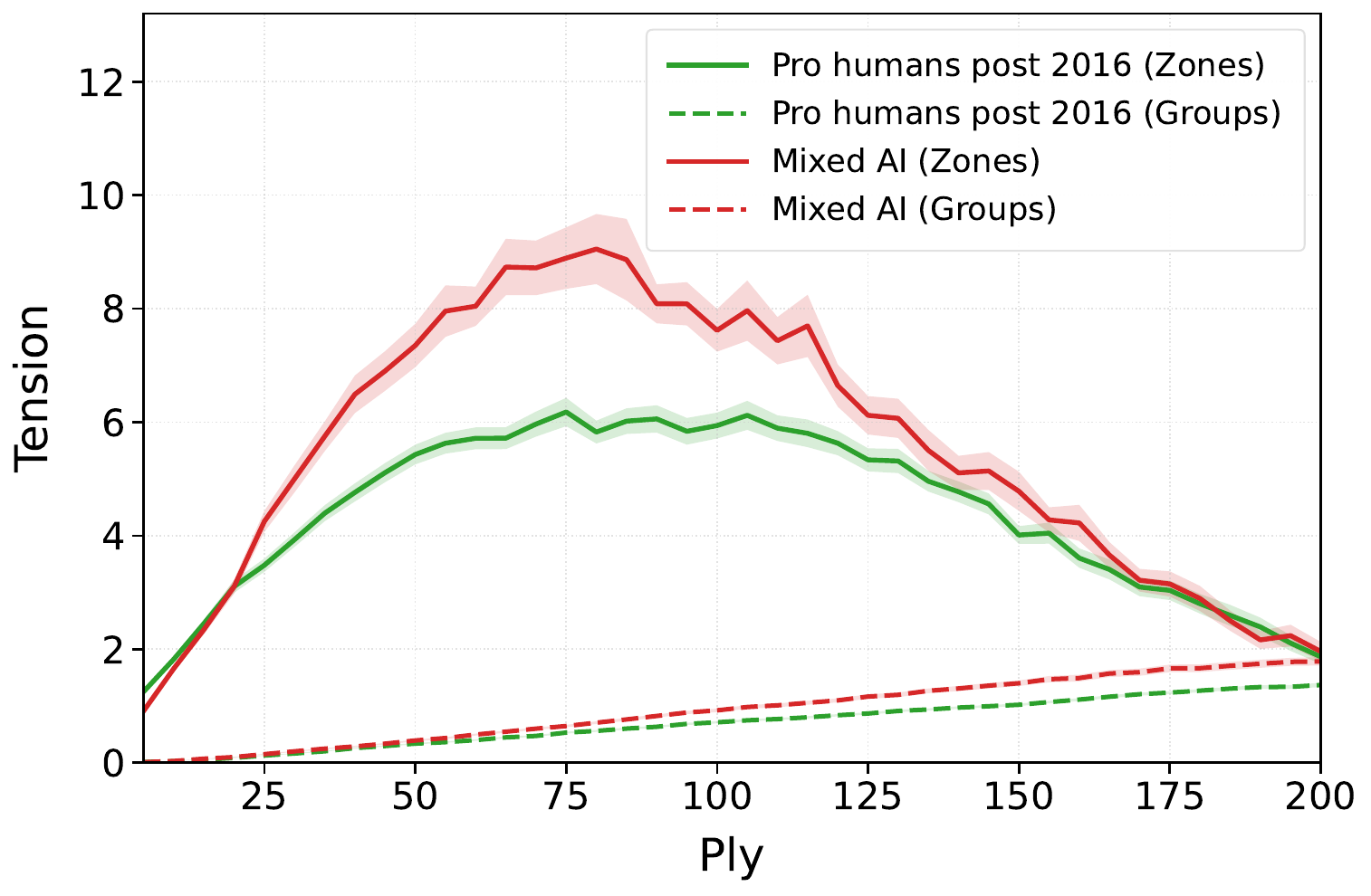}
\caption{\label{fig:tension_decomposition} \textbf{Channel-resolved decomposition of strategic tension.} 
Relative contributions of group--group and group--zone coupling across plies for post-2016 human professionals and Mixed AI ($N = 200$ games per cohort, matching Fig.~\ref{fig:tension_evolution}). Solid and dashed curves denote group--zone and direct group--group components, respectively. Shaded ribbons indicate the standard error of the mean.}
\end{figure}

As shown in Fig.~\ref{fig:tension_decomposition}, the overall tension is overwhelmingly dominated by the group--zone interaction channel. The tension mediated by contested zones exhibits a pronounced non-monotonic trajectory, surging during the opening and middlegame to reach peak values ($\lambda_1 \approx 6.2$ for humans and $\lambda_1 \approx 9.0$ for AI around plies 75--100) before decaying as territorial boundaries crystallize in the endgame. In contrast, the direct group--group tension remains substantially lower in magnitude throughout the game, growing monotonically from near zero to $\lambda_1 \approx 1.4$--$1.8$ as the board becomes crowded with groups toward ply 200.

Crucially, the strategic hierarchy between artificial intelligence and human play is preserved across both decoupled channels. AI players sustain higher tension than elite human professionals not only across contested territorial frontiers (solid curves) but also in direct group-to-group tactical entanglements (dashed curves) throughout the entire game trajectory. This demonstrates that the elevated strategic complexity exhibited by AI is an intrinsic macroscopic property across all structural interaction modes, rather than an artifact localized to territorial boundary estimation.

\section{Robustness to game length and competitive balance}
\label{subsec:robustness_competitive_balance}
\noindent
To ensure that the systematic divergence in strategic tension between artificial intelligence and elite human play is not an artifact of extrinsic match characteristics, we evaluate the robustness of our findings against two primary potential confounders: the temporal duration of games and the degree of competitive asymmetry between opponents. Differences in tactical resignation thresholds, endgame execution, and playing conventions could conceivably distort position-level statistics. We therefore characterize the empirical distributions of game length and final score margins, and subsequently assess strategic tension across identical competitive regimes using a rolling-window framework.

\begin{figure}[!htbp]
\centering
\includegraphics[width=1\linewidth]{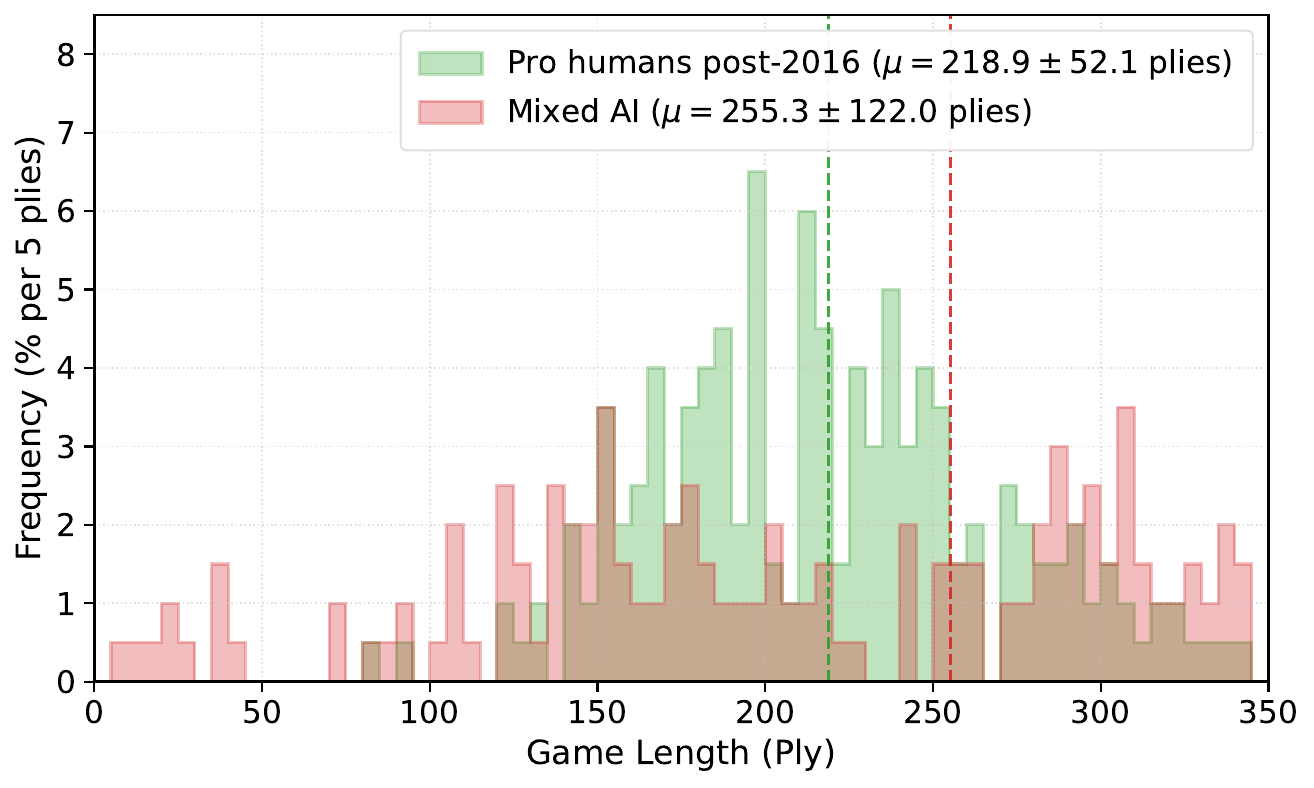}
\caption{\label{fig:game_lengths_distribution} \textbf{Game duration profiles across human and machine cohorts.} 
Empirical move counts grouped into 5-ply bins for post-2016 human professionals (green) and Mixed AI (red), from the ensembles analyzed in Fig.~\ref{fig:tension_evolution} ($N = 200$ games per cohort). Vertical dashed lines indicate the respective sample means.}
\end{figure}

The empirical game length distributions, displayed in Fig.~\ref{fig:game_lengths_distribution}, reveal distinct temporal profiles across the two populations. Human professional games exhibit a well-behaved unimodal distribution tightly centered around $\mu = 218.9 \pm 52.1$ plies, reflecting standardized pacing where middlegame tactical decisions typically mature into clear endgame conversions or professional resignations near ply 220. Conversely, Mixed AI matches display significantly greater dispersion, characterized by a broader mean and higher variance ($\mu = 255.3 \pm 122.0$ plies). Artificial agents frequently push tactical clashes beyond 300 plies when technical counting is contested, while occasionally producing abrupt early resignations under irreversible tactical refutations. Despite these distributional differences, both populations maintain dense statistical coverage throughout the critical ply $1$--$200$ window where macroscopic tension builds, peaks, and resolves.

\begin{figure}[!htbp]
\centering
\includegraphics[width=1\linewidth]{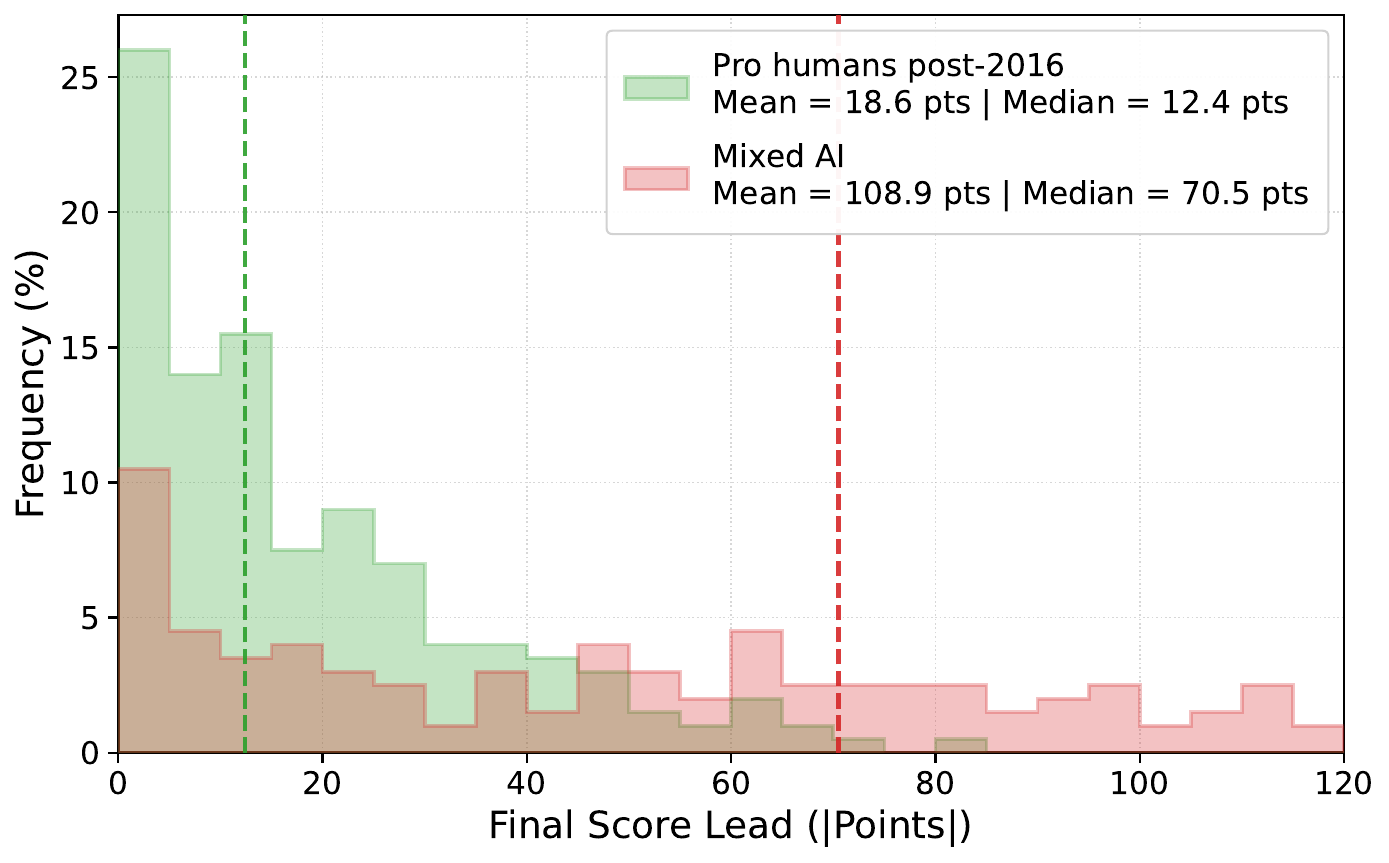}
\caption{\label{fig:score_lead_distribution} \textbf{Terminal score margins in human and machine play.} 
Relative frequency of the absolute final score lead $|S_{\mathrm{final}}|$ predicted by KataGo (500 visits per move) for post-2016 human professionals (green) and Mixed AI (red). Vertical dashed lines mark the respective sample medians.}
\end{figure}

A more subtle potential confounder stems from score margins. A conceivable concern is that elevated tension in Mixed AI encounters might be driven by runaway imbalances, where severe territorial swings artificially inflate spatial dependencies. To evaluate this hypothesis, we quantify the absolute score advantage at game termination, denoted by $|S_{\text{final}}|$ and evaluated via KataGo at 500 visits per ply. As shown in Fig.~\ref{fig:score_lead_distribution}, the competitive balance profiles of human and artificial play differ markedly. Human professionals predominantly conclude matches within close territorial margins, yielding a mean final lead of $18.6$ points and a median of only $12.4$ points, with an overwhelming concentration of games decided by fewer than 20 points. In contrast, Mixed AI encounters exhibit an expansive heavy tail, with an average score lead of $108.9$ points and a median of $70.5$ points. This pronounced divergence reflects human resignation etiquette: human professionals reliably resign once a deficit exceeds manageable territory, whereas artificial agents continue play until terminal conditions or threshold-based resignations are triggered, often resulting in catastrophic group deaths that generate nominal score margins exceeding 100 points.

\begin{figure}[!htbp]
\centering
\includegraphics[width=1\linewidth]{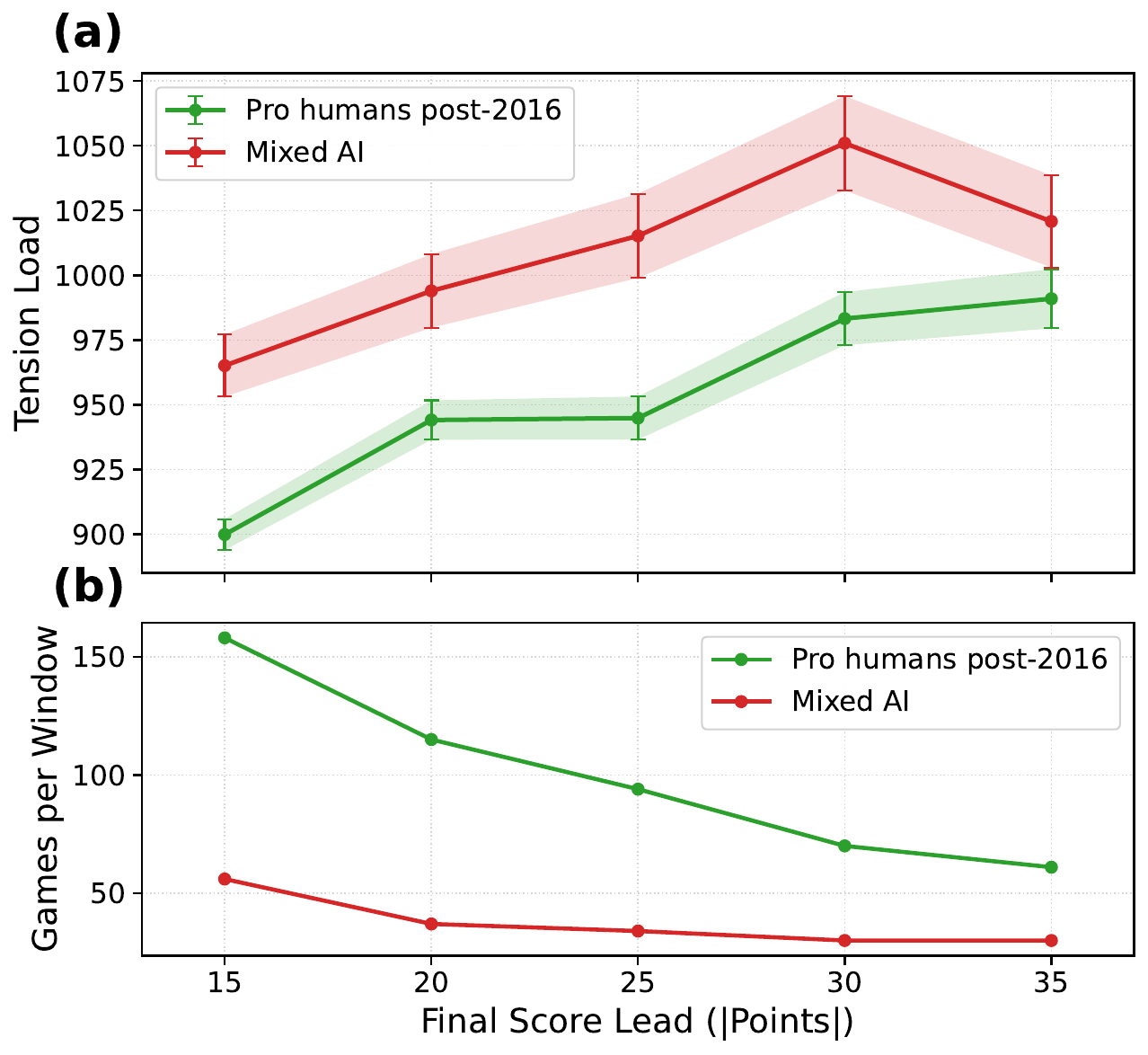}
\caption{\label{fig:tension_load_vs_score_lead} \textbf{Strategic tension load under controlled competitive balance.} 
Sliding-window evaluation across matches stratified by outcome margin. 
\textbf{(a)} Accumulated tension load plotted against the absolute final score lead $|S_{\mathrm{final}}|$, computed using an overlapping sliding window of width $\Delta S = 30$ points with a step size of 5 points across the common support $15 \le |S_{\mathrm{final}}| \le 35$ points. Shaded ribbons and error bars denote the standard error of the mean. 
\textbf{(b)} Number of contributing games per window across the shared margin domain.}
\end{figure}

Because the score lead distributions occupy substantially different supports, partitioning each dataset into internal tertiles would compare disparate physical regimes, matching narrow human territorial finishes against one-sided AI runaways. To establish an unbiased comparison, we restrict our analysis to the shared support of competitive margins and compute the cumulative tension load across plies $1$--$200$ as a continuous function of $|S_{\text{final}}|$ using an overlapping sliding window ($\Delta S = 30$ points, step size $5$ points). As demonstrated in Fig.~\ref{fig:tension_load_vs_score_lead}a, Mixed AI systematically sustains a higher tension load than elite human professionals across the entire shared competitive domain ($15 \le |S_{\text{final}}| \le 35$ points). While both ensembles display a moderate positive correlation between tension load and final margin, indicating that larger endgame swings often stem from more volatile middlegame conflicts, the Mixed AI trajectory remains uniformly shifted upward. Fig.~\ref{fig:tension_load_vs_score_lead}b confirms that both datasets retain robust statistical sample sizes across all evaluated windows. The higher tension of Mixed AI is therefore not explained by more one-sided outcomes, since it persists when comparing games with similar final score margins.

\section{Macroscopic board structure and game dynamics}
\label{sec:structural_analysis}
\noindent
Before examining micro-spatial move preferences, we analyze the macroscopic evolution of the game state through four fundamental structural metrics: the number of groups, mean group size, total captured stones, and mean liberties per group. This analysis is based on the same set of games considered in Fig.~\ref{fig:tension_evolution}.

\begin{figure}[!htbp]
\centering
\includegraphics[width=1\linewidth]{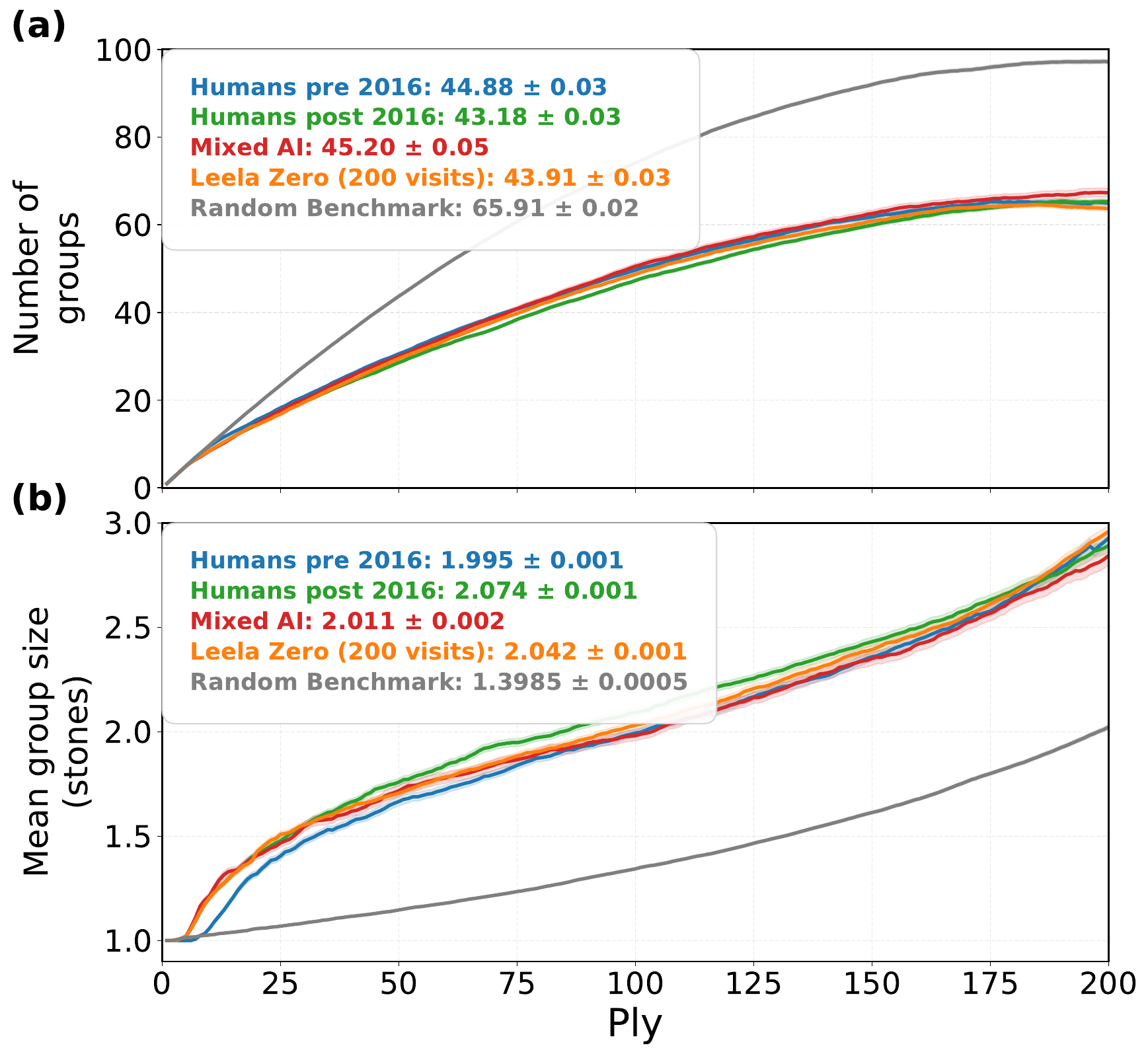}
\caption{\label{fig:structural_groups} \textbf{Macroscopic structure of connected stone groups across plies.} We consider the same set of games analyzed in Fig.~\ref{fig:tension_evolution}. (a) Mean number of distinct groups on the board. (b) Mean group size measured in total stones per connected component. The shaded bands represent the standard error of the mean.}
\end{figure}

Fig.~\ref{fig:structural_groups} illustrates the topological evolution of connected stone clusters (groups). Panel (a) shows that the number of distinct groups increases rapidly during the early stages of the game before gradually reaching a plateau. Mixed AI maintains a slightly higher average number of groups than human players. In contrast, Panel (b) shows that the mean group size increases monotonically as the game progresses and stones merge into larger connected structures. Post-2016 human players tend to form larger contiguous groups earlier in the game, whereas AI maintains a more distributed network topology, resembling the playing style of pre-2016 human players.

\begin{figure}[!htbp]
\centering
\includegraphics[width=1\linewidth]{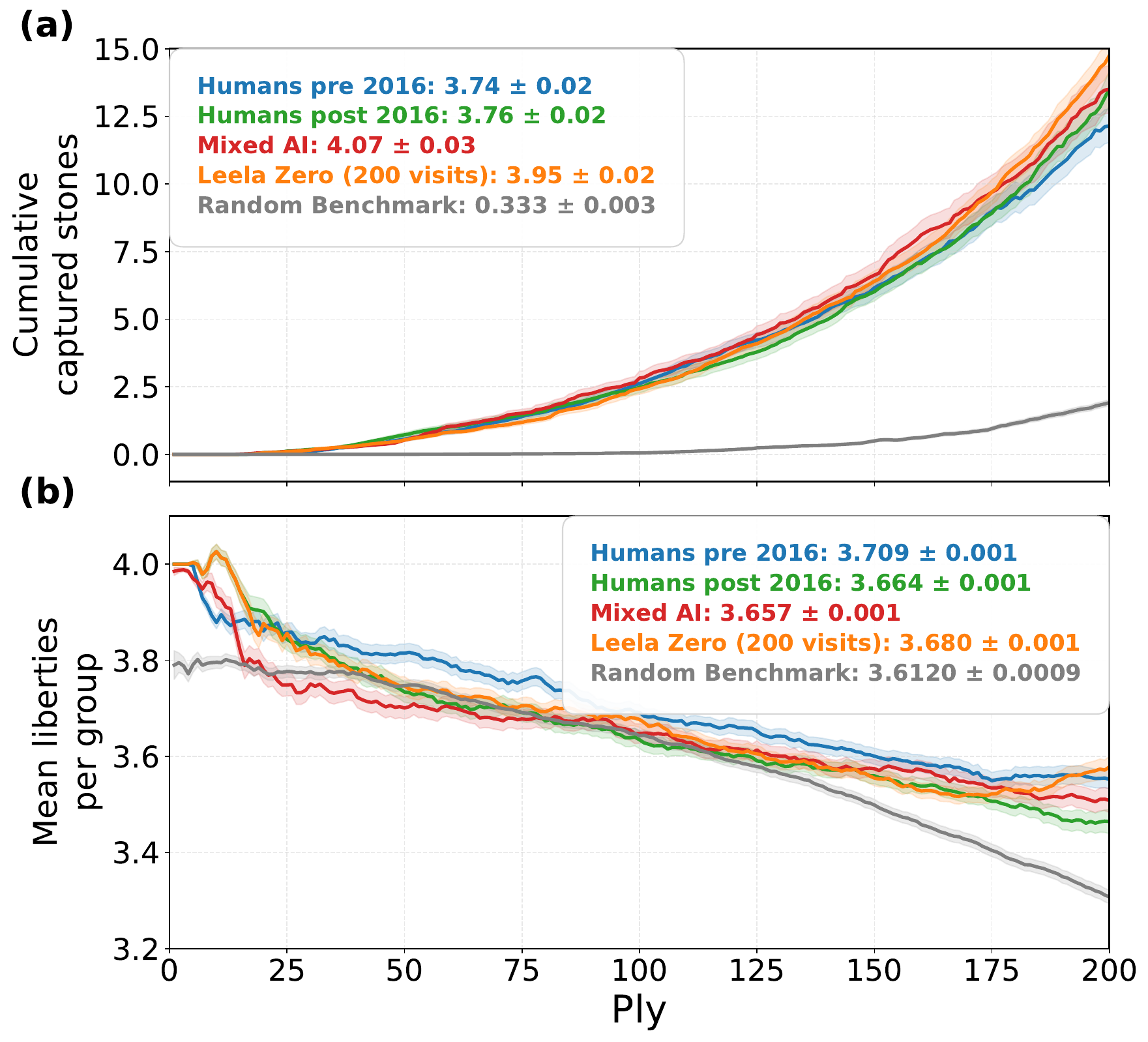}
\caption{\label{fig:structural_captures} \textbf{Evolution of tactical game dynamics across plies.} We consider the same set of games analyzed in Fig.~\ref{fig:tension_evolution}. (a) Cumulative mean number of captured stones. (b) Mean liberties per connected group. The shaded bands represent the standard error of the mean.}
\end{figure}

The dynamic interplay between tension and tactical resolution is further highlighted by tactical captures and group liberties (Fig.~\ref{fig:structural_captures}). Panel (a) shows that captured stones remain near zero during early opening plays before rising steadily during mid-game tactical clashes. State-of-the-art AI achieves an higher mean capture count than human play. Finally, Panel (b) reveals that the mean number of liberties per group decays continuously as the board fills, with pre-2016 humans retaining slightly higher average liberties compared to post-2016 human play and AI.

\section{Spatial and topological playstyle analysis}
\label{sec:spatial_analysis}
\noindent
To uncover the mechanistic drivers behind the tension hierarchy observed in the main text (see Fig.~\ref{fig:tension_evolution}), we evaluate the statistical evolution of four fundamental spatial metrics across the datasets.

\begin{figure}[!htbp]
\centering
\includegraphics[width=1\linewidth]{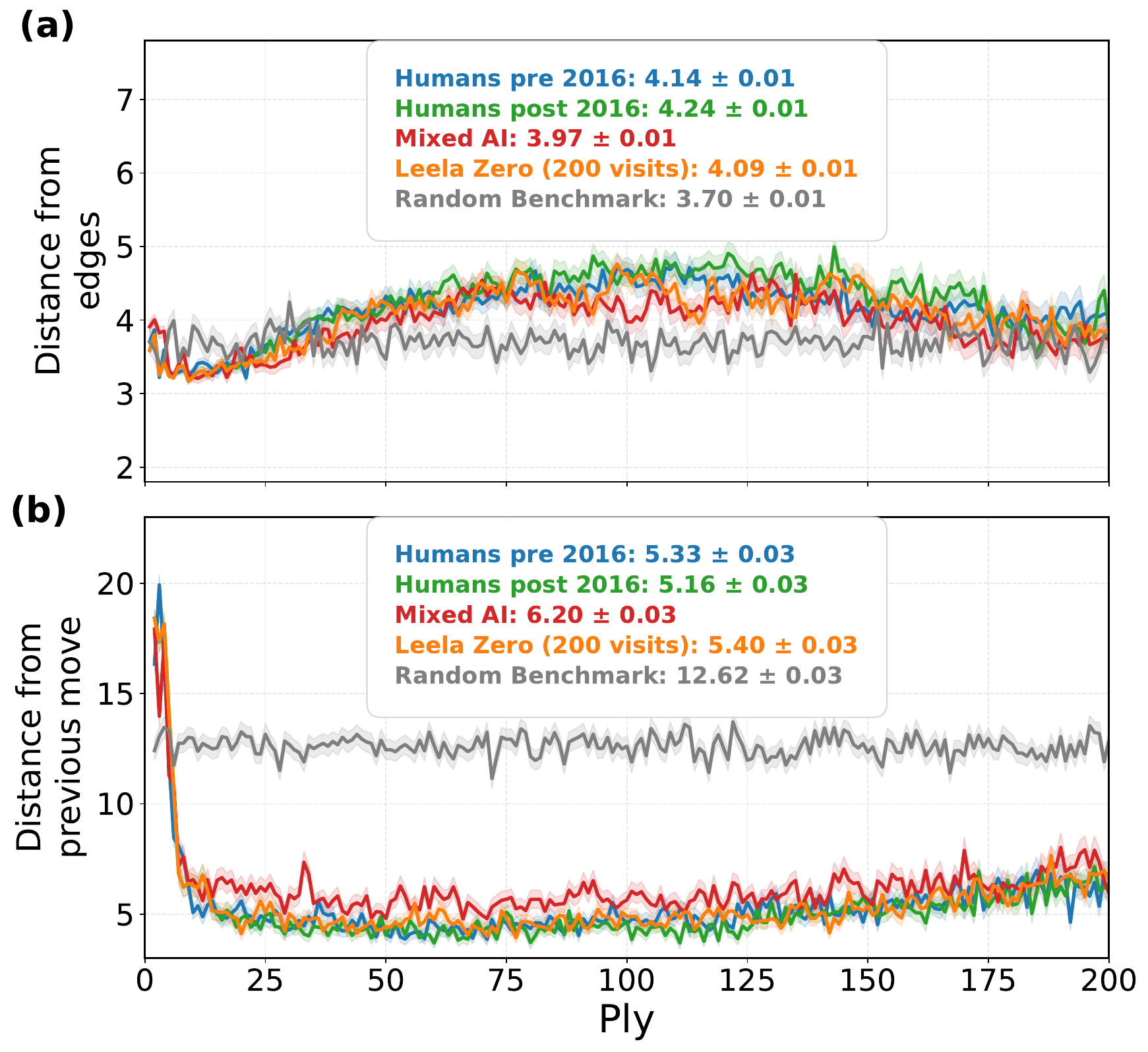}
\caption{\label{fig:spatial_distances} \textbf{Evolution of spatial distance metrics across plies.} We consider the same set of games analyzed in Fig.~\ref{fig:tension_evolution}. (a) Mean Manhattan distance from the nearest edges. (b) Mean Manhattan distance from the previous move played on the board. The shaded bands represent the standard error of the mean.}
\end{figure}

Figure~\ref{fig:spatial_distances} highlights macro-strategic divergences in board utilization. Panel (a) tracks the mean Manhattan distance of played stones from the board edges. Contrary to traditional human opening theories that prioritize high-influence star points, state-of-the-art AI exhibits a slightly lower mean edge distance. Conversely, Panel (b) shows that AI maintains a significantly higher distance from the opponent's previous move. While human play remains reactive and localized around recent skirmishes, AI demonstrates superior global mobility through frequent \textit{tenuki} (playing elsewhere).

\begin{figure}[!htbp]
\centering
\includegraphics[width=1\linewidth]{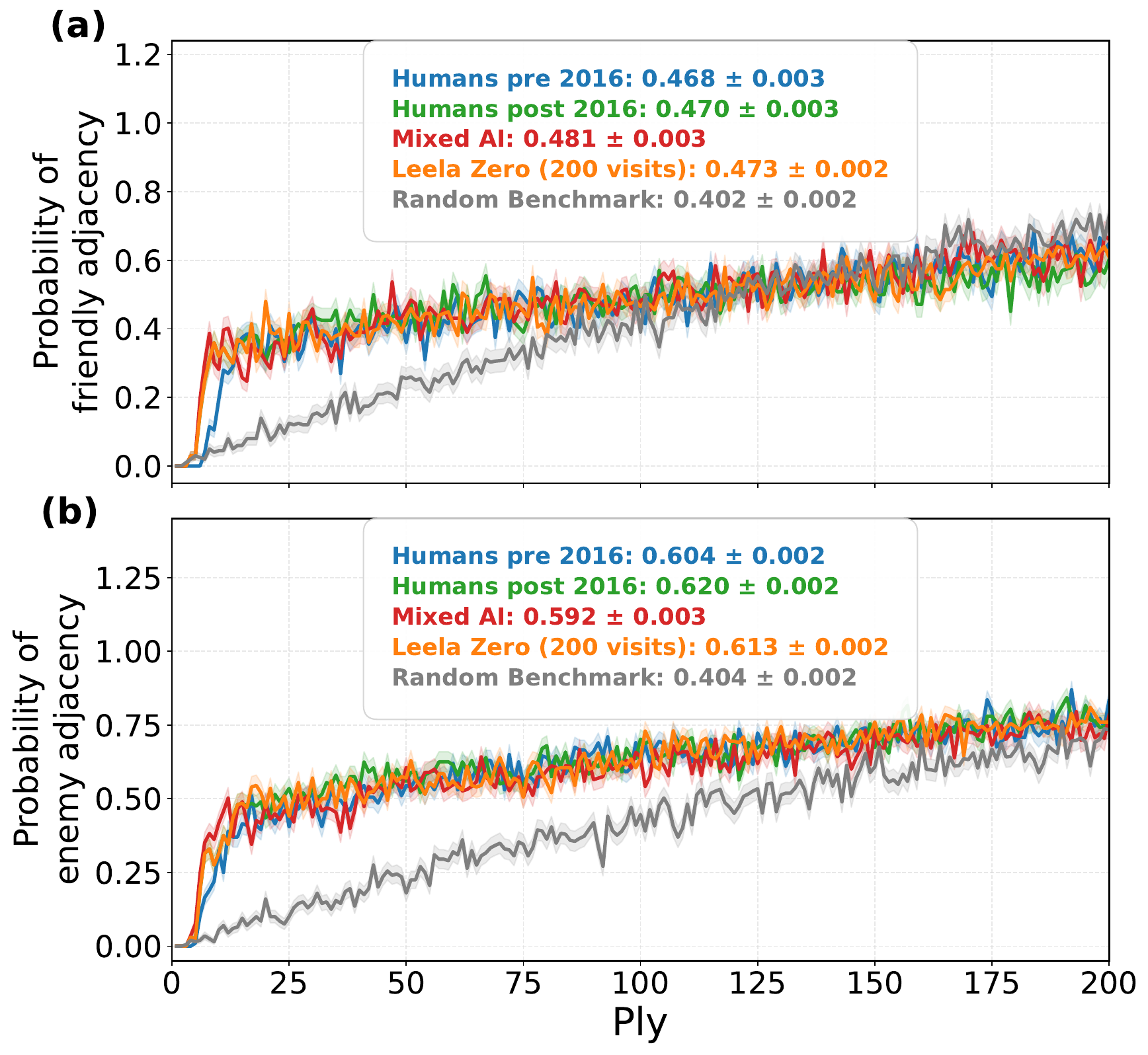}
\caption{\label{fig:spatial_adjacencies} \textbf{Evolution of stone adjacency probabilities across plies.} We consider the same set of games analyzed in Fig.~\ref{fig:tension_evolution}. (a) Probability of playing adjacent to at least one friendly stone. (b) Probability of playing adjacent to at least one enemy stone. The shaded bands represent the standard error of the mean.}
\end{figure}

Direct tactical engagements further reinforce this structural contrast (Fig.~\ref{fig:spatial_adjacencies}). While the probability of friendly stone adjacency (Panel a) grows monotonically across all competent players, AI achieves the highest overall rate. In contrast, enemy adjacency (Panel b) shows that human players, especially in the post-2016 era, exhibit the highest propensity for direct contact. Mixed AI records the lowest enemy contact probability among non-random agents, systematically avoiding premature local attachment.

Together, these spatial signatures show how Mixed AI plays differently from humans: it plays closer to the edge, moves more often to distant areas of the board, and avoids early contact with enemy stones, thus delaying local simplification. These habits alone, however, do not explain its higher tension. As shown by the generative null models (Appendix~\ref{sec:null_model}), games that reproduce only Mixed AI spatial statistics have the lowest tension of all, because scattered stones placed without lookahead tend to form independent, weakly coupled structures. The high tension of Mixed AI therefore arises when these dispersed moves are coordinated by search, so that distant groups and zones become strongly interdependent.

\section{Generative spatial null models}
\label{sec:null_model}
\noindent
To ascertain whether the elevated strategic tension observed in actual AI games is a trivial byproduct of local spatial habits or a genuine non-local topological phenomenon related to strategy, we construct a suite of generative spatial null models. Rather than enforcing global tactical constraints, these models generate synthetic games move-by-move using a Naive Bayes stochastic sampler calibrated on the empirical spatial metrics extracted in Appendix~\ref{sec:spatial_analysis}.

At each ply $k$, the generation algorithm identifies all physically legal intersections on the $19 \times 19$ board by strictly enforcing rules against occupancy, Ko violations, and suicide moves. For each legal candidate move $m$, we calculate its four local spatial features: edge Manhattan distance $d_{\text{edge}}$, distance from the previous move $d_{\text{prev}}$, and binary indicators for friendly $\text{adj}_{\text{friendly}}$ and enemy $\text{adj}_{\text{enemy}}$ orthogonal adjacencies. The candidate move weight $w_k^{(m)}$ is determined by the joint likelihood product of these independent features:
\begin{equation*}
\begin{aligned}
w_k^{(m)} = \; & \exp\left( -\frac{1}{2}\left[\frac{d_{\text{edge}}^{(m)} - \mu_{\text{edge}}(k)}{\sigma_{\text{edge}}(k)}\right]^2 \right) \cdot \\
& \exp\left( -\frac{1}{2}\left[\frac{d_{\text{prev}}^{(m)} - \mu_{\text{prev}}(k)}{\sigma_{\text{prev}}(k)}\right]^2 \right) \cdot \\
& P(\text{adj}_{\text{friendly}}^{(m)} \mid k) \cdot P(\text{adj}_{\text{enemy}}^{(m)} \mid k)
\end{aligned}
\end{equation*}
The continuous spatial features are modeled using Gaussian likelihoods centered at the empirical means $\mu(k)$, with the population standard deviations $\sigma(k)$ used to capture the full variability of move locations across individual games rather than the uncertainty of the mean. Adjacencies are evaluated using empirical Bernoulli likelihoods matching target binary probabilities ($P = \mu_{\text{adj}}$ if the condition is met, and $1 - \mu_{\text{adj}}$ otherwise). Move selection is then performed by stochastically sampling from the normalized probability distribution $p_m = w_m / \sum_j w_j$.

\begin{figure}[!htbp]
\centering
\includegraphics[width=1\linewidth]{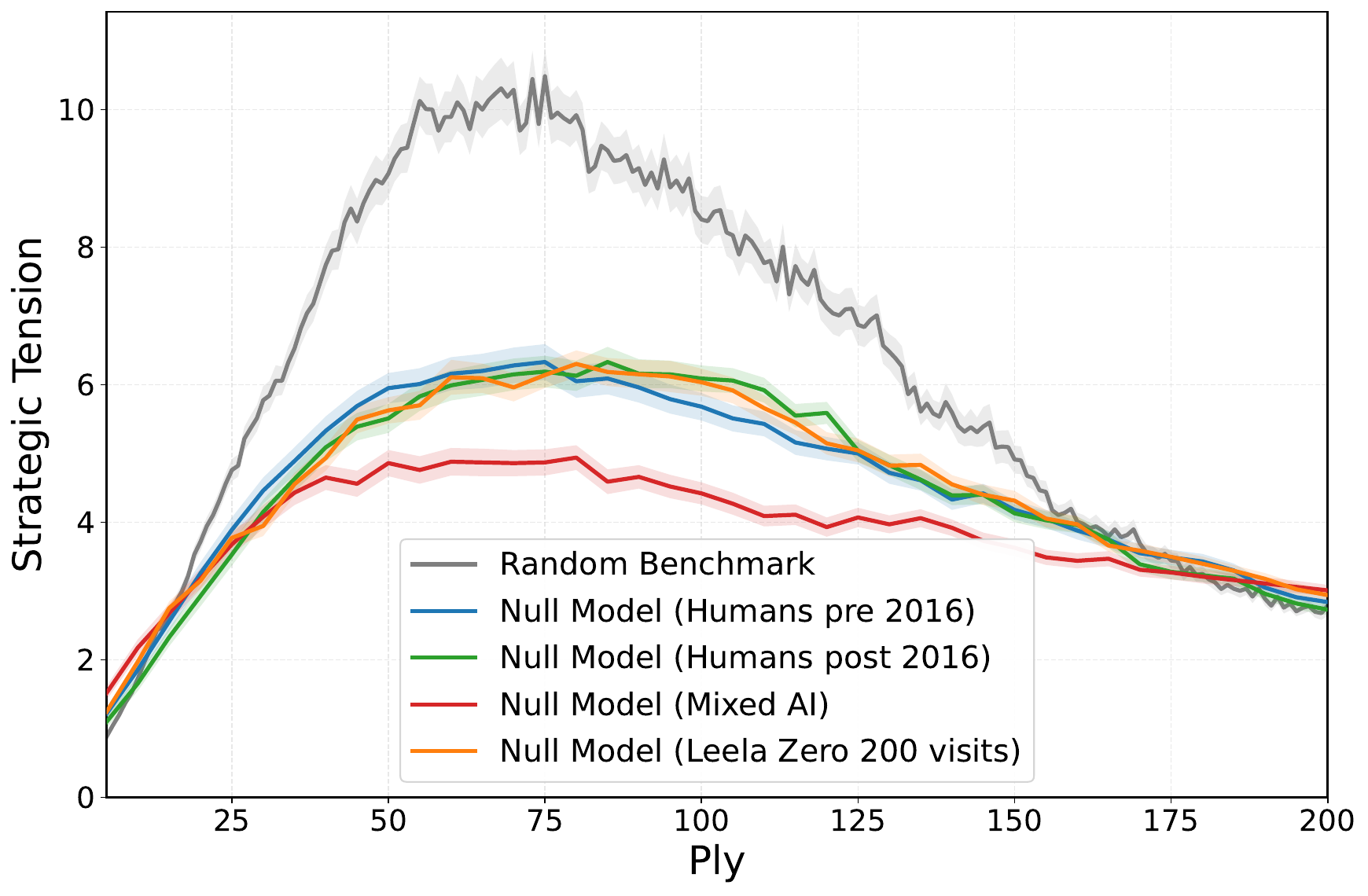}
\caption{\label{fig:null_model_tension} \textbf{Strategic tension evolution in generative spatial null models.} Synthetic games are produced by stochastically sampling legal moves constrained by the empirical spatial feature distributions of each target dataset analyzed in Fig.~\ref{fig:tension_evolution}.}
\end{figure}

Figure~\ref{fig:null_model_tension} displays the resulting tension evolution across the synthetic ensembles. Crucially, the spatial null models reveal a striking reversal compared to actual gameplay: synthetic games constrained by Mixed AI spatial habits sustain a distinctly lower tension than those derived from human spatial profiles.

This divergence offers a helpful perspective on the interplay between spatial habits and strategic planning. In the absence of lookahead search and positional evaluation, diversified AI-like spatial tendencies, such as wider move-to-move transitions ($d_{\text{prev}}$) and a lower frequency of immediate contact ($P_{\text{enemy}}$), naturally foster a more diffuse arrangement of stones across the board. Without active tactical coordination to weave these distant stones together, however, such dispersion tends to settle into relatively independent, self-contained formations that naturally moderate overall strategic tension.

These findings suggest that the high strategic tension characteristic of genuine AI play cannot be fully accounted for by local spatial preferences alone. Instead, it reflects the role of deep tactical lookahead, which purposefully coordinates geographically dispersed stones into tightly coupled, globally interdependent structures.

\section{Temporal error dynamics and competitive phase filtering}
\label{sec:appendix_error_analysis}
\noindent
To examine how macroscopic strategic tension connects to tactical precision, we analyze the temporal evolution of move inaccuracies across game progression in human and artificial agents. Each board state is evaluated using KataGo at 1000 visits per position with a 7.5-point komi. For each ply, point loss is measured as the score degradation incurred by the played move relative to KataGo's optimal choice. A move is formally classified as an error if this loss exceeds $\Delta S_{\text{thresh}} = 3.0$ points, isolating genuine tactical blunders while filtering out minor evaluation fluctuations. Since KataGo serves as the reference, this criterion may favor engines that share its design (neural networks combined with tree search) relative to human players, even though KataGo itself is not among the evaluated engines. Neural evaluators can also be systematically misled in specific group configurations~\cite{wang2023adversarial}, which is a further limitation of using an engine as the reference.

\begin{figure}[!htbp]
\centering
\includegraphics[width=1\linewidth]{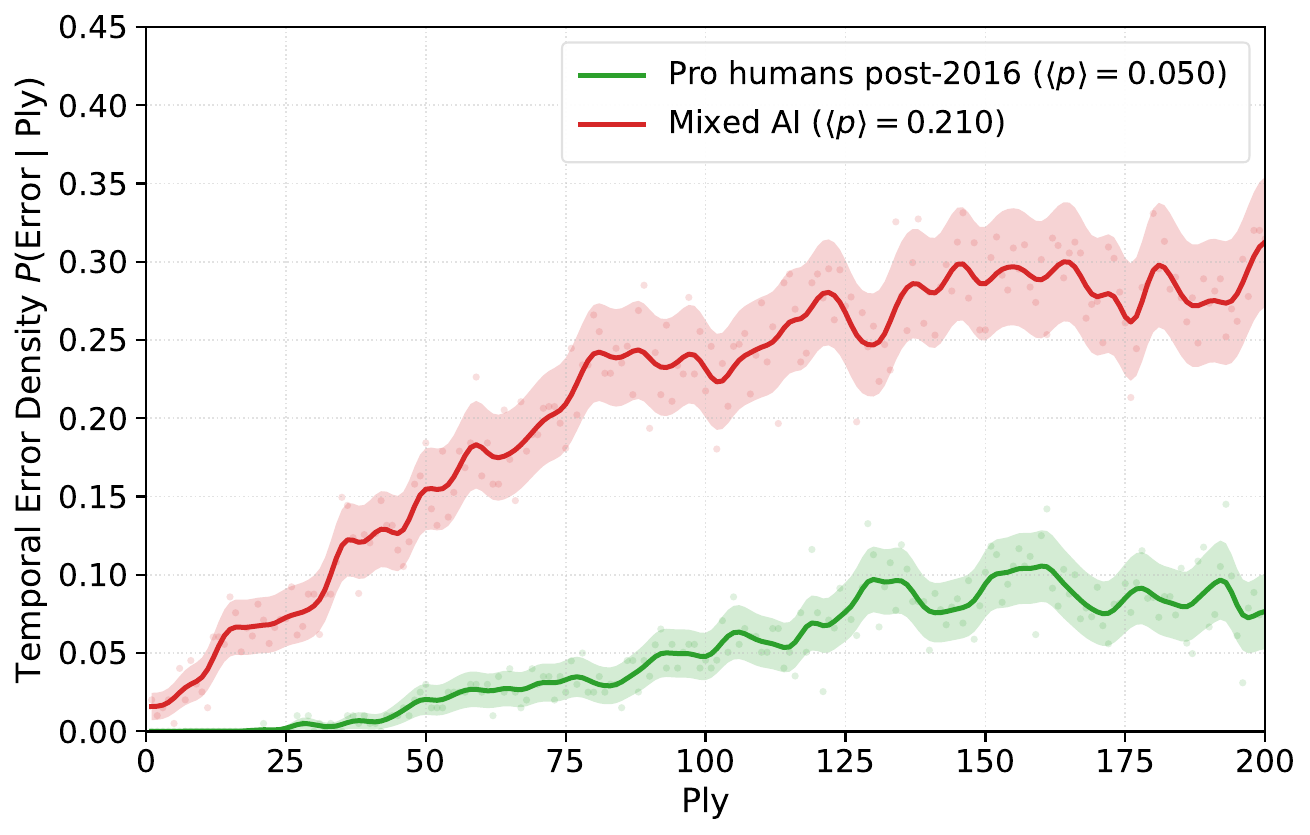}
\caption{\label{fig:error_density_raw} \textbf{Temporal error density across human and machine play.} 
Unconstrained empirical probability $P(\text{Error} \mid \text{Ply})$ across plies 1--200 for post-2016 human professionals (green) and Mixed AI (red), evaluated at a score-lead threshold $\Delta S_{\text{thresh}} = 3.0$ points ($N = 200$ games per cohort, matching Fig.~\ref{fig:tension_evolution}). Solid curves denote trajectories smoothed via a 1D Gaussian filter ($\sigma = 2.0$ plies), shaded ribbons represent the standard error of the mean (s.e.m.), and dots indicate raw per-ply empirical frequencies. Global average error rates $\langle p \rangle$ are reported in the legend.}
\end{figure}

Evaluating the raw error density across the entire $200$-ply progression yields an apparent paradox (see Fig.~\ref{fig:error_density_raw}): AI players appear to commit more than four times as many errors as human professionals ($\langle p \rangle_{\text{AI}} = 0.210$ vs. $\langle p \rangle_{\text{Human}} = 0.050$), with AI error density escalating beyond $25$--$30\%$ in the late game. This discrepancy arises from two behavioral artifacts inherent to artificial gameplay:
\begin{enumerate}
    \item \textit{Win-rate optimization vs. point maximization (slack moves):} Value-network engines maximize win probability rather than final point margins. Once a decisive lead is secured ($\text{WR} > 90\%$), the engine often plays safe, suboptimal moves (\textit{slack moves}) to simplify the position and remove tactical risk. Conceding a few points guarantees the win, but an objective score metric flags these moves as errors.
    \item \textit{Absence of resignation and desperation play:} Human professionals typically resign once a deficit becomes insurmountable, keeping recorded games focused on accurate play. In contrast, tournament AI engines often play out completely lost positions ($\text{WR} < 5\%$), attempting hopeless invasions into settled territory (\textit{desperation play}) where each captured stone is registered as an error.
\end{enumerate}

To isolate authentic tactical fragility, we filter the dataset to include strictly contested positions ($0.10 \le \text{WR} \le 0.90$). To maintain continuous temporal integrity, each game trajectory is analyzed sequentially and truncated at the first ply where the win rate exits the $[10\%, 90\%]$ interval, discarding all subsequent moves. 

\begin{figure}[!htbp]
\centering
\includegraphics[width=1\linewidth]{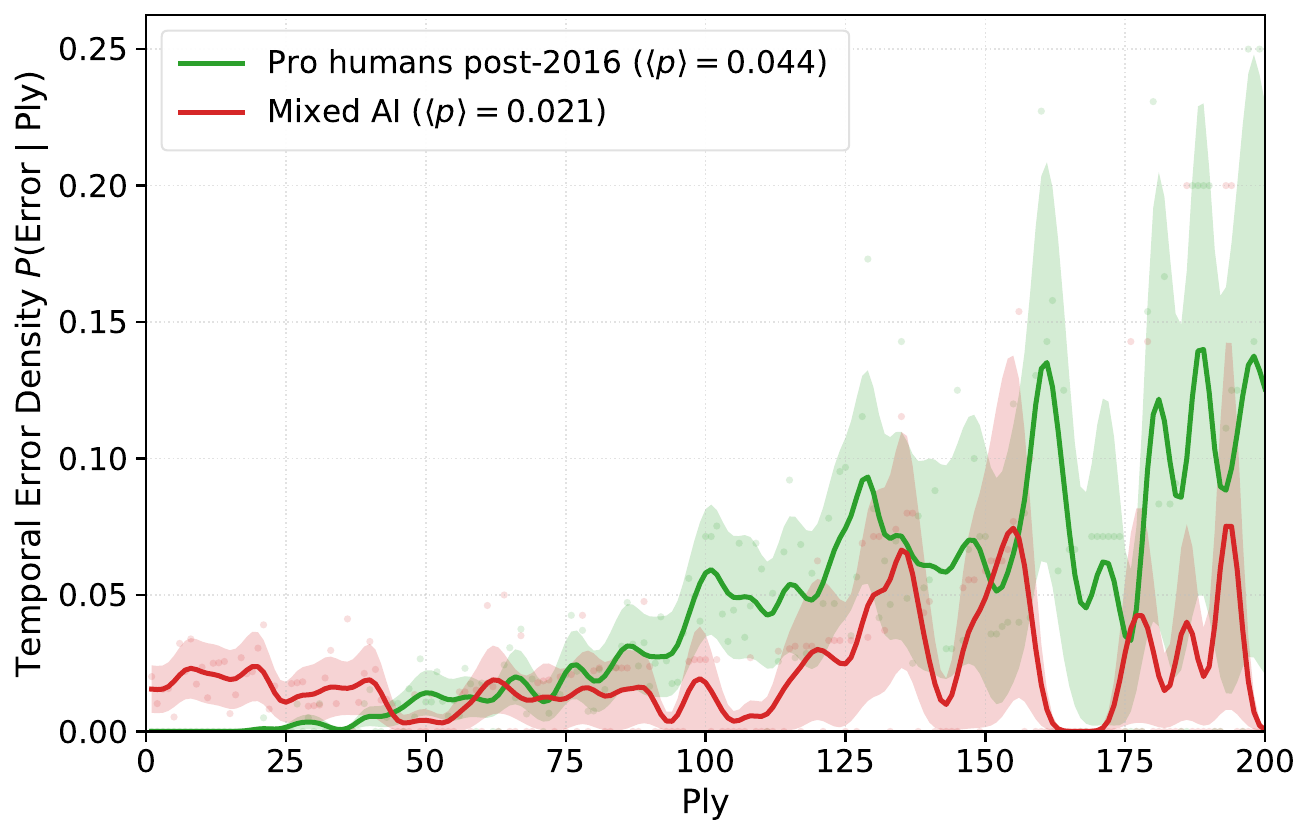}
\caption{\label{fig:error_density_filtered} \textbf{Temporal error dynamics conditioned on competitive balance.} 
Empirical probability $P(\text{Error} \mid \text{Ply})$ restricted strictly to contested positions ($10\% \le \mathrm{WR} \le 90\%$) across plies 1--200 for post-2016 human professionals (green) and Mixed AI (red), evaluated at a score-lead threshold $\Delta S_{\mathrm{thresh}} = 3.0$ points ($N = 200$ games per cohort, matching Fig.~\ref{fig:tension_evolution}). Solid curves denote trajectories smoothed via a 1D Gaussian filter ($\sigma = 2.0$ plies), shaded ribbons indicate the standard error of the mean (s.e.m.), and dots represent raw per-ply empirical frequencies. Global average error rates $\langle p \rangle$ are reported in the legend.}
\end{figure}

As shown in Fig.~\ref{fig:error_density_filtered}, contested filtering resolves the artifact: in genuinely contested states, the AI error rate drops to less than half that of human professionals ($\langle p \rangle_{\text{AI}} = 0.021$ vs. $\langle p \rangle_{\text{Human}} = 0.044$). Human professionals make almost no errors in the opening (plies 1--40), but their error rate rises to about $5$--$10\%$ in the middlegame (plies 80--150) and approaches $15\%$ in the endgame (plies 170--190). Possible explanations include the growing complexity of the position, fatigue, and time pressure in \textit{byo-yomi}. Conversely, artificial agents sustain a highly stable, dampened error profile throughout all phases, experiencing minor fluctuations during all possible game moments.

\begin{figure}[!htbp]
\centering
\includegraphics[width=1\linewidth]{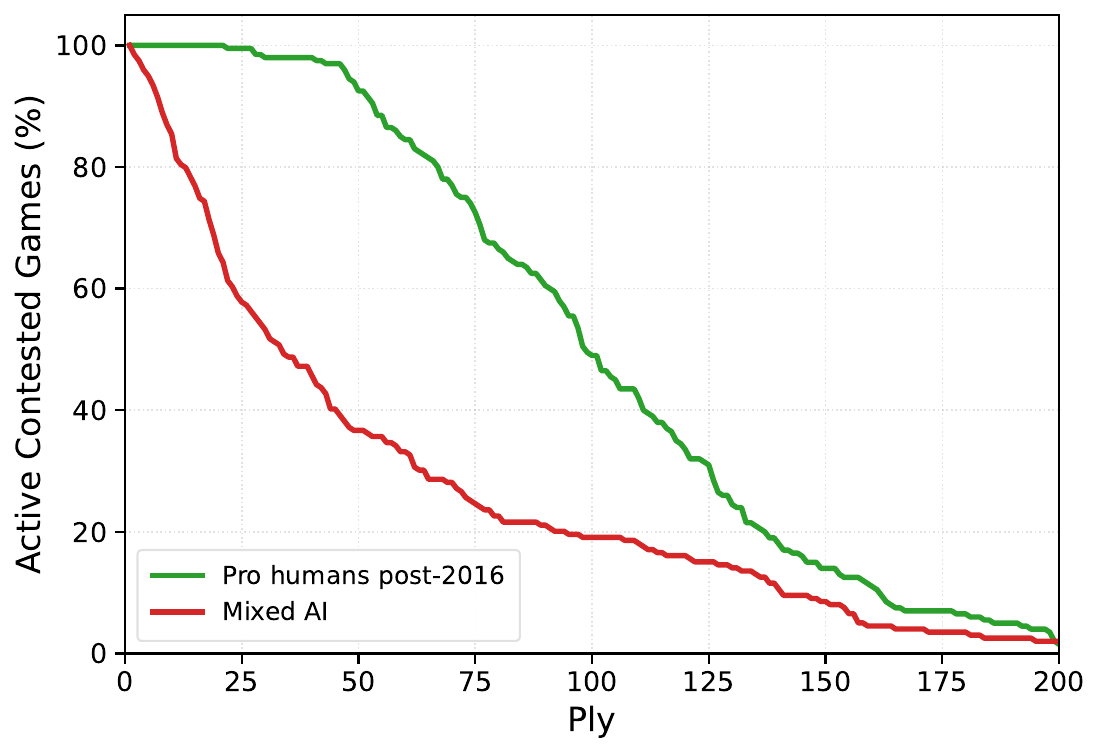}
\caption{\label{fig:contested_survival} \textbf{Persistence of competitive parity across game progression.} 
Survival trajectories showing the percentage of active matches remaining within the contested win-rate window ($10\% \le \mathrm{WR} \le 90\%$) as a function of ply number for post-2016 human professionals (green) and Mixed AI (red), evaluated on $N = 200$ games per cohort.}
\end{figure}

The early departure of games from the contested window generates the survival profile shown in Fig.~\ref{fig:contested_survival}. Human professional matches maintain an extended plateau of competitive balance, with $100\%$ of games remaining contested through ply 35. This prolonged equilibrium stems from reciprocal human inaccuracies that repeatedly reopen the score margin. In contrast, AI matches exhibit rapid competitive divergence, with active contested games decaying sharply after the opening. Because artificial agents convert small opening leads ($1.5$--$2.0$ points) into decisive win-rate advantages without allowing tactical recovery, AI tournament games settle the balance of power much earlier in the match.

\section{Skill-dependent scaling of structural and spatial metrics}
\label{sec:rank_metrics_appendix}
\noindent
To complement the rank-dependent analysis of strategic tension presented in the main text (see Fig.~\ref{fig:tension_rank}), we examine how the underlying structural and spatial metrics scale across player expertise from 18 Kyu to 9 Dan.

\begin{figure}[!htbp]
\centering
\includegraphics[width=1\linewidth]{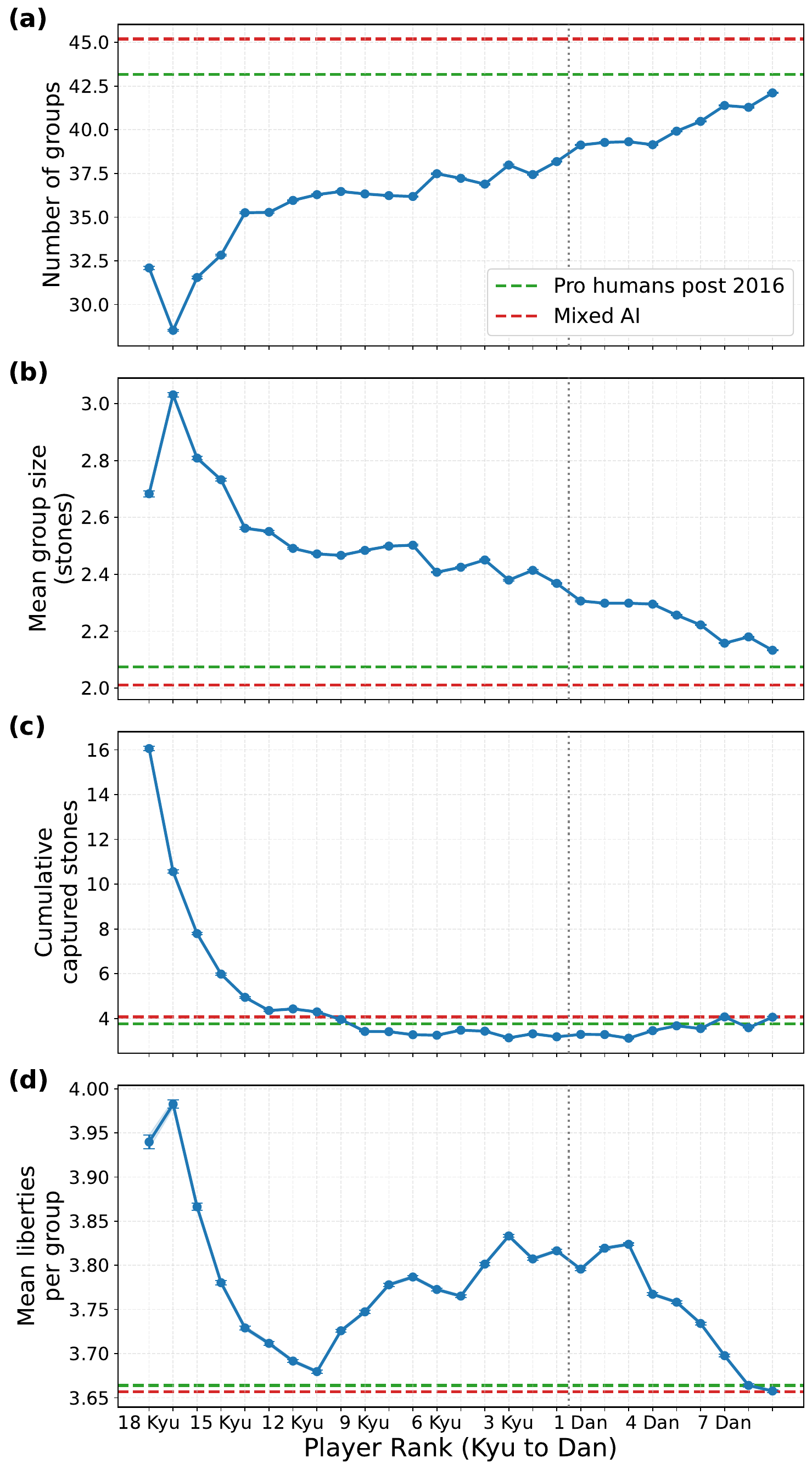}
\caption{\label{fig:structural_metrics_rank} \textbf{Structural metrics across skill levels.} 
Ranks from 18 Kyu to 9 Dan, as in Fig.~\ref{fig:tension_rank}. To remove biases from different game lengths, all quantities are computed at each ply and averaged over plies 1--200.
\textbf{(a)} Number of stone groups. 
\textbf{(b)} Group size (stones per connected group). 
\textbf{(c)} Cumulative captured stones, $\sum_{k=1}^{200} C_k$, where $C_k$ is the number of prisoners up to ply $k$. 
\textbf{(d)} Liberties per group. 
Dashed lines: post-2016 professionals (green) and Mixed AI (red). Shaded ribbons: $\pm 1$ standard error of the mean.}
\end{figure}

Fig~\ref{fig:structural_metrics_rank} illustrates the structural transformation of the board state as player skill advances. Panel (a) shows that the average number of distinct groups increases systematically with player rank, as novices tend to consolidate stones into a few clusters while stronger players maintain a more distributed board architecture that approaches high-level professional and AI baselines. Conversely, Panel (b) reveals that mean group size drops sharply from beginner levels before stabilizing among Dan players, indicating that novices routinely over-invest in heavy, contiguous stone formations. Panel (c) highlights a dramatic decrease in tactical captures across the skill spectrum: beginner games are characterized by widespread tactical blunders and high capture counts, whereas advancing expertise fosters defensive awareness and shape stability, bringing capture rates down toward professional thresholds. Finally, Panel (d) shows that average group liberties do not change monotonically with rank: they drop among lower kyu players, rise again around 3 kyu--1 dan, and decrease among the strongest dan players, reaching the professional and Mixed AI values.

\begin{figure}[!htbp]
\centering
\includegraphics[width=1\linewidth]{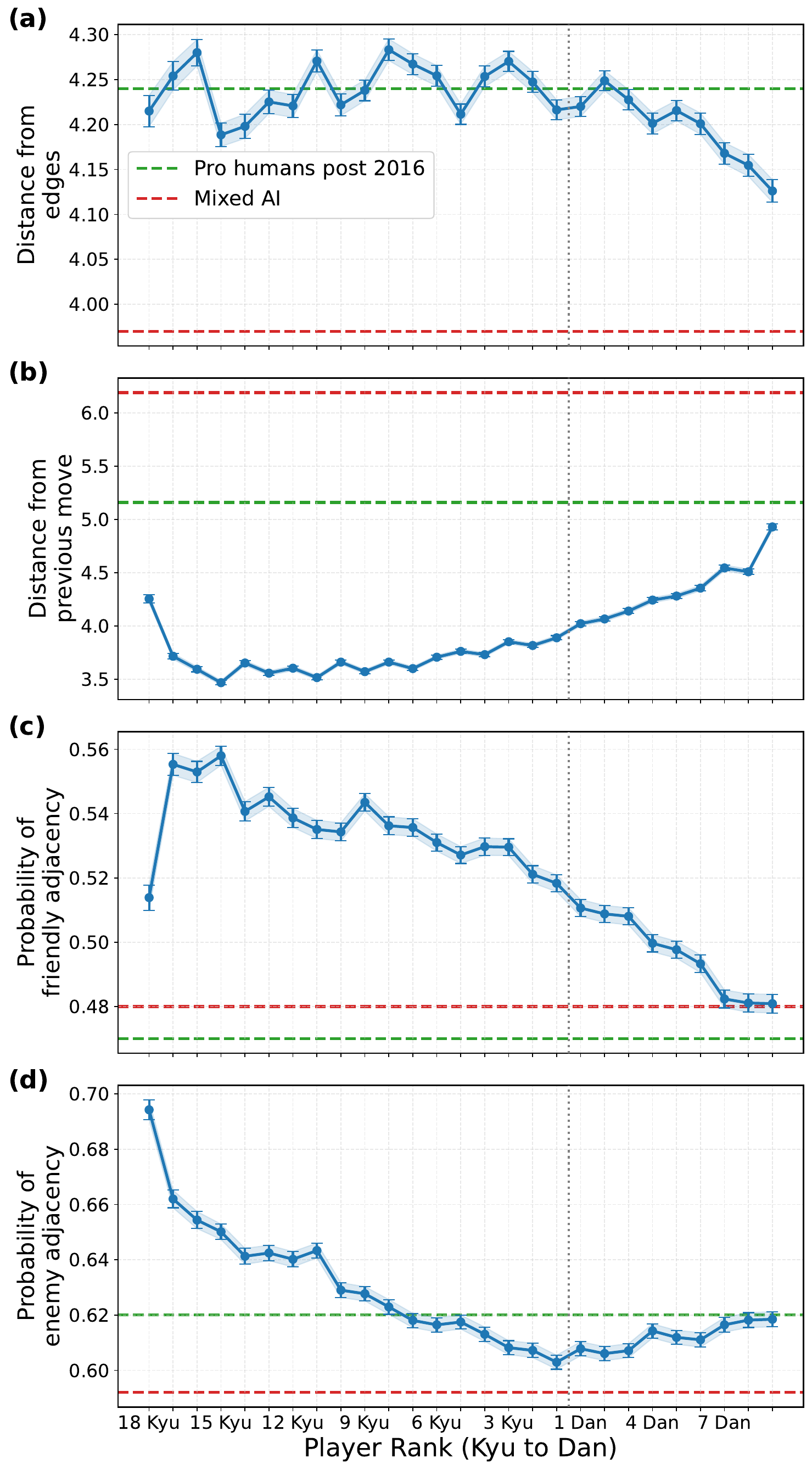}
\caption{\label{fig:spatial_metrics_rank} \textbf{Spatial move statistics across skill levels.} 
Ranks from 18 Kyu to 9 Dan, as in Fig.~\ref{fig:tension_rank}. To remove biases from different game lengths, all metrics are computed at each ply and averaged over plies 1--200.
\textbf{(a)} Manhattan distance of played stones from the nearest edge. 
\textbf{(b)} Manhattan distance from the previous move. 
\textbf{(c)} Probability of playing adjacent to a friendly stone. 
\textbf{(d)} Probability of playing adjacent to an enemy stone. 
Dashed lines: post-2016 professionals (green) and Mixed AI (red). Shaded ribbons: $\pm 1$ standard error of the mean.}
\end{figure}

The spatial move preferences across ranks (Fig.~\ref{fig:spatial_metrics_rank}) further elucidate the mechanical origin of these structural trends. Panel (a) shows that the distance from the board edge changes little across most ranks and stays close to the professional value, but decreases among the strongest dan players, moving toward the lower values of Mixed AI. Panel (b) demonstrates an almost steady, monotonic rise in inter-move distance with increasing rank: as players improve, they break away from local, reactive skirmishes and adopt broader global vision, progressively closing the mobility gap with superhuman AI. This transition is further reinforced by local adjacency statistics. Beginners exhibit a strong propensity for immediate local attachment, characterized by high probabilities of both friendly (Panel c) and enemy (Panel d) adjacencies. As expertise grows, both contact probabilities decline sharply, demonstrating that advancing skill is defined by cleaner stone efficiency, reduced local clutter, and a deliberate avoidance of premature tactical entanglement.

\newpage

\bibliography{references}

@book{vonNeumann1944,
  title={Theory of Games and Economic Behavior},
  author={Von Neumann, John and Morgenstern, Oskar},
  year={1944},
  publisher={Princeton University Press}
}

@book{kahneman2011thinking, title={Thinking, fast and slow}, author={Kahneman, Daniel}, year={2011}, publisher={Farrar, Straus and Giroux}}

@article{cerioli2025ai,
  title = {Artificial intelligence sustains higher strategic tension than humans in chess},
  author = {Cerioli, Adamo and Lee, Edward D. and Servedio, Vito D. P.},
  journal = {APS Open Sci.},
  volume = {1},
  pages = {000131},
  numpages = {30},
  year = {2026},
  month = {Sep},
  publisher = {American Physical Society},
  doi = {10.1103/63cv-52fj},
  url = {https://link.aps.org/doi/10.1103/63cv-52fj}
}

@book{kennedy2010rise,
  title={The rise and fall of the great powers: economic change and military conflict from 1500 to 2000},
  author={Kennedy, Paul},
  year={2010},
  publisher={Vintage}
}

@article{march1991exploration,
  title={Exploration and exploitation in organizational learning},
  author={March, James G},
  journal={Organization science},
  volume={2},
  number={1},
  pages={71--87},
  year={1991},
  publisher={INFORMS},
  doi={10.1287/orsc.2.1.71}
}

@article{harre2011aggregate,
  title={The aggregate complexity of decisions in the game of Go},
  author={Harr{\'e}, Michael S and Bossomaier, Terence and Gillett, A and Snyder, A},
  journal={The European Physical Journal B},
  volume={80},
  number={4},
  pages={555--563},
  year={2011},
  publisher={Springer},
  doi={10.1140/epjb/e2011-10905-8}
}

@inproceedings{morandin2019sai,
  title={SAI a sensible artificial intelligence that plays go},
  author={Morandin, Francesco and Amato, Gianluca and Gini, Rosa and Metta, Carlo and Parton, Maurizio and Pascutto, Gian-Carlo},
  booktitle={2019 International Joint Conference on Neural Networks (IJCNN)},
  pages={1--8},
  year={2019},
  organization={IEEE},
  doi={10.1109/ijcnn.2019.8852266}
}

@book{allis1994searching,
  title={Searching for solutions in games and artificial intelligence},
  author={Allis, Louis Victor and others},
  volume={14},
  year={1994},
  publisher={Ponsen \& Looijen Wageningen}
}

@article{silver2017mastering,
  title={{Mastering the game of Go without human knowledge}},
  author={Silver, David and Schrittwieser, Julian and Simonyan, Karen and Antonoglou, Ioannis and Huang, Aja and Guez, Arthur and Hubert, Thomas and Baker, Lucas and Lai, Matthew and Bolton, Adrian and others},
  journal={nature},
  volume={550},
  number={7676},
  pages={354--359},
  year={2017},
  publisher={Nature Publishing Group UK London},
  doi={10.1038/nature24270}
}

@inproceedings{bossomaier2012neuro,
  title={Neuro-cognitive model of move location in the game of Go},
  author={Bossomaier, Terry and Traish, Jason and Gobet, Fernand and Lane, Peter CR},
  booktitle={The 2012 International Joint Conference on Neural Networks (IJCNN)},
  pages={1--7},
  year={2012},
  organization={IEEE},
  doi={10.1109/ijcnn.2012.6252377}
}

@article{benson1976life,
  title={Life in the game of Go},
  author={Benson, David B},
  journal={Information Sciences},
  volume={10},
  number={1},
  pages={17--29},
  year={1976},
  publisher={Elsevier},
  doi={10.1016/0020-0255(76)90059-1}
}

@article{silver2016mastering,
  title={Mastering the game of Go with deep neural networks and tree search},
  author={Silver, David and Huang, Aja and Maddison, Chris J and Guez, Arthur and Sifre, Laurent and Van Den Driessche, George and Schrittwieser, Julian and Antonoglou, Ioannis and Panneershelvam, Veda and Lanctot, Marc and others},
  journal={nature},
  volume={529},
  number={7587},
  pages={484--489},
  year={2016},
  publisher={Nature Publishing Group UK London},
  doi={10.1038/nature16961}
}

@inproceedings{shin2021human,
  title={Human learning from artificial intelligence: Evidence from human Go players’ decisions after AlphaGo},
  author={Shin, Minkyu and Kim, Jin and Kim, Minkyung},
  booktitle={Proceedings of the annual meeting of the cognitive science society},
  volume={43},
  number={43},
  year={2021}
}

@article{adler1965topological,
  title={Topological entropy},
  author={Adler, Roy L and Konheim, Alan G and McAndrew, M Harry},
  journal={Transactions of the American Mathematical Society},
  volume={114},
  number={2},
  pages={309--319},
  year={1965},
  publisher={JSTOR},
  doi={10.1090/S0002-9947-1965-0175106-9}
}

@article{bowen1971entropy,
  title={Entropy for group endomorphisms and homogeneous spaces},
  author={Bowen, Rufus},
  journal={Transactions of the American Mathematical Society},
  volume={153},
  pages={401--414},
  year={1971},
  doi={10.1090/S0002-9947-1971-0274707-X}
}

@article{dinaburg1970correlation,
  title={A correlation between topological entropy and metric entropy},
  author={Dinaburg, Efim Isaakovich},
  journal={Doklady Akademii Nauk SSSR},
  volume={190},
  number={1},
  pages={19--22},
  year={1970},
  organization={Russian Academy of Sciences}
}

@article{wu2019accelerating,
  title={Accelerating self-play learning in go},
  author={Wu, David J},
  journal={arXiv preprint arXiv:1902.10565},
  year={2019},
  url={https://arxiv.org/abs/1902.10565}
}

@book{van2023graph,
  title={{Graph Spectra for Complex Networks}},
  author={Van Mieghem, Piet},
  year={2023},
  publisher={Cambridge University Press},
  chapter={4},
  doi={10.1017/9781009366793}
}

@article{pillai2005perron,
  title={The Perron-Frobenius theorem: some of its applications},
  author={Pillai, S Unnikrishna and Suel, Torsten and Cha, Seunghun},
  journal={IEEE Signal Processing Magazine},
  volume={22},
  number={2},
  pages={62--75},
  year={2005},
  doi={10.1109/MSP.2005.1407714}
}

@article{castellano2017relating,
  title={{Relating topological determinants of complex networks to their spectral properties: Structural and dynamical effects}},
  author={Castellano, Claudio and Pastor-Satorras, Romualdo},
  journal={Physical Review X},
  volume={7},
  number={4},
  pages={041024},
  year={2017},
  publisher={APS},
  doi={10.1103/PhysRevX.7.041024}
}

@article{hammond2006hidden,
  title={The hidden traps in decision making},
  author={Hammond, John S and Keeney, Ralph L and Raiffa, Howard and others},
  journal={Harvard business review},
  volume={84},
  number={1},
  pages={118},
  year={2006}
}

@article{baddeley2012working,
  title={{Working memory: Theories, models, and controversies}},
  author={Baddeley, Alan},
  journal={Annual review of psychology},
  volume={63},
  number={1},
  pages={1--29},
  year={2012},
  publisher={Annual Reviews}
}

@article{sweller1988cognitive,
  title={{Cognitive load during problem solving: Effects on learning}},
  author={Sweller, John},
  journal={Cognitive science},
  volume={12},
  number={2},
  pages={257--285},
  year={1988},
  publisher={Elsevier},
  doi={10.1207/s15516709cog1202_4}
}

@article{inglis2000central,
  title={The central role of uncertainty reduction in determining behaviour},
  author={Inglis, Ian},
  journal={Behaviour},
  volume={137},
  number={12},
  pages={1567--1599},
  year={2000},
  publisher={Brill},
  doi={ 10.1163/156853900502727}
}

@article{heinemann2009measuring,
  title={Measuring strategic uncertainty in coordination games},
  author={Heinemann, Frank and Nagel, Rosemarie and Ockenfels, Peter},
  journal={The review of economic studies},
  volume={76},
  number={1},
  pages={181--221},
  year={2009},
  publisher={Wiley-Blackwell}
}

@article{wurman2023improving,
  title={Improving artificial intelligence with games},
  author={Wurman, Peter R and Stone, Peter and Spranger, Michael},
  journal={Science},
  volume={381},
  number={6654},
  pages={147--148},
  year={2023},
  publisher={American Association for the Advancement of Science},
  doi={10.1126/science.adh8135}
}

@article{popenici2017exploring,
  title={Exploring the impact of artificial intelligence on teaching and learning in higher education},
  author={Popenici, Stefan AD and Kerr, Sharon},
  journal={Research and practice in technology enhanced learning},
  volume={12},
  number={1},
  pages={22},
  year={2017},
  publisher={Springer},
  doi={10.1186/s41039-017-0062-8}
}

@article{diaz2024artificial,
  title={Artificial intelligence for teaching and learning in schools: The need for pedagogical intelligence},
  author={D{\'\i}az, Brayan and Nussbaum, Miguel},
  journal={Computers \& Education},
  volume={217},
  pages={105071},
  year={2024},
  publisher={Elsevier},
  doi={10.1016/j.compedu.2024.105071}
}

@article{gode1993allocative,
  title={Allocative efficiency of markets with zero-intelligence traders: Market as a partial substitute for individual rationality},
  author={Gode, Dhananjay K and Sunder, Shyam},
  journal={Journal of political economy},
  volume={101},
  number={1},
  pages={119--137},
  year={1993},
  publisher={The University of Chicago Press},
  doi={10.1086/261868}
}

@article{mckelvey1995quantal,
  title={Quantal response equilibria for normal form games},
  author={McKelvey, Richard D and Palfrey, Thomas R},
  journal={Games and economic behavior},
  volume={10},
  number={1},
  pages={6--38},
  year={1995},
  publisher={Elsevier},
  doi={10.1006/game.1995.1023}
}

@article{meta2022human,
  title={Human-level play in the game of Diplomacy by combining language models with strategic reasoning},
  author={{Meta Fundamental AI Research Diplomacy Team (FAIR)†} and Bakhtin, Anton and Brown, Noam and Dinan, Emily and Farina, Gabriele and Flaherty, Colin and Fried, Daniel and Goff, Andrew and Gray, Jonathan and Hu, Hengyuan and others},
  journal={Science},
  volume={378},
  number={6624},
  pages={1067--1074},
  year={2022},
  publisher={American Association for the Advancement of Science},
  doi={10.1126/science.ade9097}
}

@article{perolat2022mastering,
  title={Mastering the game of Stratego with model-free multiagent reinforcement learning},
  author={Perolat, Julien and De Vylder, Bart and Hennes, Daniel and Tarassov, Eugene and Strub, Florian and de Boer, Vincent and Muller, Paul and Connor, Jerome T and Burch, Neil and Anthony, Thomas and others},
  journal={Science},
  volume={378},
  number={6623},
  pages={990--996},
  year={2022},
  publisher={American Association for the Advancement of Science},
  doi={ 10.1126/science.add4679}
}

@article{helbing2013globally,
  title={Globally networked risks and how to respond},
  author={Helbing, Dirk},
  journal={Nature},
  volume={497},
  number={7447},
  pages={51--59},
  year={2013},
  publisher={Nature Publishing Group UK London},
  doi={10.1038/nature12047}
}

@article{scheffer2012anticipating,
  title={Anticipating critical transitions},
  author={Scheffer, Marten and Carpenter, Stephen R and Lenton, Timothy M and Bascompte, Jordi and Brock, William and Dakos, Vasilis and Van de Koppel, Johan and Van de Leemput, Ingrid A and Levin, Simon A and Van Nes, Egbert H and others},
  journal={science},
  volume={338},
  number={6105},
  pages={344--348},
  year={2012},
  publisher={American Association for the Advancement of Science},
  doi={10.1126/science.1225244}
}

@article{punzi2026learning,
  title={Learning paradigms for hybrid decision-making systems},
  author={Punzi, Clara and Pellungrini, Roberto and Setzu, Mattia and Giannotti, Fosca and Pedreschi, Dino},
  journal={ACM Computing Surveys},
  year={2026},
  publisher={ACM New York, NY},
  doi={10.1145/3802522}
}

@article{chen2025generative,
  title={{Generative AI as a reflective scaffold in a UAV-based STEM project: A mixed-methods study on students’ higher-order thinking and cognitive transformation}},
  author={Chen, Shih-Yeh and Chen, Wei-Cheng and Lai, Chin-Feng},
  journal={Education and Information Technologies},
  volume={30},
  number={17},
  pages={24787--24814},
  year={2025},
  publisher={Springer},
  doi={10.1007/s10639-025-13758-4}
}

@article{liu2026tool,
  title={{Tool, tutor, or crutch?: A grounded theory of cognitive scaffolding and offloading in AI-assisted programming education}},
  author={Liu, Dandan and Fan, Guangrui and Pan, Lihu},
  journal={International Journal of STEM Education},
  volume={13},
  number={1},
  pages={10},
  year={2026},
  publisher={Springer},
  doi={10.1186/s40594-025-00592-w}
}

@article{riva2026architecture,
  title={{The architecture of cognitive amplification: Enhanced cognitive scaffolding as a resolution to the comfort-growth paradox in AI-supported learning}},
  author={Riva, Giuseppe},
  journal={Cognitive Systems Research},
  pages={101500},
  year={2026},
  publisher={Elsevier},
  doi={10.1016/j.cogsys.2026.101500}
}

@article{garcia2025epistemic,
  title={{Epistemic Partner or Cognitive Crutch: A Conceptual Model of Cognitive Offloading in Human-Artificial Intelligence Collaboration.}},
  author={Garc{\'\i}a-Barrios, David Andr{\'e}s},
  journal={IEEE Internet Comput.},
  volume={29},
  number={5},
  pages={7--16},
  year={2025},
  doi={10.1109/mic.2025.3626694}
}

@article{jain2026cognitive,
  title={{Cognitive offloading, critical thinking and attitudes towards artificial intelligence in the era of ChatGPT: a comparative study of artificial intelligence-assisted and manual task performance in young adults}},
  author={Jain, Purvi and Shajit, Deshna and Maini, Ansh and Shahin, Yasmeen and Panda, Ayush},
  journal={Cognitive Processing},
  pages={1--18},
  year={2026},
  publisher={Springer},
  doi={10.1007/s10339-026-01375-z}
}

@article{shin2023superhuman,
  title={Superhuman artificial intelligence can improve human decision-making by increasing novelty},
  author={Shin, Minkyu and Kim, Jin and Van Opheusden, Bas and Griffiths, Thomas L},
  journal={Proceedings of the National Academy of Sciences},
  volume={120},
  number={12},
  pages={e2214840120},
  year={2023},
  publisher={National Academy of Sciences},
  doi={10.1073/pnas.2214840120}
}

@inproceedings{coulom2006efficient,
  title={{Efficient selectivity and backup operators in Monte-Carlo tree search}},
  author={Coulom, R{\'e}mi},
  booktitle={International conference on computers and games},
  pages={72--83},
  year={2006},
  organization={Springer}
}

@article{hills2015exploration,
  title={Exploration versus exploitation in space, mind, and society},
  author={Hills, Thomas T and Todd, Peter M and Lazer, David and Redish, A David and Couzin, Iain D},
  journal={Trends in cognitive sciences},
  volume={19},
  number={1},
  pages={46--54},
  year={2015},
  publisher={Elsevier},
  doi={10.1016/j.tics.2014.10.004}
}

@article{charnov1976optimal,
  title={Optimal foraging, the marginal value theorem},
  author={Charnov, Eric L},
  journal={Theoretical population biology},
  volume={9},
  number={2},
  pages={129--136},
  year={1976},
  publisher={Elsevier},
  doi={10.1016/0040-5809(76)90040-X}
}

@article{levinthal1993myopia,
  title={The myopia of learning},
  author={Levinthal, Daniel A and March, James G},
  journal={Strategic management journal},
  volume={14},
  number={S2},
  pages={95--112},
  year={1993},
  publisher={Wiley Online Library},
  doi={10.1002/smj.4250141009}
}

@article{roberson2006colonel,
  title={The colonel blotto game},
  author={Roberson, Brian},
  journal={Economic Theory},
  volume={29},
  number={1},
  pages={1--24},
  year={2006},
  publisher={Springer},
  doi={10.1007/s00199-005-0071-5}
}

@article{levitt1988organizational,
  title={Organizational learning},
  author={Levitt, Barbara and March, James G},
  journal={Annual review of sociology},
  volume={14},
  number={1},
  pages={319--338},
  year={1988},
  publisher={Annual Reviews, 4139 El Camino Way, PO Box 10139, Palo Alto, CA 94303-0139, USA},
  doi={10.1146/annurev.so.14.080188.001535}
}

@article{chase1973perception,
  title={Perception in chess},
  author={Chase, William G and Simon, Herbert A},
  journal={Cognitive psychology},
  volume={4},
  number={1},
  pages={55--81},
  year={1973},
  publisher={Elsevier},
  doi={10.1016/0010-0285(73)90004-2}
}

@article{reitman1976skilled,
  title={{Skilled perception in Go: Deducing memory structures from inter-response times}},
  author={Reitman, Judith S},
  journal={Cognitive psychology},
  volume={8},
  number={3},
  pages={336--356},
  year={1976},
  publisher={Elsevier},
  doi={10.1016/0010-0285(76)90011-6}
}

@article{georgeot2012game,
  title={The game of go as a complex network},
  author={Georgeot, Bertrand and Giraud, Olivier},
  journal={EPL (Europhysics Letters)},
  volume={97},
  number={6},
  pages={68002},
  year={2012},
  doi={10.1209/0295-5075/97/68002}
}

@article{kandiah2014move,
  title={Move ordering and communities in complex networks describing the game of go},
  author={Kandiah, Vivek and Georgeot, Bertrand and Giraud, Olivier},
  journal={The European Physical Journal B},
  volume={87},
  number={10},
  pages={246},
  year={2014},
  publisher={Springer},
  doi={10.1140/epjb/e2014-50497-1}
}

@article{coquide2017distinguishing,
  title={{Distinguishing humans from computers in the game of go: A complex network approach}},
  author={Coquid{\'e}, C{\'e}lestin and Georgeot, Bertrand and Giraud, Olivier},
  journal={Europhysics Letters},
  volume={119},
  number={4},
  pages={48001},
  year={2017},
  publisher={EDP Sciences, IOP Publishing and Societ{\`a} Italiana di Fisica},
  doi={10.1209/0295-5075/119/48001}
}

@article{choi2025does,
  title={{How does artificial intelligence improve human decision-making? Evidence from the AI-powered Go program}},
  author={Choi, Sukwoong and Kang, Hyo and Kim, Namil and Kim, Junsik},
  journal={Strategic Management Journal},
  volume={46},
  number={6},
  pages={1523--1554},
  year={2025},
  publisher={Wiley Online Library},
  doi={10.1002/smj.3694}
}

@article{choi2025dual,
  title={{The dual edges of AI: Advancing knowledge while reducing diversity}},
  author={Choi, Sukwoong and Kang, Hyo and Kim, Namil and Kim, Junsik},
  journal={PNAS nexus},
  volume={4},
  number={5},
  pages={pgaf138},
  year={2025},
  publisher={Oxford University Press, US},
  doi={10.1093/pnasnexus/pgaf138}
}

@article{scheffer2001catastrophic,
  title={Catastrophic shifts in ecosystems},
  author={Scheffer, Marten and Carpenter, Steve and Foley, Jonathan A and Folke, Carl and Walker, Brian},
  journal={Nature},
  volume={413},
  number={6856},
  pages={591--596},
  year={2001},
  publisher={Nature Publishing Group UK, London},
  doi={10.1038/35098000}
}

@inproceedings{tromp2006combinatorics,
  title={{Combinatorics of Go}},
  author={Tromp, John and Farneb{\"a}ck, Gunnar},
  booktitle={International Conference on Computers and Games},
  pages={84--99},
  year={2006},
  organization={Springer},
  doi={10.1007/978-3-540-75538-8_8}
}

@article{gelly2012grand,
  title={{The grand challenge of computer Go: Monte Carlo tree search and extensions}},
  author={Gelly, Sylvain and Kocsis, Levente and Schoenauer, Marc and Sebag, Michele and Silver, David and Szepesv{\'a}ri, Csaba and Teytaud, Olivier},
  journal={Communications of the ACM},
  volume={55},
  number={3},
  pages={106--113},
  year={2012},
  publisher={ACM, New York, NY, USA},
  doi={10.1145/2093548.2093574}
}

@article{silver2018general,
  title={{A general reinforcement learning algorithm that masters chess, shogi, and Go through self-play}},
  author={Silver, David and Hubert, Thomas and Schrittwieser, Julian and Antonoglou, Ioannis and Lai, Matthew and Guez, Arthur and Lanctot, Marc and Sifre, Laurent and Kumaran, Dharshan and Graepel, Thore and others},
  journal={Science},
  volume={362},
  number={6419},
  pages={1140--1144},
  year={2018},
  publisher={American Association for the Advancement of Science},
  doi={10.1126/science.aar6404}
}

@article{gaessler2023training,
  title={{Training with AI: Evidence from chess computers}},
  author={Gaessler, Fabian and Piezunka, Henning},
  journal={Strategic Management Journal},
  volume={44},
  number={11},
  pages={2724--2750},
  year={2023},
  publisher={Wiley Online Library},
  doi={10.1002/smj.3512}
}

@article{beheim2025opening,
  title={{Opening strategies in the game of go from feudalism to superhuman AI}},
  author={Beheim, Bret},
  journal={Evolutionary Human Sciences},
  volume={7},
  pages={e28},
  year={2025},
  publisher={Cambridge University Press},
  doi={10.1017/ehs.2025.10016}
}

@inproceedings{wang2003epidemic,
  title={{Epidemic spreading in real networks: An eigenvalue viewpoint}},
  author={Wang, Yang and Chakrabarti, Deepayan and Wang, Chenxi and Faloutsos, Christos},
  booktitle={22nd international symposium on reliable distributed systems, 2003. proceedings.},
  pages={25--34},
  year={2003},
  organization={IEEE},
  doi={10.1109/RELDIS.2003.1238052}
}

@inproceedings{mcilroy2020aligning,
  title={{Aligning superhuman ai with human behavior: Chess as a model system}},
  author={McIlroy-Young, Reid and Sen, Siddhartha and Kleinberg, Jon and Anderson, Ashton},
  booktitle={Proceedings of the 26th ACM SIGKDD international conference on knowledge discovery \& data mining},
  pages={1677--1687},
  year={2020},
  doi={10.1145/3394486.3403219}
}

@article{brinkmann2023machine,
  title={Machine culture},
  author={Brinkmann, Levin and Baumann, Fabian and Bonnefon, Jean-Fran{\c{c}}ois and Derex, Maxime and M{\"u}ller, Thomas F and Nussberger, Anne-Marie and Czaplicka, Agnieszka and Acerbi, Alberto and Griffiths, Thomas L and Henrich, Joseph and others},
  journal={Nature Human Behaviour},
  volume={7},
  number={11},
  pages={1855--1868},
  year={2023},
  publisher={Nature Publishing Group UK, London},
  doi={10.1038/s41562-023-01742-2}
}

@misc{LeelaZero,
  author       = {Pascutto, Gian-Carlo and others},
  title        = {{Leela Zero}: open-source {Go} engine},
  year         = {2017},
  howpublished = {\url{https://github.com/leela-zero/leela-zero}},
  note         = {Version 0.17}
}

@misc{sgfmill,
  author       = {Woodcraft, Matthew},
  title        = {{Sgfmill: {Python}} library for {SGF} files and {Go} boards},
  howpublished = {\url{https://mjw.woodcraft.me.uk/sgfmill/}},
  note         = {Version 1.1.1}
}

@techreport{hagberg2008exploring,
  title={{Exploring network structure, dynamics, and function using NetworkX}},
  author={Hagberg, Aric and Swart, Pieter J and Schult, Daniel A},
  year={2008},
  institution={Los Alamos National Laboratory (LANL)},
  doi={10.25080/TCWV9851}
}

@inproceedings{wang2023adversarial,
  title={{Adversarial policies beat superhuman Go AIs}},
  author={Wang, Tony Tong and Gleave, Adam and Tseng, Tom and Pelrine, Kellin and Belrose, Nora and Miller, Joseph and Dennis, Michael D and Duan, Yawen and Pogrebniak, Viktor and Levine, Sergey and others},
  booktitle={International Conference on Machine Learning},
  pages={35655--35739},
  year={2023},
  organization={PMLR}
}

@inproceedings{kang2022ai,
  title={{How AI-based training affected the performance of professional Go players}},
  author={Kang, Jimoon and Yoon, June Seop and Lee, Byungjoo},
  booktitle={Proceedings of the 2022 CHI Conference on Human Factors in Computing Systems},
  pages={1--12},
  year={2022},
  doi={10.1145/3491102.3517540}
}

\end{document}